\documentclass[runningheads]{llncs}

\usepackage{eccv}

\usepackage{eccvabbrv}

\usepackage{graphicx}
\usepackage{booktabs}

\usepackage[accsupp]{axessibility}  

\usepackage[dvipsnames]{xcolor}
\usepackage{color, colortbl}
\usepackage{pifont}  
\usepackage{mathtools}
\usepackage{tabu}
\usepackage[inline]{enumitem}
\usepackage{threeparttable}
\usepackage{multirow}
\usepackage{tikz}
\usepackage[abs,overlay]{overpic}
\usepackage{subcaption}
\usepackage{float}

\newcommand{\ADEOoD}{ADE-OoD}

\newcommand{\fpr}{FPR}
\newcommand{\SETR}[1]{\color{Blue}\texttt{#1}}
\newcommand{\MF}[1]{\color{OliveGreen}\texttt{#1}}

\definecolor{cvprblue}{rgb}{0.21,0.49,0.74}

\definecolor{mylightgreen}{RGB}{51,153,102}
\definecolor{mydarkgreen}{RGB}{17,121,74}

\definecolor{mygreenblue}{RGB}{47,111,112}
\definecolor{mylightblue}{RGB}{172,207,208} 

\definecolor{myyellow}{RGB}{215,179,18} 
\definecolor{myorange}{RGB}{215,136,18}

\definecolor{mylightred}{RGB}{192,72,72} 
\definecolor{mydarkred}{RGB}{158,28,28} 

\definecolor{mylightgray}{RGB}{238,238,238} 
\definecolor{mymedgray}{RGB}{202,204,206} 
\definecolor{mydarkgray}{RGB}{42,44,46}

\definecolor{mymaxturbo}{RGB}{161,17,1}
\definecolor{myminturbo}{RGB}{57,42,119}

\definecolor{appearance}{RGB}{103,169,207}
\definecolor{positions}{RGB}{239,138,98}

\colorlet{bgcolor}{mylightgray}
\colorlet{poscolor}{mydarkgreen}
\colorlet{negcolor}{mydarkred}
\colorlet{outputscalingcolor}{blue}
\definecolor{draftcolor}{RGB}{0,73,95}

\definecolor{tabfirst}{rgb}{1, 0.7, 0.7} 
\definecolor{tabsecond}{rgb}{1, 0.85, 0.7} 
\definecolor{tabthird}{rgb}{1, 1, 0.7} 

\newcommand{\fakeparagraph}[1]{\noindent\textbf{#1}{ }}
\newcommand\blfootnote[1]{%
  \begingroup
  \renewcommand\thefootnote{}\footnote{#1}%
  \addtocounter{footnote}{-1}%
  \endgroup
}

\DeclareMathOperator*{\argmin}{arg\,min}
\newcommand{\norm}[1]{\left\lVert#1\right\rVert}

\usepackage{hyperref}

\usepackage{orcidlink}

\title{
Diffusion for Out-of-Distribution Detection\\on Road Scenes and Beyond
}

\author{Silvio Galesso$^*$\orcidlink{0009-0007-0249-6029}
\quad
Philipp Schröppel$^*$\orcidlink{0000-0003-0717-6896}
\quad
Hssan Driss
\quad
Thomas Brox\\
\footnotesize{$^*$Equal contribution}
}

\institute{
University of Freiburg, Germany \\
\email{\{galessos,schroepp\}@cs.uni-freiburg.de}
}

\authorrunning{S.~Galesso, P.~Schröppel et al.}

\begin{document}

\maketitle
\begin{abstract}

In recent years, research on out-of-distribution (OoD) detection for semantic segmentation has mainly focused on road scenes -- a domain with a constrained amount of semantic diversity.
In this work, we challenge this constraint and extend the domain of this task to general natural images. 
To this end, we introduce 
\textbf{1.} the \ADEOoD{} benchmark, which is based on the ADE20k dataset and includes images from diverse domains with a high semantic diversity, and
\textbf{2.} a novel approach that uses Diffusion score matching for OoD detection (DOoD) and is robust to the increased semantic diversity.

\ADEOoD{} features indoor and outdoor images, defines 150 semantic categories as in-distribution, and contains a variety of OoD objects.
For DOoD, we train a diffusion model with an MLP architecture on semantic in-distribution embeddings and build on the score matching interpretation to compute pixel-wise OoD scores at inference time. 
On common road scene OoD benchmarks, DOoD performs on par or better than the state of the art, without using outliers for training or making assumptions about the data domain. On \ADEOoD{}, DOoD outperforms previous approaches, but leaves much room for future improvements.
\blfootnote{\ADEOoD{} webpage: {{\url{https://ade-ood.github.io}}}}
\blfootnote{DOoD code: {{\url{https://github.com/lmb-freiburg/diffusion-for-ood}}}}
\end{abstract}    
\section{Introduction}

Deep neural networks achieve remarkable results when applied to their training domain, but their robustness and reliability on out-of-distribution (OoD) data is often inadequate.
Furthermore, they typically do not know that they do not know, \ie the complexity and overconfidence of neural networks makes it difficult to reliably detect whether inputs are out-of-distribution. In practice, this makes it hard to decide whether a model prediction can be trusted or not. For deployment in the real world, and especially in safety critical applications, this is a major problem; consequently OoD detection is an active research area.

This work falls in the category of \emph{out-of-distribution detection for semantic segmentation}. 
In this task, a semantic segmentation training dataset is given, consisting of images and segmentation ground truth maps, which assign each pixel to a semantic class. The set of classes in the training set are considered \textit{in-distribution}. Given an image at inference time, the task is to detect and localize entities that do not belong to the in-distribution classes via per-pixel \textit{out-of-distribution scores}.
Several benchmarks exist to evaluate performance on this task. The most common ones -- such as RoadAnomaly~\cite{lis2019detecting}, SegmentMeIfYouCan~\cite{segmentmeifyoucan2021}, and Fishyscapes Lost\&Found~\cite{blum2019fishyscapes, pinggera2016lost} -- are designed for the application of autonomous driving. Therefore, all samples in these benchmarks are from the constrained domain of road scenes and define the 19 road-specific classes (\textit{road}, \textit{car}, \etc) from Cityscapes~\cite{Cordts2016Cityscapes} as in-distribution.

\begin{figure}[!h]
\centering
    \begin{subfigure}[b]{0.15\textwidth}
    \centering
    \includegraphics[width=\textwidth]{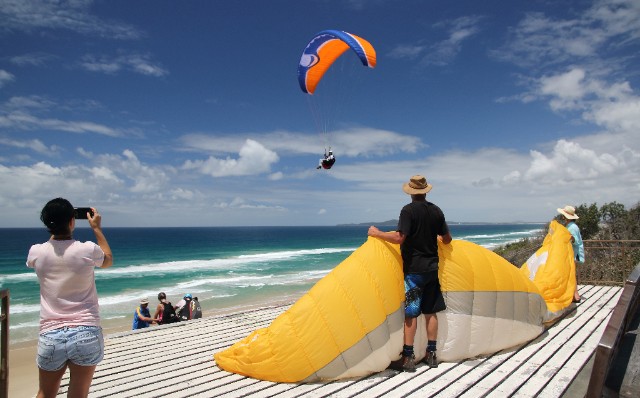}
    \end{subfigure}
    \begin{subfigure}[b]{0.14\textwidth}
    \centering
    \includegraphics[width=\textwidth]{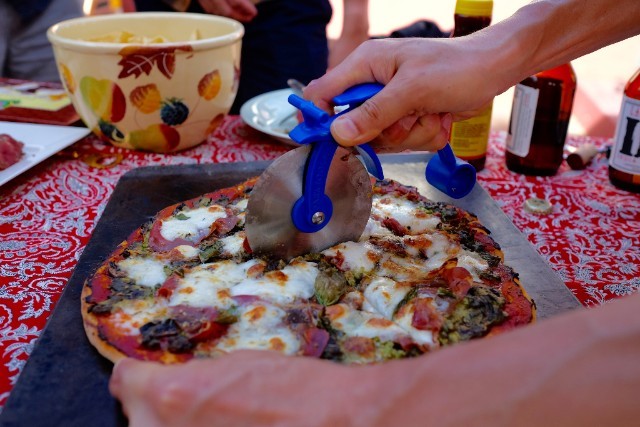}
    \end{subfigure}
    \begin{subfigure}[b]{0.14\textwidth}
    \centering
    \includegraphics[width=\textwidth]{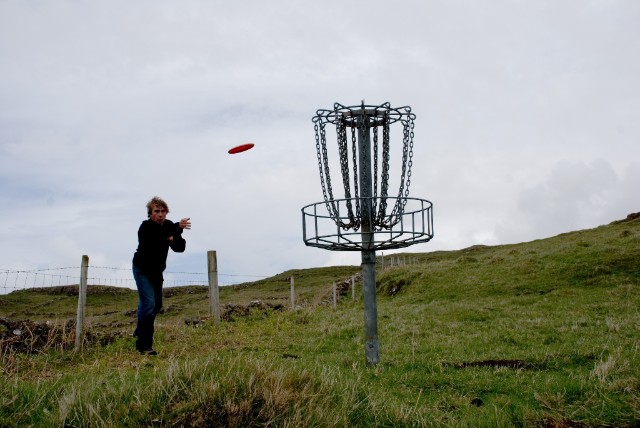}
    \end{subfigure}
    \begin{subfigure}[b]{0.143\textwidth}
    \centering
    \includegraphics[width=\textwidth]{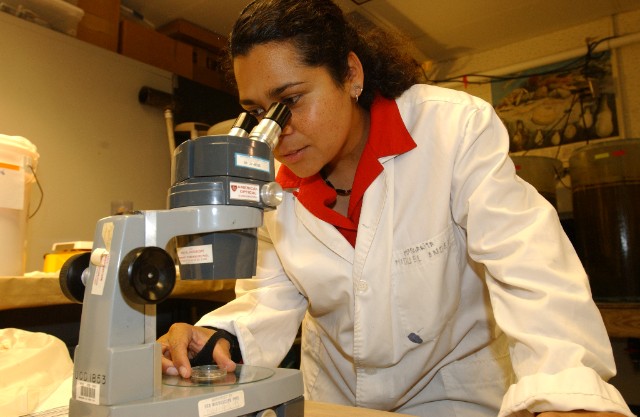}
    \end{subfigure}
    \begin{subfigure}[b]{0.125\textwidth}
    \centering
    \includegraphics[width=\textwidth]{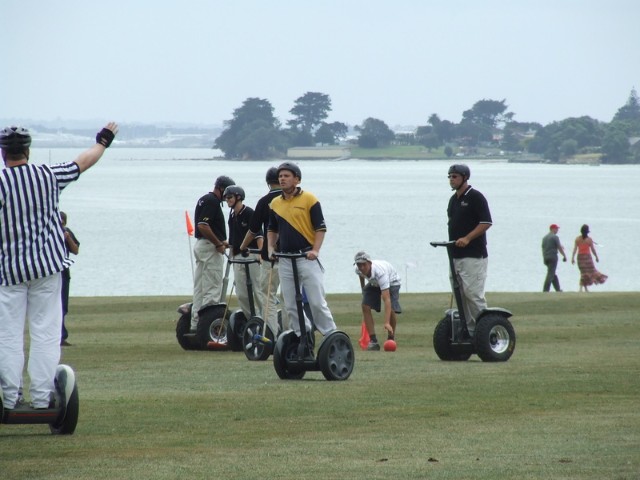}
    \end{subfigure}
    \begin{subfigure}[b]{0.125\textwidth}
    \centering
    \includegraphics[width=\textwidth]{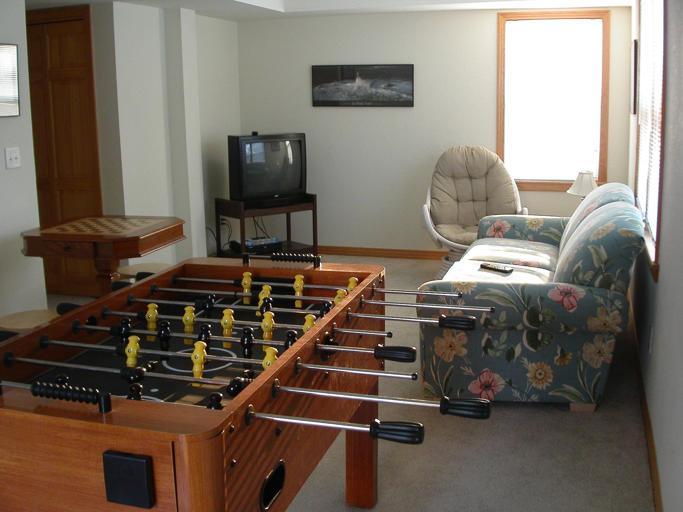}
    \end{subfigure}
    
    \begin{subfigure}[b]{0.15\textwidth}
    \centering
    \includegraphics[width=\textwidth]{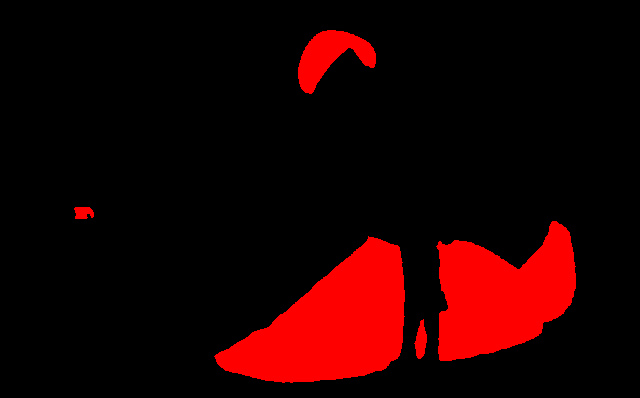}
    \end{subfigure}
    \begin{subfigure}[b]{0.14\textwidth}
    \centering
    \includegraphics[width=\textwidth]{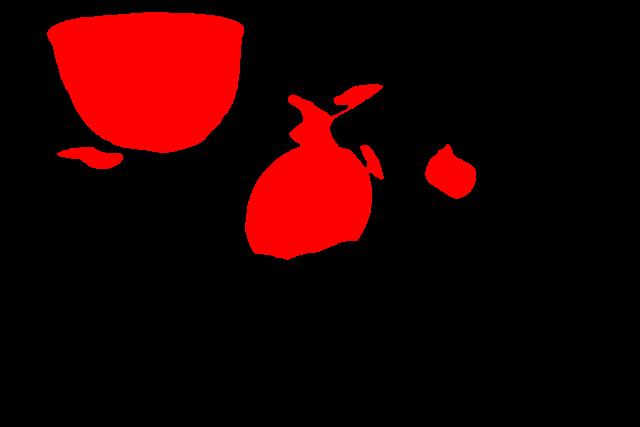}
    \end{subfigure}
    \begin{subfigure}[b]{0.14\textwidth}
    \centering
    \includegraphics[width=\textwidth]{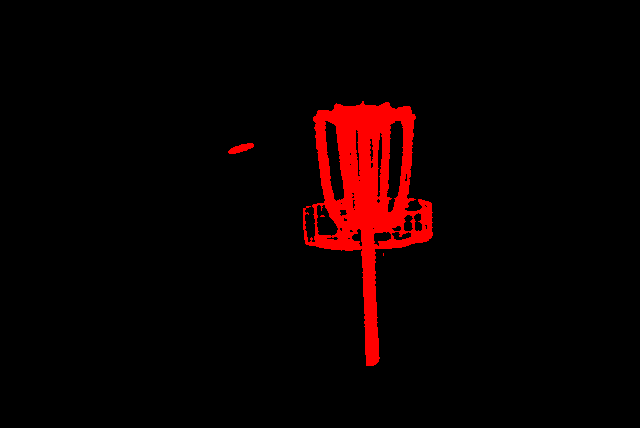}
    \end{subfigure}
    \begin{subfigure}[b]{0.143\textwidth}
    \centering
    \includegraphics[width=\textwidth]{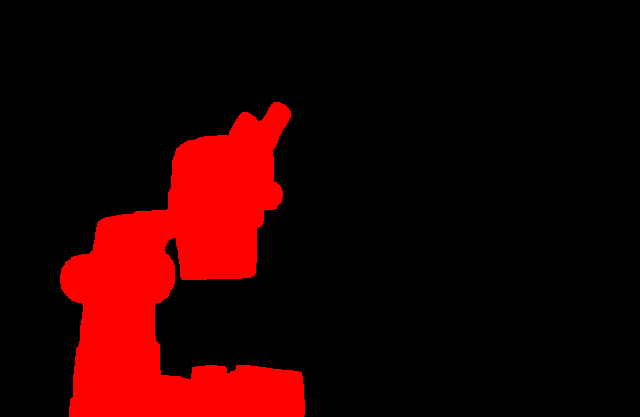}
    \end{subfigure}
    \begin{subfigure}[b]{0.125\textwidth}
    \centering
    \includegraphics[width=\textwidth]{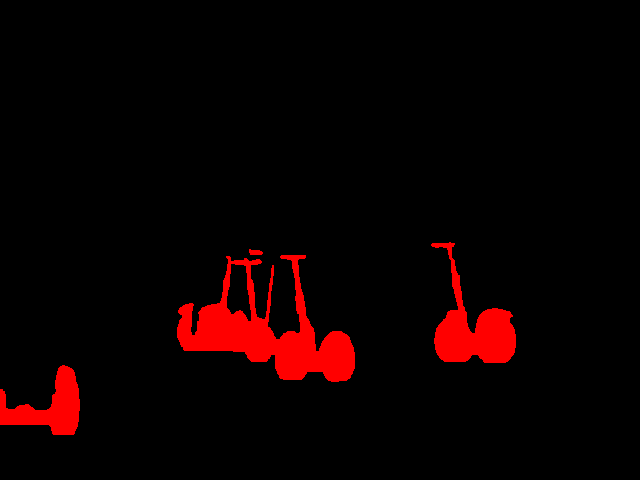}
    \end{subfigure}
    \begin{subfigure}[b]{0.125\textwidth}
    \centering
    \includegraphics[width=\textwidth]{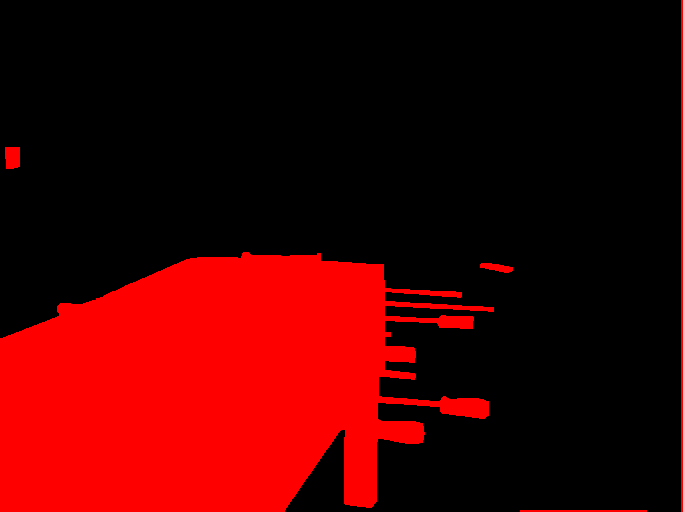}
    \end{subfigure}
  \caption{
  \textbf{Samples from the \ADEOoD{} benchmark.} The top row shows the input images, which feature diverse indoor and outdoor scenes. The bottom row shows the corresponding ground truth OoD segmentations 
  (red indicates OoD), 
  which contain all regions not covered by the 150 classes from ADE20k that we define as in-distribution. 
    }
    \label{fig:adeoodteaser}
\end{figure}

In this work, we extend this setting to \emph{general natural images}, which induces a higher complexity and semantic diversity. To this end, we introduce a novel benchmark called \ADEOoD{}. \ADEOoD{} defines the 150 classes from ADE20k~\cite{zhou2017scene} as in-distribution, which comprises a larger set of commonplace classes, for example \textit{computer}, \textit{streetlight}, or \textit{river}. 
The test images in \ADEOoD{} feature diverse indoor and outdoor scenes, including a high variety of OoD objects such as \textit{kites}, \textit{pizza cutters}, or \textit{disk golf baskets}. This poses a challenge for OoD detection methods, which need to model a more diverse in-distribution semantic landscape in order to successfully recognize the outliers. \cref{fig:adeoodteaser} shows samples from the proposed \ADEOoD{} benchmark.

\begin{figure*}[!t]
    \centering
    \includegraphics[width=0.98\textwidth]{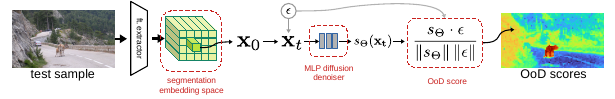}
    \caption{
    \textbf{Overview of DOoD.} We train a diffusion model on features extracted with a pre-trained feature extractor from in-distribution data. The diffusion model -- a small MLP -- is trained on \textit{individual feature vectors}, in order to discard harmful spatial correlations. 
    At inference time, we compute out-of-distribution scores by perturbing the input $\mathbf{x}_0$ with noise $\mathbf{\epsilon}$ via forward diffusion, estimating the gradient of the inlier log-density $s_\Theta$ with the diffusion model, and using the directional error as OoD score.
    }
    \label{fig:method_training}
\end{figure*}

In addition to that, we introduce a novel approach for OoD detection based on diffusion models, which works particularly well for modeling the highly diverse inlier distribution of the task at hand. 

A few recent works have applied diffusion models to out-of-distribution detection, albeit in different settings: classification, medical anomaly segmentation, and industrial inspection. 
Most of these approaches~\cite{graham2023denoising, liu2023unsupervised, goodier2023, Wolleb2022, Wyatt2022, Zhang_2023_ICCV, mousakhan2023anomaly, Gao_2023_ICCV, Wu_2023_ICCV} rely on reconstruction of the input data. This requires sequential application of multiple reverse diffusion process steps, which results in long runtimes.
A few approaches~\cite{shin2023, mahmood2020} build on the connection of diffusion models and score matching and compute OoD scores based on the estimated score from the diffusion model. In particular, Shin~\etal~\cite{shin2023} perturb the input data, estimate the score with the diffusion model, and use the mean-squared error between perturbations and estimated scores as OoD scores.

We follow a similar strategy, but propose to use the directional error between perturbations and estimated scores as OoD scores. We validate this choice with experiments and analyses, showing that it consistently outperforms previous ones. Further, instead of images, we apply the diffusion model to the latent distribution of features from \emph{deep segmentation embedding spaces}, which have been shown to be well-suited for OoD detection~\cite{knns}. Given the high spatial correlation between segmentation features -- which is not desirable for out-of-distribution detection -- we replace the common 2D U-Net with an MLP and \emph{learn to reconstruct individual features instead of spatial feature maps}. This way, the diffusion model is faster and cannot easily reconstruct anomalies by extrapolation.

We evaluate DOoD on the commonly used RoadAnomaly, Fishyscapes, and SMIYC-Anomaly benchmarks, and on the proposed \ADEOoD{} benchmark.
With the evaluations on these benchmarks, we show that our approach can detect and localize novel objects competitively or better than the state of the art. 
Our approach is especially effective when the novel objects are semantically anomalous, and when the test data has a large domain shift, such as in RoadAnomaly and \ADEOoD{}. This validates that the proposed adaptations make diffusion models effective for OoD detection on road scenes and beyond.

\section{Related Work}

In the following, we provide an overview of benchmarks and approaches for OoD detection in semantic segmentation data, and of diffusion model based approaches for other OoD detection tasks. 

\subsection{Out-of-Distribution Detection for Semantic Segmentation}
Detecting semantic out-of-distribution objects in segmentation data has risen in popularity in recent years, especially in the context of autonomous driving.

\fakeparagraph{Methods.}
Generative models have been used for out-of-distribution detection in many settings.
In semantic segmentation, some approaches use GANs for image re-synthesis, and use reconstruction quality as a proxy for anomaly scores~\cite{lis2019detecting,Xia2020,Biase2021}. Due to the implicit nature of GANs, these methods require comparison networks trained with proxy outliers, whereas our method can exploit the score estimate from the diffusion model.

A similar direction, but with non-parametric retrieval instead of parametric generation, is taken by cDNP~\cite{knns}, which shows that strong OoD detection results can be obtained with nearest-neighbor distances in certain deep segmentation embedding spaces. We build on their findings but improve upon it by using diffusion models as parametric density estimators instead of non-parametric kNNs.

Many recent methods for dense OoD detection use the Mask2Former~\cite{mask2former} segmentation model. Its separate mask/class prediction system is equipped with a rejection category, which was found to yield high quality anomaly scores~\cite{RbA,rai2023unmasking,Maskomaly,Grcic_2023_CVPR}, especially when exposed to outliers during training~\cite{hendrycks2018deep}. These works belong to a widespread practice of exploiting the prediction confidence of semantic segmentation models as a proxy score for anomaly~\cite{Jung_2021_ICCV,Chan2021EntropyMA,Grcic2021DenseAD,galesso2022probing,cen2021deep,Grcic2022,Tian2021,liang2022gmmseg, OVNNI, nekrasov2023ugains}.

Some approaches specifically target the driving domain: notably, JSRNet~\cite{jsrnet} achieves strong results on driving benchmarks by modelling the road surface. In contrast to ours, such methods are not applicable to other domains.

\fakeparagraph{Benchmarks.} 
Commonly used benchmarks are RoadAnomaly~\cite{lis2019detecting}, SegmentMeIfYouCan~\cite{segmentmeifyoucan2021}, and Fishyscapes Lost\&Found~\cite{blum2019fishyscapes, pinggera2016lost}. All of them focus on road scenes, which are limited in terms of scene structure and semantic categories (see overview in Sec.~\ref{sec:ade20k_ood}). 
Our proposed \ADEOoD{} benchmark uses the ontology of ADE20k, a large scale scene segmentation dataset, to overcome this constraint.

\subsection{Out-of-Distribution Detection with Diffusion Models}
\label{sec:related_ood_diffusion}
Multiple recent works have used diffusion models for OoD detection~\cite{graham2023denoising, liu2023unsupervised, goodier2023, Wolleb2022, Wyatt2022, Zhang_2023_ICCV, mousakhan2023anomaly, Gao_2023_ICCV, Wu_2023_ICCV, shin2023, mahmood2020}. The approaches differ in terms of downstream task and use of the diffusion model to derive OoD scores.

\fakeparagraph{Downstream Tasks.}
One set of works~\cite{graham2023denoising, liu2023unsupervised, goodier2023, Gao_2023_ICCV} applies diffusion models for OoD classification, where the task is to classify whole images at inference time as out-of-distribution. The approach DFDD~\cite{Wu_2023_ICCV} applies diffusion to OoD object detection, where the task is to detect OoD objects in form of bounding boxes.

Another set of works~\cite{Wolleb2022, Wyatt2022, Zhang_2023_ICCV, mousakhan2023anomaly, Gao_2023_ICCV} applies diffusion models for OoD segmentation, where the task is to predict a pixel-wise segmentation of OoD regions. All of these works are either in a medical or an industrial inspection domain. 

In this work, we present the novel approach for OoD segmentation on semantic segmentation data. In this task, a training dataset of images with semantic segmentation annotations is given. At inference time, the task is to segment OoD regions in images, where OoD is defined as not belonging to the set of training classes. 
Our setting differs from industrial and medical anomaly detection for two main reasons: the larger size of the input images (unsuitable for direct application of diffusion models), and the higher visual diversity and scene complexity which makes it more challenging to model the in-distribution data.

\fakeparagraph{OoD Score Computation.}
Most related works compute OoD scores based on reconstruction errors~\cite{graham2023denoising, liu2023unsupervised, Wolleb2022, Wyatt2022, Zhang_2023_ICCV, mousakhan2023anomaly, Gao_2023_ICCV}. This is done by corrupting the input image, reconstructing it with the diffusion model, and using the error between the input and the reconstruction as OoD score. 
Goodier~\etal~\cite{goodier2023} estimate the likelihood of given input data and use it as OoD score.
Most related to our work, are methods~\cite{shin2023,mahmood2020} that avoid expensive input reconstructions by leveraging the theory of denoising score matching~\cite{hyvarinen2005, a_connection, Song2019} and deriving OoD scores from estimated scores. Shin~\etal~\cite{shin2023} do this by perturbing the input data with Gaussian noise, estimating the score for the perturbed data point, and using its $L_2$ distance to the perturbation as OoD score.
Mahmood~\etal~\cite{mahmood2020} estimate the score for given data points and use the norm of the estimated score. In contrast to this, our approach computes OoD scores based on the directional difference of the estimated score and the perturbation.

\section{\ADEOoD{}: OoD Detection Beyond Road Scenes}
\label{sec:ade20k_ood}
In this section we introduce \ADEOoD{}, a novel benchmark for OoD detection on semantic segmentation data. The key characteristic of the proposed benchmark is higher diversity. It stands out by a much larger set of in-distribution categories than the typically used 19 Cityscapes classes. Furthermore, it contains images from more diverse indoors and outdoors scenes.

\subsubsection{Benchmark Overview.} The benchmark uses the training set of the ADE20k~\cite{zhou2017scene} dataset, with its 150 categories, to define the in-distribution domain; for this reason, the benchmark is named \ADEOoD{}. It consists of 111 real-world images, each containing in-distribution and out-of-distribution segments. For each image, a binary OoD mask is provided. A few examples are shown in \cref{fig:adeoodteaser} and \cref{fig:ade_results}. A comparison to other established benchmarks is provided in the following table:
\begin{table}[!h]
    \centering
    \scriptsize
    \begin{tabular}{|l|c|c|c|c|c|>{\columncolor{bgcolor}}c|}
    \hline
                   &    {\textbf{Road-}} & {\textbf{Fishyscapes}} & {\textbf{Fishyscapes}} & {\textbf{SMIYC}}  & {\textbf{SMIYC}}  & \textbf{\ADEOoD{}}    \\
                            &   {\textbf{Anomaly}} & {\textbf{L\&F}}      & {\textbf{Static}}      & {\textbf{Anomaly}} & {\textbf{Obstacle}}&   (\textbf{Ours})      \\
                
    \hline
    \hline
                \multirow{ 2}{*}{\textbf{Type}}        &   \multirow{ 2}{*}{Real}    &    \multirow{ 2}{*}{Real}       & Synthetic     &   \multirow{ 2}{*}{Real}    &  Real         &  Real  \\ 
                            &           &               & (pasted cutouts)&         &   (road only) &        \\ \hline
                \textbf{Domain}        &   Driving      & Driving       & Driving       & Driving    &   Driving      &   In \& outdoor \\ \hline
                \textbf{Num. Samples}        &   60      & 100+275       & 30+1000       & 10+100    &   327+30      &   111 \\ \hline
                \textbf{In-dist. classes} & 19   &   19          & 19            & 19        & 1            &   150 \\
    \hline
    \end{tabular}
\label{tab:benchmarks_overview}
\end{table}

\subsubsection{Benchmark Construction.} We obtained the images for the benchmark from the validation set of ADE20k and from OpenImages~\cite{OpenImages}. The images were selected according to the following criteria: \begin{enumerate*}[label=(\arabic*)] \item presence of clear and unambiguous in- and out-of-distribution entities only, \item presence of diverse in- and out-of-distribution categories, as well as varying indoor and outdoor settings \item adequate scene complexity to match the original ADE20k training data, \item collecting an adequate number of samples for statistically significant results\end{enumerate*}. For annotation, we have set up a semi-automatic tool, based on the popular promptable segmentation model SAM~\cite{Kirillov_2023_ICCV}.

\section{Diffusion Models Background}
\label{sec:diffusion_background}

In this section, we provide background information on score matching and diffusion models. Both are relevant for our approach DOoD, as we build on the score matching interpretation of diffusion models to compute OoD scores.

\subsection{Denoising Score Matching}
\label{sec:denoising_score_matching}
Let $q(\mathbf{x})$ be an unknown data distribution and $\{\mathbf{x}_i\in\mathbb{R}^D\}_{i=1}^N$ a dataset of samples from that distribution. Let $p(\mathbf{x}, \Theta)$ be a probability density model that should fit the data distribution. 
Score matching~\cite{hyvarinen2005} is an approach to learn the parameters $\Theta$ such that the score $\nabla_{\mathbf{x}} \log{p(\mathbf{x}, \Theta)}$ of the model distribution matches the score $\nabla_{\mathbf{x}} \log{q(\mathbf{x})}$ of the data distribution. In the following, we denote the score of the model distribution as $s_\Theta(\mathbf{x}) \coloneqq \nabla_{\mathbf{x}} \log{p(\mathbf{x}, \Theta)}$.

As the score of the data distribution is usually not known, techniques exist to match the score based on the given finite set of data samples~\cite{hyvarinen2005, a_connection}.
Denoising score matching~\cite{a_connection} is one such technique. It perturbs data points $\mathbf{x}$ according to a noise distribution $q(\tilde{\mathbf{x}}\mid \mathbf{x})$ and matches the score of the resulting perturbed distribution $q(\tilde{\mathbf{x}}) = \int q(\tilde{\mathbf{x}}\mid \mathbf{x}) q(\mathbf{x})\mathrm{d} \mathbf{x}$. It has been shown that matching the score of the perturbed distribution, amounts to matching the score $\nabla_{\tilde{\mathbf{x}}} \log q(\tilde{\mathbf{x}}\mid \mathbf{x})$~\cite{a_connection}.

Typically, Gaussian noise is used as perturbation: $q(\tilde{\mathbf{x}}\mid \mathbf{x})=\mathcal{N}(\tilde{\mathbf{x}} \mid \mathbf{x}, \sigma^2 \mathbf{I})$~\cite{a_connection, Song2019, Song2020}. Intuitively, this amounts to matching the score of a smoothed data distribution where the amount of smoothing depends on the chosen $\sigma$. 
With $\nabla_{\tilde{\mathbf{x}}} \log q(\tilde{\mathbf{x}}\mid \mathbf{x}) = - (\tilde{\mathbf{x}} - \mathbf{x} ) / \sigma^2$, this leads to the following objective:
\begin{equation}
   \hat{\Theta} = \argmin_{\Theta} \mathbb{E}_{q(\tilde{\mathbf{x}})}\bigg[ \frac{1}{2} \norm{  s_\Theta(\tilde{\mathbf{x}}) - \frac{-(\tilde{\mathbf{x}} - \mathbf{x})}{\sigma^2}  }_2^2  \bigg]\,.
   \label{eq:denoising_score_matching}
\end{equation}
Note that with a reparameterization $\tilde{\mathbf{x}} = \mathbf{x} + \sigma\mathbf{\epsilon}$ with $\mathbf{\epsilon}\sim \mathcal{N}(\mathbf{0}, \mathbf{I})$, one can reformulate this as matching the score $s_\Theta(\tilde{\mathbf{x}})$ to $-\mathbf{\epsilon} / \sigma$.

In the following, we provide background on Gaussian diffusion models, which can be interpreted as denoising score matching over Gaussian perturbations of different smoothing levels~\cite{ddpm}.

\subsection{Diffusion Models and their Relation to Score Matching}
\label{sec:diffusion_and_denoising_score_matching}
Diffusion models~\cite{Sohldickstein2015} are a class of generative models that learn a data distribution $q(\mathbf{x}_0)$ by defining a forward diffusion process in form of a Markov chain with steps $q(\mathbf{x}_t|\mathbf{x}_{t-1})$ that gradually transform the data distribution into a simple known distribution. A model $p_\theta(\mathbf{x}_{t-1}|\mathbf{x}_{t})$ with parameters $\theta$ is then trained to approximate the steps $q(\mathbf{x}_{t-1}|\mathbf{x}_{t})$ in the Markov chain of the reverse process.

In case of Gaussian diffusion models, the forward diffusion process gradually replaces the data with Gaussian noise following a noise schedule $\beta_1, .., \beta_T$:
\begin{equation}
q(\mathbf{x}_t|\mathbf{x}_{t-1}) \coloneqq \mathcal{N}(\mathbf{x}_t;\sqrt{1-\beta_t}\mathbf{x}_{t-1},\beta_t \mathbf{I})\,.
\label{eq:forwardprocess}
\end{equation}
Following this, $q(\mathbf{x}_t|\mathbf{x}_{0})$ is also a Gaussian that is parameterized as:
\begin{equation}
q(\mathbf{x}_t|\mathbf{x}_{0}) = \mathcal{N}(\mathbf{x}_t;\sqrt{\bar\alpha_t}\mathbf{x}_{0},(1-\bar\alpha_t) \mathbf{I})
\,\text{ with }\, 
\alpha_t \coloneqq 1-\beta_t 
\,\text{ and }\, 
\bar\alpha_t \coloneqq \prod_{s=1}^t \alpha_s\,.
\label{eq:forwardprocessskip}
\end{equation}

DDPM~\cite{ddpm} reparametrizes \cref{eq:forwardprocessskip} to 
\begin{equation}
\mathbf{x}_t(\mathbf{x}_0, \mathbf{\epsilon}) = \sqrt{\bar\alpha_t}\mathbf{x}_0 + \sqrt{1-\bar\alpha_t}\mathbf{\epsilon} \text{ with } \mathbf{\epsilon} \sim \mathcal{N}(\mathbf{0}, \mathbf{I}) 
\label{eq:forwardprocessskip2}
\end{equation}
and trains a model $\mathbf{\epsilon}_\theta(\mathbf{x}_t, t) $ to predict the noise $\mathbf{\epsilon}$ with the following loss:
\begin{equation}
   \hat{\Theta} = \argmin_{\Theta} \mathbb{E}_{\mathbf{x}_0, \mathbf{\epsilon}, t}\bigg[ \norm{   \mathbf{\epsilon}_\theta(\mathbf{x}_t, t)   -  \mathbf{\epsilon}  }_2^2 \bigg]\,.
   \label{eq:ddpmloss}
\end{equation}
Based on the similarity of this loss and the denoising score matching objective in Equation~\ref{eq:denoising_score_matching}, an estimated $\mathbf{\epsilon}_\theta(\mathbf{x}_t, t)$ can be interpreted as estimated score for the smoothed data distribution $q(\mathbf{x}_t)$ that arises at the diffusion timestep $t$:
\begin{equation}
\label{eq:score}
s_\Theta(\mathbf{x}_t)= \nabla_{\mathbf{x}_t} \log p(\mathbf{x}_t) = -\frac{\mathbf{\epsilon}_\theta(\mathbf{x}_t, t)} {\sigma_t} \quad\text{with }\,  \sigma_t = \sqrt{1-\bar\alpha_t}\,.
\end{equation}

In this work, we train a diffusion model on in-distribution data. The estimated score is hence trained to point towards regions of high-probability in smoothed inlier distributions, where different diffusion timesteps correspond to different amounts of smoothing. At inference time, we perturb given data $\mathbf{x}_0$ for different diffusion timesteps $t$ via $\mathbf{x}_t(\mathbf{x}_0, \mathbf{\epsilon}) = \sqrt{\bar\alpha_t}\mathbf{x}_0 + \sqrt{1-\bar\alpha_t}\mathbf{\epsilon}$ from Equation~\ref{eq:forwardprocessskip2}, estimate the score $s_\theta(\mathbf{x}_t)$, and use the directional difference of the estimated score $s_\theta(\mathbf{x}_t)$ and the perturbation $\mathbf{\epsilon}$ as OoD score.

\section{DOoD: Diffusion Score Matching for OoD Detection}

In the following, we describe DOoD, our proposed method for the detection and localization of out-of-distribution entities in semantic segmentation scenarios. As illustrated in \cref{fig:method_training}, the method consists of two steps: \begin{enumerate*}[label=(\arabic*)]\item feature extraction and \item OoD score estimation\end{enumerate*}. In the first step, we use a semantic segmentation model trained on in-distribution data to extract feature maps from the input images (Sec.~\ref{sec:ft_extraction}). For estimating the OoD scores, we train a diffusion model on the features extracted from the training dataset (Sec.~\ref{sec:denoiser_setup_training}). At inference time, we use this diffusion model to estimate the OoD scores (Sec.~\ref{sec:ood_detection_scores}).

\subsection{Feature Extraction}
\label{sec:ft_extraction}
To obtain feature representations suitable for OoD detection on segmentation data, we apply a feature extractor that was pretrained on in-distribution data of the given segmentation task. More specifically, we build on the findings from~\cite{knns}, which show that the self-attention features of transformer encoders induce meaningful distances between in-distribution and out-of-distribution features.
In particular, we employ the \textit{key} features of the  ViT~\cite{dosovitskiy2020image} and MiT~\cite{xie2021segformer} used as encoders in semantic segmentation models. These are trained with a standard supervised loss on the respective predictions from the Segmenter~\cite{strudel2021segmenter} or SETR~\cite{zheng2021rethinking} decoders.

After the feature extraction, our approach operates on dense feature maps of dimensions $\mathbf{F}\in\mathbb{R}^{H{\times}W{\times}C}$, \ie each feature vector has dimensionality $C$ and corresponds to one of the $H{\cdot}W$ square patches of the original image.

\subsection{Diffusion Model Architecture and Training}
\label{sec:denoiser_setup_training}
In the second stage of the method, we train a diffusion model on the features extracted from the training dataset of the in-distribution semantic segmentation data.
In the following, we describe the design of the diffusion model in terms of the denoiser architecture and the training procedure. 

\subsubsection{Architecture.}
Typical diffusion models for image generation use a convolutional 2D U-Net~\cite{ronneberger2015u} architecture for the denoiser network~\cite{ddpm, improved_ddpm}. This architecture works well for image generation, because it embeds spatial context information in the sampling process. For our application, however, information from neighboring features is harmful. This is because our approach relies on the diffusion model failing to estimate the score for OoD features; if the model can exploit the strongly correlated neighboring features it can accomplish this task regardless of whether the input is anomalous or not (as validated in \cref{fig:unet_analysis}).

As a remedy, we propose to use a MLP-based denoiser network that has an input dimensionality of $C$ and operates on each local feature individually. Our diffusion model hence models the distribution $q(\mathbf{x}_0)$ of individual features $\mathbf{x}_0\in\mathbb{R}^{C}$. This is illustrated in Figure~\ref{fig:method_training}.
The architecture of our MLP denoiser network has the same structure of residual and skip connections as the U-Net architecture in~\cite{improved_ddpm}. However, it replaces the convolutional layers with linear ones, maintains a constant hidden dimensionality, uses no attention layers, and uses a simplified time embedding. Further details are provided in the Appendix. 

\subsubsection{Training.}
\label{sec:training_denoiser}
Training of the denoiser is done with a DDPM formulation, \ie with the same noise schedule, model parametrization, and loss. We first extract feature maps $\mathbf{F}$ from all images in the training dataset and disassemble them into individual features $\mathbf{x}_0$. Those features represent samples of the inlier feature distribution $q(\mathbf{x}_0)$. We compute statistics across all features and use it for normalization to a range from $-1$ to $1$ before feeding data to the diffusion model. We use the same statistics for normalization at inference time. 

During the training, the features are perturbed with noise $\mathbf{\epsilon} \sim \mathcal{N}(\mathbf{0}, \mathbf{I})$ to $\mathbf{x}_t(\mathbf{x}_0, \mathbf{\epsilon})$ as stated in Eq.~\ref{eq:forwardprocessskip2}, and the denoiser is trained to estimate the perturbation noise $\mathbf{\epsilon}_\theta(\mathbf{x}_t, t)$ using the MSE loss from Eq.~\ref{eq:ddpmloss}. As described in the Background section, this can be interpreted as the denoiser learning to estimate the gradient of the smoothed data distribution $\mathbf{\epsilon}_\theta(\mathbf{x}_t, t) = -\sigma_t\nabla_{\mathbf{x}_t} \log p(\mathbf{x}_t)$ and the score $s_\Theta(\mathbf{x}_t)$ can be computed from the model output via $s_\Theta(\mathbf{x}_t)=-\mathbf{\epsilon}_\theta(\mathbf{x}_t, t)/\sigma_t$.

\subsection{OoD Score Computation}
\label{sec:ood_detection_scores}

\subsubsection{OoD Score Computation for a Single Diffusion Timestep.}
At inference time, given an input image, the goal is to estimate pixel-wise OoD scores. %
First, we extract the feature map $\mathbf{F}$ from the image and normalize it as described above. We disassemble the feature map to individual features $\mathbf{x}_0$ and compute an OoD score for each feature individually. 

For computing the OoD score of a single feature $\mathbf{x}_0$, we perturb the feature by applying the forward diffusion process  up to a timestep $t$: $\mathbf{x}_t(\mathbf{x}_0, \mathbf{\epsilon}) = \sqrt{\bar\alpha_t}\mathbf{x}_0 + \sqrt{1-\bar\alpha_t}\mathbf{\epsilon} \text{ with } \mathbf{\epsilon} \sim \mathcal{N}(\mathbf{0}, \mathbf{I})$ (same as Eq.~\ref{eq:forwardprocessskip}). We feed this perturbed feature $\mathbf{x}_t(\mathbf{x}_0, \mathbf{\epsilon})$ to our denoiser to estimate the score $s_\Theta(\mathbf{x}_t)=-\mathbf{\epsilon}_\theta(\mathbf{x}_t, t)/\sigma_t$ with $\sigma_t = \sqrt{1-\bar\alpha_t}$. Intuitively, this score points towards higher probability in the smoothed data distribution of the respective timestep. Based on this, we compute the OoD score $e_{\measuredangle,t}$ as directional error (\ie cosine similarity) between the estimated score $s_\Theta(\mathbf{x}_t)$ and the perturbation $\mathbf{\epsilon}$:
\begin{equation}
    e_{\measuredangle,t} = \frac{ s_\Theta(\mathbf{x}_t) \cdot {\epsilon}}{\lVert s_\Theta(\mathbf{x}_t)\rVert \, \lVert {\epsilon} \rVert} = 
    \frac{ - \frac{1}{\sigma_t} \mathbf{\epsilon}_\theta(\mathbf{x}_t, t) \cdot {\epsilon}}{\lVert - \frac{1}{\sigma_t} \mathbf{\epsilon}_\theta(\mathbf{x}_t, t) \rVert \, \lVert {\epsilon} \rVert} =
    \frac{-\mathbf{\epsilon}_\theta(\mathbf{x}_t, t)\cdot\epsilon}{\lVert \mathbf{\epsilon}_\theta(\mathbf{x}_t, t) \rVert \, \lVert \epsilon \rVert}\,.
\end{equation}
Finally, we reassemble the OoD scores $e_{\measuredangle,t}$ of each individual feature to an OoD score map $\mathbf{E}_{\measuredangle,t}$. In practice, the OoD scores for all features are computed in parallel as a single batch.

\subsubsection{OoD Score Aggregation Across Timesteps.}
We compute OoD score maps for multiple diffusion timesteps, \ie we compute $\mathbf{E}_{\measuredangle,t}$ for each $t\in\mathcal{T}=\{t_1, ..., t_N\}$. As the perturbations $\sqrt{1-\bar\alpha_t}\mathbf{\epsilon}$ in each timesteps have different expected magnitudes $\sqrt{1-\bar\alpha_t}$, we weight directional errors in different timesteps accordingly. The final OoD score map $\mathbf{E}_{\measuredangle}$ is therefore computed as a weighted sum as follows:
\begin{equation}
    \mathbf{E}_{\measuredangle} = \sum\limits_{t\in\mathcal{T}}\sqrt{1-\bar\alpha_t}\mathbf{E}_{\measuredangle,t}\,.
\end{equation}
In practice, again, the OoD score maps $\mathbf{E}_{\measuredangle,t}$ of the different timesteps are computed in parallel as a single batch.

Regarding reconstruction-based approaches, the advantage of this OoD score computation is that all steps can be parallelized, which enables much faster runtimes. The difference to previous score-based methods is that here the OoD scores are computed from directional errors and aggregated across multiple timesteps. In \cref{sec:scores_comparison} we show experimentally that this improves performance consistently.

\subsubsection{Compound OoD Scores.}
Following~\cite{knns}, we average the OoD scores $\mathbf{E}_{\measuredangle,t}$ obtained from the diffusion model with the uncertainty scores of the segmentation model (in the form of the LogSumExp of the predicted logits). The average is computed after standardization according to means and standard deviations obtained on the training set.
\section{Experiments}

In this section, we provide experimental results for the presented method and for the proposed benchmark. In \cref{sec:sota_driving}, we compare our method to the state-of-the-art in OoD detection for semantic segmentation on road scene benchmarks, which are the current standard in the field. In \cref{sec:sota_ade}, we provide evaluations on the proposed \ADEOoD{} benchmark.
Following that, we validate the principal design choices of our approach, namely the use of an MLP architecture for the denoiser network (\cref{sec:mlp_vs_unet}), and the proposed OoD score computation (\cref{sec:scores_comparison}).

\subsubsection{Experimental Setup.}
\label{sec:exp_setup}
To train the segmentation models for feature extraction, we follow~\cite{knns} or use off-the-shelf pre-trained parameters whenever available. For the MLP architecture of our diffusion model, we use a hidden dimension equal to the input feature channel size. We use a learning rate of $5e\!\!-\!\!5$ and a batch size of 4096. We train for 70k iterations, which takes ca. 5 hours on an RTX 3090 GPU. We aggregate OoD scores across the last 25 timesteps, \ie $\mathcal{T} = \{1, ..., 25\}$. Please note that we use the same experimental settings across all benchmarks. 

We report the standard evaluation metrics Average Precision (AP) and FPR at $95\%$TPR (\fpr). As the results vary strongly across datasets, we additionally report the average AP and FPR rank across benchmarks for each method.

\subsection{Comparing DOoD to the State-of-the-Art on Road Scenes}
\label{sec:sota_driving}
In this section we evaluate DOoD on the common driving-oriented benchmarks RoadAnomaly~\cite{lis2019detecting}, Fishyscapes\cite{blum2019fishyscapes,pinggera2016lost} validation, and SegmentMeIfYouCan~\cite{segmentmeifyoucan2021} (SMIYC). In all cases, the Cityscapes dataset is used for training and the Cityscapes categories are considered as in-distribution.

\subsubsection{Compared Methods.} We compare with a diverse set of state-of-the-art methods. 
Notably, many are similarly based on Mask2Former architecture (Rba, M2A, EAM, Maskomaly). 
Our evaluation includes approaches that use external datasets as proxy outliers for negative training (Outlier Exposure/OE) and approaches which are designed specifically for driving (DS) scenarios. These methods build on data or assumptions that make them less general (as detailed in \cref{sec:sota_ade}) and therefore not directly comparable. For example the DS method M2A suppresses OoD scores for all ``stuff'' classes except ``road'', and JSRNet explicitly models the road surface. We therefore included them in the evaluation for completeness, but excluded them from the ranking. 

The compared methods build on diverse segmentation models. Following~\cite{knns}, we apply our approach on the representation spaces of two different segmentation model architectures: Segmenter-B and SETR-L. These are popular high-performance models with pre-trained Cityscapes weights available off-the-shelf. Since SMIYC is submission-based, we report results for one model per dataset.

\subsubsection{Results.} Quantitative results are provided in \cref{tab:sota}.
DOoD performs as good or better than the state of the art on the RoadAnomaly, SMIYC-Anomaly, and Fishyscapes Static benchmarks. Further, on RoadAnomaly and SMIYC-Anomaly, our approach challenges most methods that are driving-specific or use outlier exposure, even though they require additional data or are less general. 

On Fishyscapes Lost\&Found the performance of our method varies depending on which segmentation model is used as feature extractor. With the \mbox{SETR-L} feature extractor, our approach is competitive on the AP metric with all comparable approaches. We attribute the higher \fpr{} metric of our approach to the fact that our estimated OoD score maps are of lower resolution (as we operate on transformer feature maps), which is unfavorable with the small-sized obstacles featured in Lost\&Found, resulting in a higher \fpr. The same applies to SMIYC-Obstacle, which also features small-sized anomalies.

Overall, we observe that no method is capable to prevail on all benchmarks; the ones which score excellently on some, perform poorly on others. In comparison with others, the results of DOoD are more consistent and give best average performance across all benchmarks, as indicated by the Average Rank.

\begin{table*}[!h]
    \centering
\vspace{-0.3cm}
    \caption{
    \textbf{State-of-the-art comparison on common road scene benchmarks.} While no approach is superior on all benchmarks and metrics, our approach DOoD often performs better or on-par with the state-of-the-art, which is indicated by the strong Average Rank across benchmarks. ``OE'' (Outlier Exposure) and ``DS'' (Driving Specific) methods have higher data requirements or stricter assumptions, which makes them less general and therefore not directly comparable (as shown in \cref{sec:sota_ade}). Nevertheless, our approach is competitive with them on RoadAnomaly and SMIYC-Anomaly. 
    }
    
    \scriptsize

\begin{threeparttable}
        \begin{tabular}{l|c|c|rr|rr|rr|rr|rr|rr|rr} 
    \toprule
                    &    &    & \multicolumn{2}{c|}{Road-} & \multicolumn{2}{c|}{Fishysc.} & \multicolumn{2}{c|}{Fishysc.} & \multicolumn{2}{c|}{SMIYC} & \multicolumn{2}{c|}{SMIYC LF} & \multicolumn{2}{c|}{SMIYC} & \multicolumn{2}{c}{Average} \\
                    &    &    & \multicolumn{2}{c|}{Anomaly} & \multicolumn{2}{c|}{L\&F val.} & \multicolumn{2}{c|}{Static val.} & \multicolumn{2}{c|}{Anomaly} & \multicolumn{2}{c|}{NoKnown} & \multicolumn{2}{c|}{Obstacle}& \multicolumn{2}{c}{Rank} \\
        Method      & OE & DS & \multicolumn{1}{c}{AP} & \multicolumn{1}{c|}{\fpr} & \multicolumn{1}{c}{AP} & \multicolumn{1}{c|}{\fpr} & \multicolumn{1}{c}{AP} & \multicolumn{1}{c|}{\fpr} & \multicolumn{1}{c}{AP} & \multicolumn{1}{c|}{\fpr} & \multicolumn{1}{c}{AP} & \multicolumn{1}{c|}{\fpr} & \multicolumn{1}{c}{AP} & \multicolumn{1}{c|}{\fpr} & \multicolumn{1}{c}{AP} & \multicolumn{1}{c}{\fpr} \\
        \midrule
        GMMSeg~\cite{liang2022gmmseg}      &            &              & 57.7 & 44.3 & 50.0 & \cellcolor{tabthird}12.6 & - & - & - & - & - & - & - & - & 5.0 & 4.5 \\
        NFlowJS~\cite{Grcic2021DenseAD}     &           &               & - & - & 39.4 & \cellcolor{tabfirst}9.0 & - & - & 56.9 & 34.7 & \cellcolor{tabfirst}89.3 & \cellcolor{tabfirst}0.7 & \cellcolor{tabsecond}85.5 & \cellcolor{tabfirst}0.4 & 3.5 & \cellcolor{tabfirst}2.0 \\
        Maskomaly~\cite{Maskomaly}                           &           &               &  80.8 & 12.0 & - & - & \cellcolor{tabthird}68.8 & \cellcolor{tabthird}15.0 & \cellcolor{tabfirst}93.4 & \cellcolor{tabfirst} 6.9  & - & - & - & - & 2.67 & \cellcolor{tabthird}2.67 \\
        RbA~\cite{RbA}                      &            &              & 78.5 & \cellcolor{tabthird}11.8 & \cellcolor{tabthird}61.0 & \cellcolor{tabsecond}{10.6} & $^{\dagger}$59.1 & $^{\dagger}$17.7 & 86.1 & 15.9 & - & - & \cellcolor{tabfirst}87.9 & 3.3 & 3.6 & 3.4 \\
        cDNP~\cite{knns}                    &            &              & \cellcolor{tabthird}85.6 & \cellcolor{tabsecond}9.8 & \cellcolor{tabfirst}{66.2} & 27.8 & $^{\dagger}$63.1 & $^{\dagger}$19.3 & \cellcolor{tabthird}88.9 &\cellcolor{tabthird} 11.4 & - & - & \cellcolor{tabthird}72.7 & \cellcolor{tabsecond}1.4 & \cellcolor{tabthird}2.8 & 3.2 \\
        \hline
\color{black}EAM~\cite{Grcic_2023_CVPR}      & \color{black}\checkmark & \color{black}              & \color{black}69.4 & \color{black}7.7 & \color{black}81.5 & \color{black}4.2 & \color{black}- & \color{black}- & \color{black}93.8 & \color{black}4.1 & \color{black}- & \color{black}- & \color{black}92.9 & \color{black}0.5 & \color{black}- & \color{black}- \\
\color{black}M2A~\cite{rai2023unmasking}         & \color{black}\checkmark & \color{black}\checkmark   & \color{black}79.7 & \color{black}13.5 & \color{black}46.0 & \color{black}4.4 & \color{black}90.5 & \color{black}1.2 & \color{black}88.7 & \color{black}14.6 & \color{black}- & \color{black}- & \color{black}92.9 & \color{black}0.5 & \color{black}- & \color{black}- \\
\color{black}RbA-OE ~\cite{RbA}                    & \color{black}\checkmark & \color{black}              & \color{black}85.4 & \color{black}6.9  & \color{black}70.8 & \color{black}6.3  & \color{black}75.5 & \color{black}3.5 & \color{black}94.5 & \color{black}4.6 & \color{black}- & \color{black}- & \color{black}95.1 & \color{black}0.1 & \color{black}- & \color{black}- \\
\color{black}RPL~\cite{liu2023residual} & \checkmark & & 71.6 & 17.7 & 70.6 & 2.5 & 92.5 & 0.9 & 83.5 & 11.7 & - & - & 85.9 & 0.6 & - & -\\
\color{black}JSRNet~\cite{jsrnet}                  & \color{black}            & \color{black}\checkmark   & \color{black}94.4 & \color{black}9.2 & \color{black}- & \color{black}- & \color{black}- & \color{black}- & 33.6 &
43.9 & \color{black}74.2 & \color{black}6.6 & \color{black}28.1 & \color{black}28.9 & \color{black}- & \color{black}- \\
        \hline
        Ours Segm.-B &          &               & \cellcolor{tabfirst}{89.1} & \cellcolor{tabfirst}{8.8}  & 43.0 & 30.2 & \cellcolor{tabsecond}75.0 & \cellcolor{tabfirst}6.2 & \cellcolor{tabsecond}{90.8} & \cellcolor{tabsecond}{8.4} & - & - & - & - & \cellcolor{tabsecond}2.5 & \cellcolor{tabsecond}2.5 \\
        Ours SETR-L  &          &               & \cellcolor{tabsecond}88.6 & 13.1 & \cellcolor{tabsecond}{64.4} & 27.7 & \cellcolor{tabfirst}79.3 & \cellcolor{tabsecond}6.5 & - & - & \cellcolor{tabsecond}79.5 & \cellcolor{tabsecond}3.7 & 64.3 & \cellcolor{tabthird}2.6 & \cellcolor{tabfirst}2.2 & 3.4 \\
    \bottomrule
    \end{tabular}
\begin{tablenotes}\tiny 
\item Ranking: \colorbox{tabfirst}{best}, \colorbox{tabsecond}{second}, \colorbox{tabthird}{third} \hspace{50px}  $\dagger$ Computed using official code and model. 
\end{tablenotes}
\end{threeparttable}
\label{tab:sota}
\vspace{-0.3cm}
\end{table*}

\subsection{\ADEOoD{} Evaluation}
\label{sec:sota_ade}

In this section we provide evaluations on the proposed \ADEOoD{} benchmark. 

\subsubsection{Evaluated Methods.} We report results for a diverse set of methods, including the hybrid generative-discriminative GMMSeg, the Mask2Former based RbA and M2A, and cDNP.
For M2A category partitioning we follow ADE20k's stuff/things split.
We use off-the-shelf ADE20k weights for all methods, matching the architectures used in the respective paper for driving data. 
We use SETR-ViT-L for our approach, as it is compatible to our previous experiments and available for download.
Note that it is difficult to evaluate many methods that use outlier exposure or are specific to driving scenes, as they require additional outlier data or their assumptions do not apply. 

\begin{figure}[!h]
 \centering
\caption{\textbf{ADE-OoD Evaluation.} 
        In the \textbf{Table} we provide results on the proposed benchmark for a diverse set of approaches, including parametric generative (GMMSeg), retrieval-based (cDNP), and Mask2Former-based (RbA, M2A, Maskomaly). Our approach DOoD performs second best in terms of AP, and best in terms of FPR. {Overall there is much room for improvement for all methods, which testifies the challenge posed by \ADEOoD{}}. In the \textbf{Figure} we show three samples from the benchmark, along with OoD score maps from GMMSeg, RbA, and our approach. 
        }
\qquad
\begin{subfigure}[b]{0.3\textwidth}
        \centering
        \small
        \begin{tabular}{l|r|r}
        \toprule
            Method & \multicolumn{1}{c|}{AP} & \multicolumn{1}{c}{\fpr} \\
            \midrule
            M2A         & 18.23 & 86.53 \\
            Maskomaly   & 28.23 & 82.68 \\
            GMMSeg      & 47.6 & \cellcolor{tabthird} 43.5 \\
            cDNP        & \cellcolor{tabthird} 62.35 & \cellcolor{tabsecond} 39.20 \\
            RbA         & \cellcolor{tabfirst} 66.82 & 82.42 \\
            Ours        & \cellcolor{tabsecond} 63.03 & \cellcolor{tabfirst}36.50 \\
        \bottomrule
        \end{tabular}
        \label{tab:ade_results}
\end{subfigure}
\hfill
\begin{subfigure}[b]{0.6\textwidth}
\input{figures/examples/ade20k_ood/ade20k_ood_examples}
\vspace{4px}
\end{subfigure}
\label{fig:ade_results}
\vspace{-0.5cm}
\end{figure}

\subsubsection{Results.} Quantitative results are reported in the Table in \cref{tab:ade_results}. In terms of AP, the best approaches are RbA and our approach DOoD, followed by cDNP. M2A has a lower performance: its strategy of restricting its scope to foreground classes works well in the driving domain, but is undesirable here. In terms of \fpr{} our approach performs best, and surprisingly RbA ranks last.

In \cref{fig:ade_results} we show some samples from the benchmark, along with predicted OoD scores from GMMSeg, RbA and our method.
Neither approach perfectly segments the anomalies, showcasing the difficulty of the benchmark.
The examples show how RbA's predictions are smooth within objects, but they incur in more critical failure cases, such as the false negatives in the last example.
On the other hand, our method's outputs are of lower resolution and noisier.

\subsection{Analysis of the Diffusion Model Architecture}
\label{sec:mlp_vs_unet}
\begin{figure}[!h]
\begin{subfigure}[t]{0.48\textwidth}
\centering
    \begin{subfigure}[t]{0.18\textwidth}
        \centering
        \includegraphics[width=\textwidth]{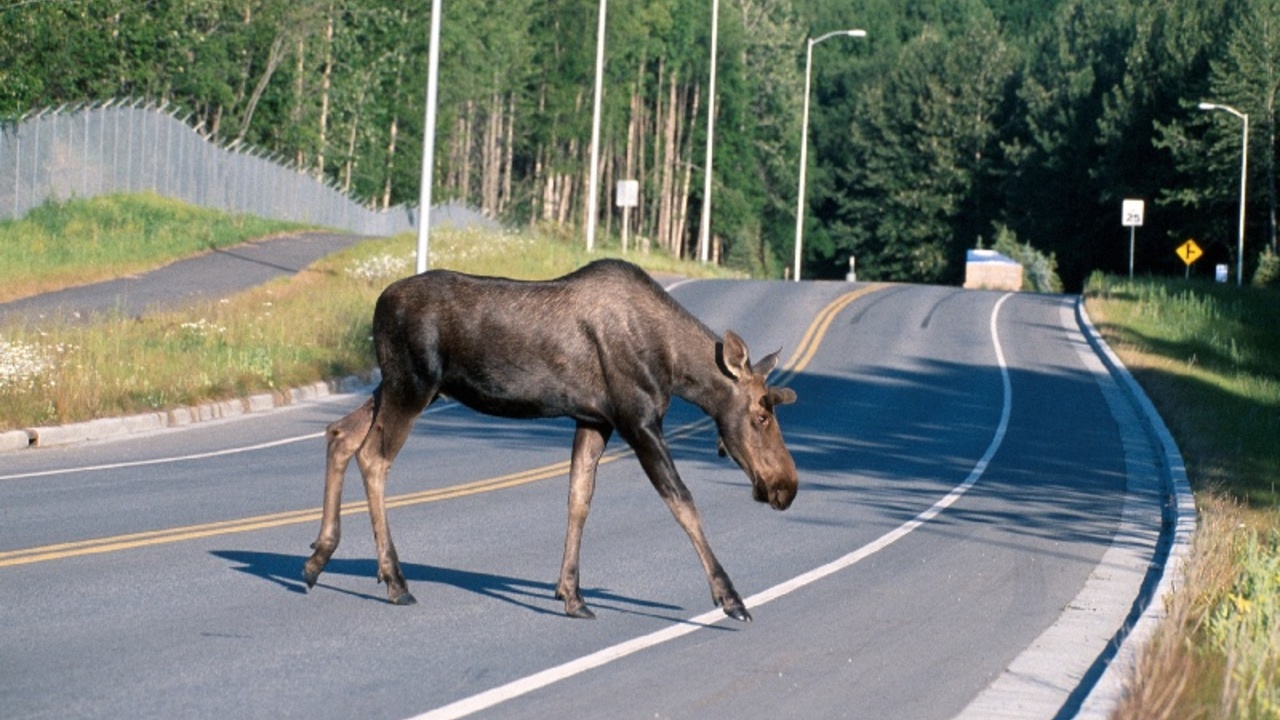}
    \end{subfigure}
    \begin{subfigure}[t]{0.18\textwidth}
        \centering
        \includegraphics[width=\textwidth]{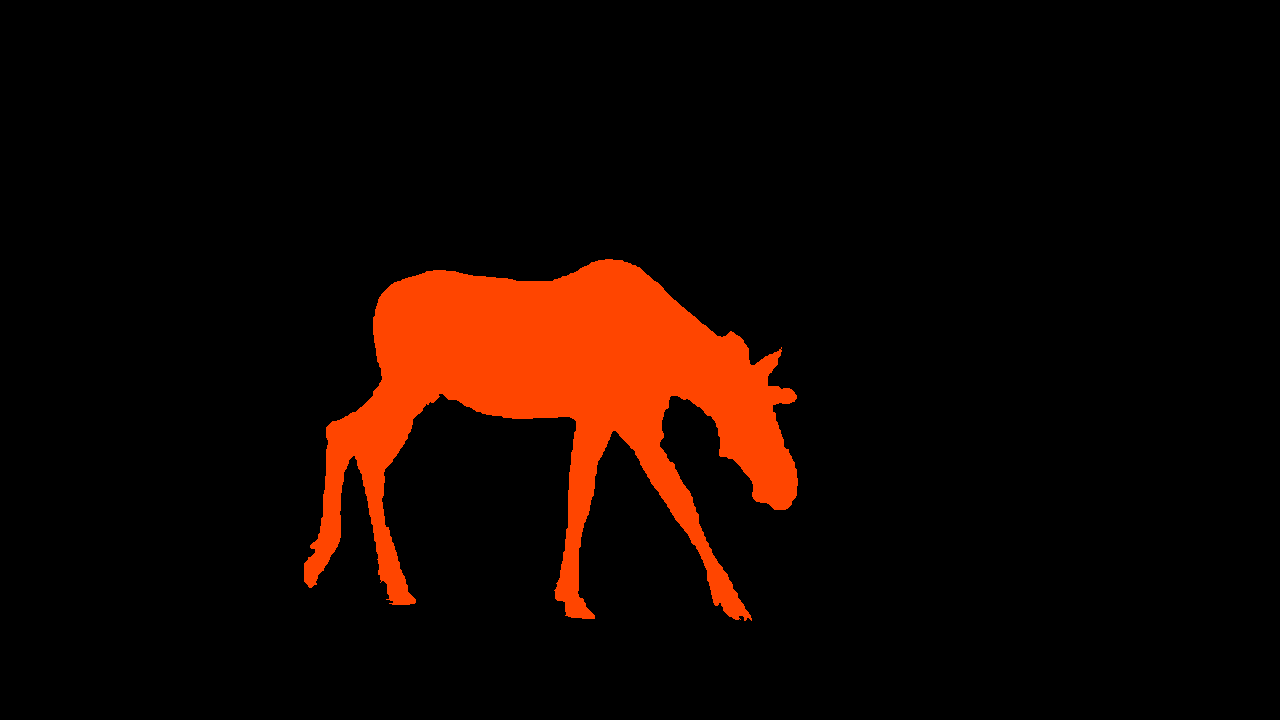}
    \end{subfigure}
    \begin{subfigure}[t]{0.17\textwidth}
        \centering
        \includegraphics[width=\textwidth]{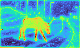}
    \end{subfigure}
    \begin{subfigure}[t]{0.17\textwidth}
        \centering
        \includegraphics[width=\textwidth]{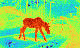}
    \end{subfigure}

    \begin{subfigure}[t]{0.18\textwidth}
    \captionsetup{labelformat=empty}
        \centering
        \includegraphics[width=\textwidth]{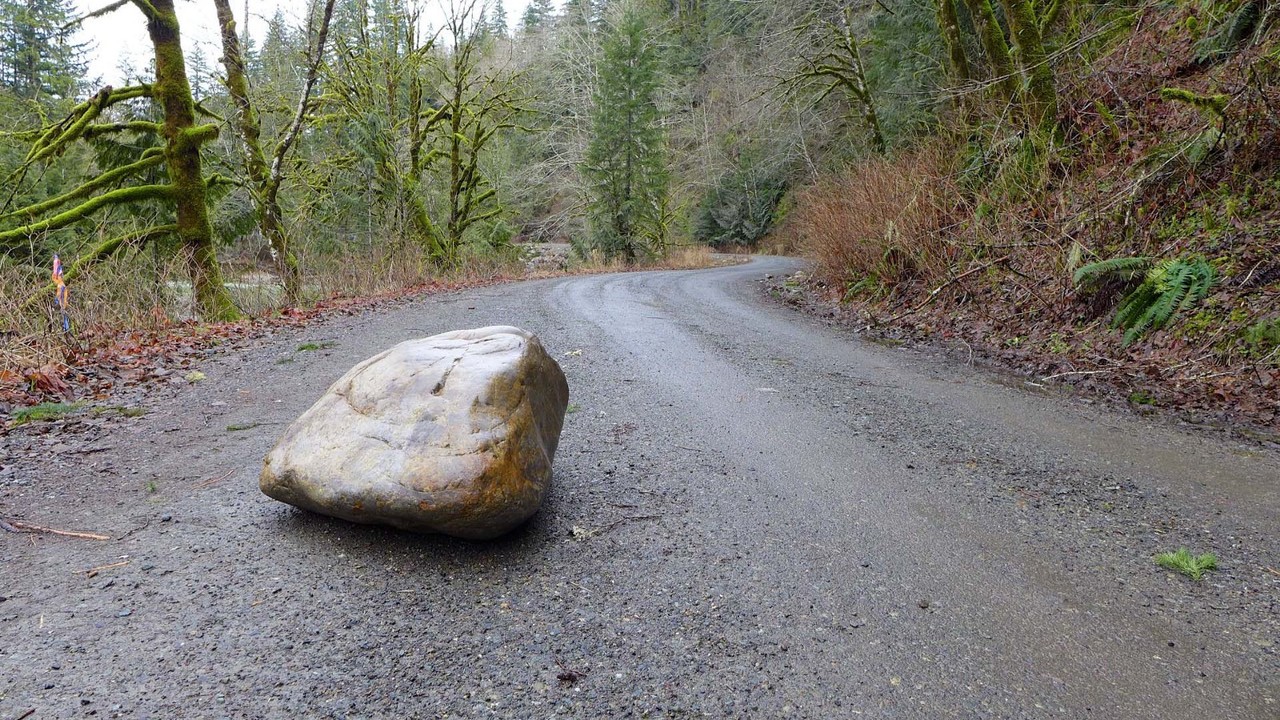}
        \caption*{\tiny Image}
    \end{subfigure}
    \begin{subfigure}[t]{0.18\textwidth}
    \captionsetup{labelformat=empty}
        \centering
        \includegraphics[width=\textwidth]{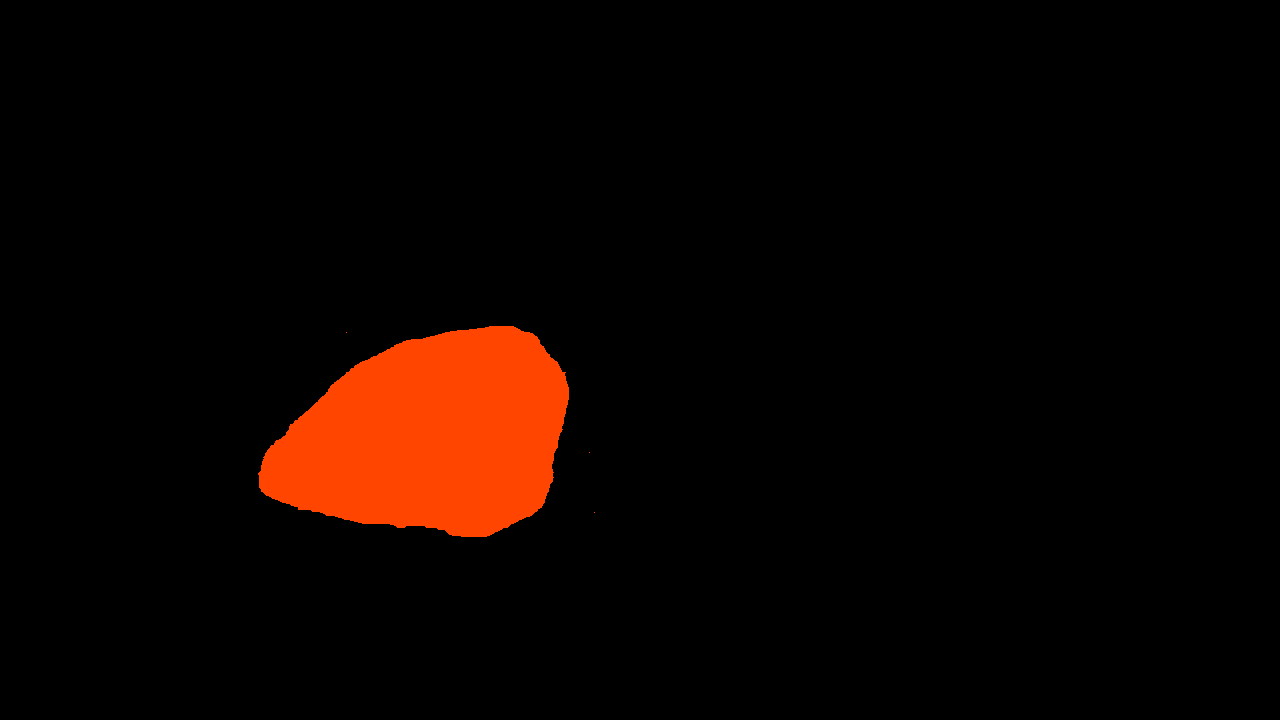}
        \caption*{\tiny G.T.}
    \end{subfigure}
    \begin{subfigure}[t]{0.17\textwidth}
    \captionsetup{labelformat=empty}
        \centering
        \includegraphics[width=\textwidth]{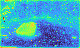}
        \caption*{\tiny U-Net}
    \end{subfigure}
    \begin{subfigure}[t]{0.17\textwidth}
    \captionsetup{labelformat=empty}
        \centering
        \includegraphics[width=\textwidth]{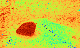}
        \caption*{\tiny MLP}
    \end{subfigure}

    \label{fig:unet_vs_mlp}
\end{subfigure}
  \begin{subfigure}[t]{0.45\textwidth}
    \centering
    \scriptsize
    \begin{tabular}{l|c|c|c|c}
    \toprule
        & \multicolumn{2}{c|}{RoadAnomaly} & \multicolumn{2}{c}{FS-L\&F} \\
    Arch. & AP & \fpr & AP & \fpr \\
    \midrule
    U-Net   & 67.3 & 19.8 & 25.9 & 43.0 \\
    MLP    & \textbf{84.7} & \textbf{15.3} & \textbf{31.5} & \textbf{41.1} \\
    \bottomrule
    \end{tabular}
    \label{tab:unet_vs_mlp}
  \end{subfigure}
  \caption{
    \textbf{Comparison of MLP and U-Net diffusion model architectures.}}
    \label{fig:unet_analysis}
\end{figure}

As described in \cref{sec:denoiser_setup_training}, we use an MLP architecture for the diffusion model. In Fig.~\ref{fig:unet_analysis}, we provide a comparison to the commonly used U-Net architecture.

The OoD scores from the U-Net architecture can be correct on boundaries of OoD objects, but too low within the object. We attribute this to the receptive field of the U-Net, which allows it to extrapolate information from neighbor features and as such successfully denoise OoD inputs, which is not desirable. The MLP architecture operates on each feature individually and does not suffer from this problem. Indeed, the proposed MLP architecture outperforms the common U-Net architecture qualitatively and quantitatively by a large margin.

\subsection{Analysis of the OoD Score Computation}
\label{sec:scores_comparison}

As described in \cref{sec:related_ood_diffusion}, we propose to compute OoD scores from the directional error between a perturbation and the estimated score.
Moreover, we propose to aggregate OoD scores across multiple timesteps. A natural question that arises is which timesteps should be used for the aggregation. Related works either face similar questions~\cite{graham2023denoising,Zhang_2023_ICCV,goodier2023}, or work with a fixed noise magnitude~\cite{shin2023}. 

In the following, we analyze these two aspects, which shows that the proposed OoD score $\mathbf{E}_{\measuredangle}$ outperforms others and is more robust to the choice of timesteps. 

\fakeparagraph{Setup.} 
We compare three different OoD scores. For all scores, the input feature $\mathbf{x}_0$ is first perturbed with noise $\mathbf{\epsilon}$ to $\mathbf{x}_t$ via the forward diffusion process up to a timestep $t$. The OoD scores are computed from the model output $\mathbf{\epsilon}_\theta(\mathbf{x}_t, t)$ as:
\begin{itemize}
    \item $\text{MSE}_{\text{recon}}$: applies reverse diffusion process to obtain a reconstructed $\hat{\mathbf{x}}_0$, uses $\text{MSE}(\mathbf{x}_0, \hat{\mathbf{x}}_0)$ as OoD score~\cite{Wyatt2022}. Requires $t$ \emph{sequential} forward passes.
    \item $\text{MSE}_{\text{score}}$: uses $\text{MSE}(\mathbf{\epsilon}, \mathbf{\epsilon}_\theta(\mathbf{x}_t, t))$ as OoD score~\cite{shin2023}. Requires 1 forward pass.
    \item $\mathbf{E}_{\measuredangle,t}$: uses $\frac{-\mathbf{\epsilon}_\theta(\mathbf{x}_t, t)\cdot\epsilon}{\lVert \mathbf{\epsilon}_\theta(\mathbf{x}_t, t) \rVert \, \lVert \epsilon \rVert}$ as OoD score (Ours). Requires 1 forward pass.
\end{itemize}

To analyze the effects of different diffusion timesteps, we report quantitative results AP$_{t=1}$ for using only the first timestep and AP$_{\text{best}}$ for the best timestep.

\begin{figure}[!h]
\caption{\textbf{Analysis of OoD scores and diffusion timesteps.} In the table, we report the AP on three benchmarks for different ways to compute OoD scores and different diffusion timesteps. In the figure, we plot the AP$_t$ on RoadAnomaly for the different OoD scores depending on the timestep.
}
\begin{subfigure}[t]{0.60\textwidth}
    \centering
    \scriptsize
    \vspace{-2.12cm}
    \begin{tabular}[b]{l|cc|cc|cc}
    \toprule
                                & \multicolumn{2}{c|}{RoadAnomaly} & \multicolumn{2}{c|}{FS-Static} & \multicolumn{2}{c}{\ADEOoD{}} \\
                                & AP$_{t=1}$ & AP$_{\text{best}}$  & AP$_{t=1}$ & AP$_{\text{best}}$ & AP$_{t=1}$ & AP$_{\text{best}}$  \\
    \midrule
        $\text{MSE}_{\text{recon}}$      & 52.5 & 84.6 & 58.7 & 71.1 & 32.5 & 42.4    \\
        $\text{MSE}_{\text{score}}$      & 78.9 & 86.7 & 74.2 & 75.0 & 49.2 & 49.2   \\
        $\mathbf{E}_{\measuredangle}$ \tiny{(Ours)}    & \textbf{86.7} & \textbf{86.7} & \textbf{76.5} & \textbf{76.5}  & \textbf{50.4} & \textbf{50.4}  \\
    \bottomrule
    \end{tabular}
\label{tab:scores_comp}
\end{subfigure}
\begin{subfigure}[t]{0.33\textwidth}
    \includegraphics[width=\textwidth]{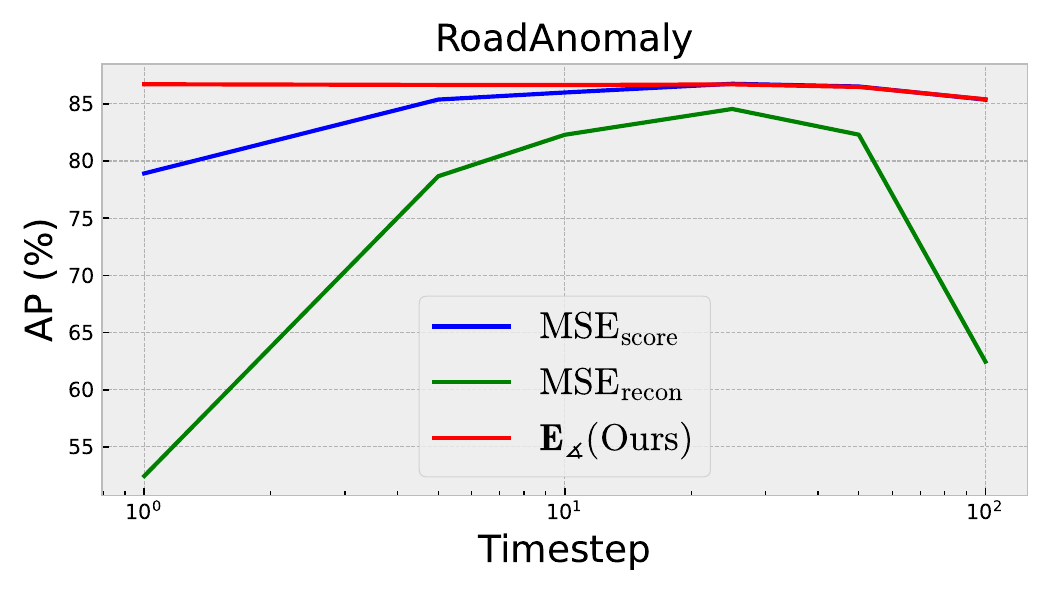}
\label{fig:scores_comp}
\end{subfigure}
\end{figure}

We report results in \cref{fig:scores_comp} and analyze them in the following.

\fakeparagraph{Comparison of OoD scores:} Overall, we observe that OoD scores $\text{MSE}_{\text{score}}$ and $\mathbf{E}_{\measuredangle}$, computed from the score estimate of the diffusion model, outperform the reconstruction-based OoD score $\text{MSE}_{\text{recon}}$. Further, we observe a consistently better performance of the proposed $\mathbf{E}_{\measuredangle}$ score compared to the $\text{MSE}_{\text{score}}$ score.

\fakeparagraph{Effect of the diffusion timestep:} We observe that the performance of the $\text{MSE}_{\text{score},t}$ score worsens sensibly for smaller timesteps, showing that it requires a carefully chosen timestep. The proposed $\mathbf{E}_{\measuredangle,t}$ score has a better and more consistent performance.

\subsection{Computational Costs}
\label{sec:costs}
We provide a comparison of computational costs on the RoadAnomaly benchmark for cDNP, RbA, and our approach (all on a RTX 3090 GPU):
\begin{table}[!h]
    \centering
    \scriptsize
    \begin{tabular}{l|c|c|c}
    \toprule
        Method & Time (s) & Mem. (GB) & AP (RA) \\
        \midrule
        cDNP Segm.-B & 0.170 & 4.58 & 85.6 \\
        RbA         & 0.124 & 1.30 & 78.5 \\
        Ours Segm.-B $|\mathcal{T}|=5$  & 0.125 & 1.87 & 86.1 \\
        Ours Segm.-B $|\mathcal{T}|=10$ & 0.126 & 1.87 & 87.9 \\
        Ours Segm.-B $|\mathcal{T}|=25$ & 0.322 & 1.87 & 89.1 \\
        
    \bottomrule
    \end{tabular}
    \label{tab:costs}
\end{table}

%
One can observe that RbA, whose runtime basically consists of a Mask2Former forward pass, is the cheapest. 
Our approach, while having a higher inference time than cDNP, it has a smaller GPU memory footprint. An advantage of our approach is that it offers the possibility to trade of runtime and performance via changing the number of diffusion timesteps that are aggregated in the OoD score computation. For example, aggregating five timesteps $\mathcal{T}=\{1,...,5\}$ enables a comparable runtime as previous methods but with better performance.

\section{Conclusion}
In this work we considered the problem of out-of-distribution detection in semantic segmentation data. We introduced, to the best of our knowledge, the first diffusion model based approach for this task. We explored design possibilities in terms of model architecture and OoD scoring functions, the best of which resulted in an approach which is competitive on the major driving-oriented OoD detection benchmarks, and achieves state of the art results on RoadAnomaly and SMIYC-Anomaly without outlier exposure and domain-specific priors.

Furthermore, we proposed a novel benchmark based on the ADE20k dataset, with the goal of assessing the quality of OoD detection approaches in settings with higher diversity, \ie beyond 19 in-distribution categories and driving scenarios. We provide evaluations of state-of-the-art approaches on the benchmark and show that our proposed method copes well with the increased diversity.

\clearpage
{
\subsubsection{Acknowledgements.} The research leading to these results is funded by the Deutsche Forschungsgemeinschaft (DFG, German Research Foundation) under the project numbers 401269959 and 417962828, and by the German Federal Ministry for Economic Affairs and Climate Action within the project “NXT GEN AI METHODS – Generative Methoden für Perzeption, Prädiktion und Planung". The authors would like to thank the consortium for the successful cooperation.
}

\bibliographystyle{splncs04}
\bibliography{main}

\newpage
\appendix
\section{Additional Information on the \ADEOoD{} Benchmark}

In this section, we provide additional information on the \ADEOoD{} benchmark. We provide details on the construction process and show samples from the benchmark alongside predictions from segmentation models. We analyze the predictions regarding the challenges of the benchmark. 

\subsection{Construction Details}

\subsubsection{Image Selection.}
As described in Sec.~5 of the main paper, the images for the benchmark were selected from the ADE20k~\cite{zhou2017scene} validation set and from OpenImages~\cite{OpenImages} according to the following criteria:
\begin{enumerate}
    \item Presence of clear and unambiguous in- and out-of-distribution entities only. 
    \item Presence of diverse in- and out-of-distribution categories, as well as varying indoor and outdoor settings. 
    \item Adequate scene complexity to match the original ADE20k training data.
    \item Adequate number of samples for statistically significant results.
\end{enumerate}

More specifically, we selected images by \begin{enumerate*}[label=\textbf{(\arabic*.)}] \item automatically selecting ADE20k images with large unlabeled regions, or manually selecting images from OpenImages (via the OpenImages web visualizer) that contain semantic classes that are not labeled in ADE20k, \item manually discarding images that do not fit to the described criteria (\eg images that contain ambiguous entities), \item validating the selected images by applying SOTA ADE20k segmentation models and manually inspecting their outputs.\end{enumerate*} 

In total we selected 84 images from OpenImages, and 27 from ADE20k. The complete benchmark data is provided on \url{https://ade-ood.github.io} and the source of each image is indicated by its filename. 

\subsubsection{Annotation Process.}
For the 27 images selected from ADE20k, we used the available ground truth segmentation annotations, using the ``unlabelled'' class mask as OoD (after checking for problematic cases as described above).

The annotation process for the 84 images selected from OpenImages was carried out in a semi-automatic fashion using the Segment Anything Model (SAM)~\cite{Kirillov_2023_ICCV}, a promptable object segmentation model trained on very large data collections, which we prompted using hand-selected points. For optimal mask coherence and quality, objects and object parts were segmented individually before merging all resulting masks. After obtaining binary masks for each image, we refined them using dilation and erosion operations. This allowed to remove small noise elements (points and holes) in the masks.

\subsubsection{Benchmark Size and Statistical Significance.}
To ensure that results on the benchmark are meaningful and reliable and that the number of samples in the benchmark is adequate, we tested its statistical robustness and compared it with that of other benchmarks. To this end, we computed results on 10 random subsets of the test data, each containing 90\% of the original samples. The mean and standard deviation of the obtained performances, computed for three benchmarks (RoadAnomaly, FS-Static, ADE20k-OoD), are reported in \cref{tab:ade_bootstrapped}.

\begin{table}[!h]
\centering
\setlength{\tabcolsep}{2pt}
\caption{\textbf{Mean and standard deviation results for 10-fold bootstrapped evaluation} on RoadAnomaly, Fishyscapes Static, and \ADEOoD{}.}
    \begin{tabular}{l|c|c|c|c|c|c}
        \toprule
        & \multicolumn{2}{c|}{RoadAnomaly} & \multicolumn{2}{c|}{FS-Static} & \multicolumn{2}{c}{\ADEOoD{}}  \\
         & AP & FPR & AP & FPR & AP & FPR \\
        \midrule
        \hline
        cDNP & 85.2$\pm$0.9 & 13.2$\pm$1.7 & 62.8$\pm$4.0 & 18.9$\pm$1.7 & 61.6$\pm$1.6 & 37.9$\pm$1.0 \\
        RbA  & 78.3$\pm$1.0 & 12.2$\pm$1.1 & 59.8$\pm$3.2 & 17.0$\pm$3.7 & 66.8$\pm$1.7 & 82.6$\pm$1.8 \\
        Ours & 89.2$\pm$0.7 & 9.1$\pm$0.7 & 74.8$\pm$2.7 & 6.1$\pm$0.6 & 62.9$\pm$2.1 & 35.9$\pm$0.8 \\
        \bottomrule
    \end{tabular}
    \label{tab:ade_bootstrapped}
\end{table}

The standard deviations are in a similar range for all benchmarks, and are generally low, indicating that \ADEOoD{} has a similar level of stability and statistical significance as other benchmarks.

\newpage
\subsection{Examples and Analysis of Segmentation Model Predictions}
\label{sec:ade_examples}
\subsubsection{Examples.} 
\begin{figure*}[!h]
\centering
\setlength{\tabcolsep}{2pt}
    \begin{minipage}[b]{0.12\textwidth}
      \raisebox{-1\baselineskip}{\includegraphics[width=\textwidth]{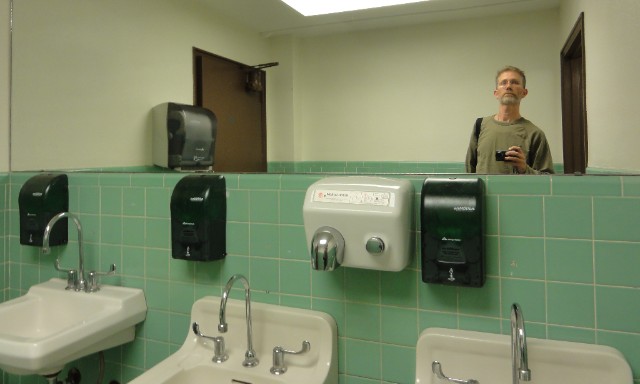}}
    \end{minipage}
    \begin{minipage}[b]{0.12\textwidth}
      \raisebox{-1\baselineskip}{\includegraphics[width=\textwidth]{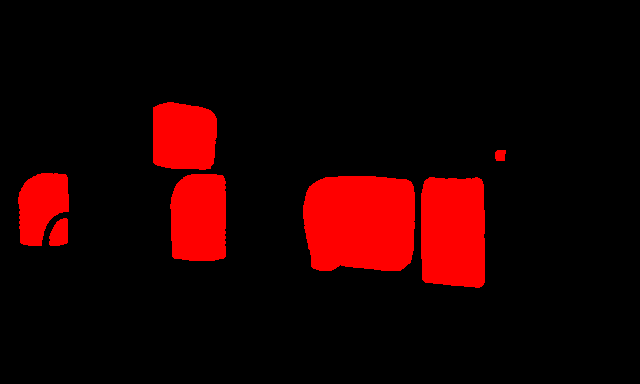}}
    \end{minipage}
    \begin{minipage}[b]{0.15\textwidth}
    \tiny
    \begin{tabular}{l}
      \SETR{wall, towel,} \\
      \SETR{mirror} \\
      \MF{windowpane,} \\
      \MF{wall, person}
    \end{tabular}
    \end{minipage}
    \begin{minipage}[b]{0.12\textwidth}
      \raisebox{-1\baselineskip}{\includegraphics[width=\textwidth]{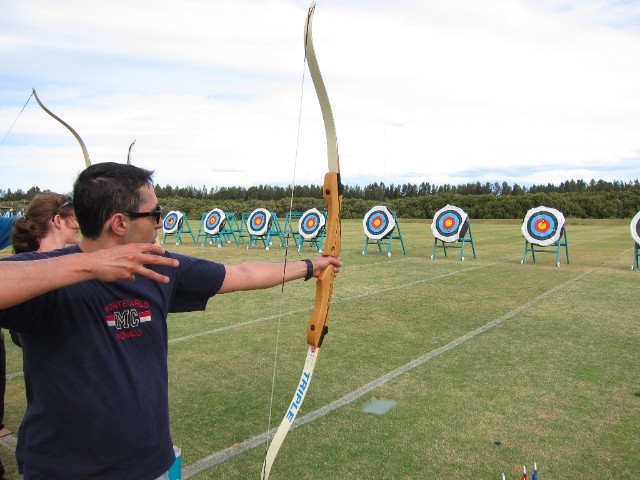}}
    \end{minipage}
    \begin{minipage}[b]{0.12\textwidth}
      \raisebox{-1\baselineskip}{\includegraphics[width=\textwidth]{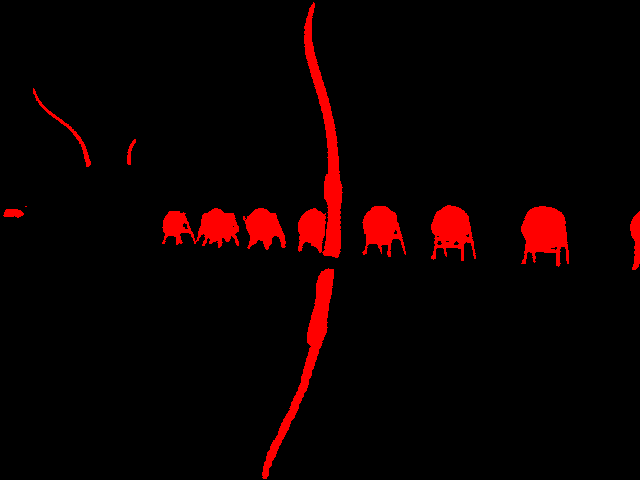}}
    \end{minipage}
    \begin{minipage}[b]{0.15\textwidth}
    \tiny
    \begin{tabular}{l}
      \SETR{grass, sky,} \\
      \SETR{fence} \\
      \MF{grass, person,} \\
      \MF{sky}
    \end{tabular}
    \end{minipage}
\vspace{13px}    
    
    \begin{minipage}[b]{0.12\textwidth}
      \raisebox{-1\baselineskip}{\includegraphics[width=\textwidth]{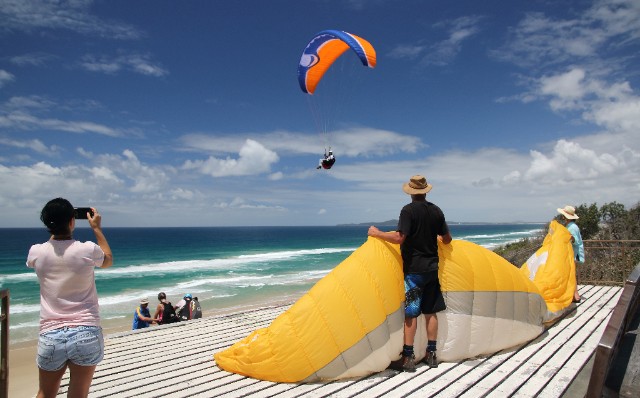}}
    \end{minipage}
    \begin{minipage}[b]{0.12\textwidth}
      \raisebox{-1\baselineskip}{\includegraphics[width=\textwidth]{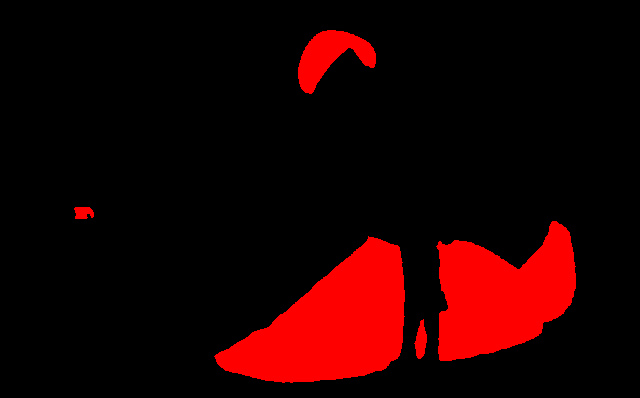}}
    \end{minipage}
    \begin{minipage}[b]{0.15\textwidth}
    \tiny
    \begin{tabular}{l}
      \SETR{tent, flag,} \\
      \SETR{person} \\
      \MF{bed, sky,} \\
      \MF{person}
    \end{tabular}
    \end{minipage}
    \begin{minipage}[b]{0.12\textwidth}
      \raisebox{-1\baselineskip}{\includegraphics[width=\textwidth]{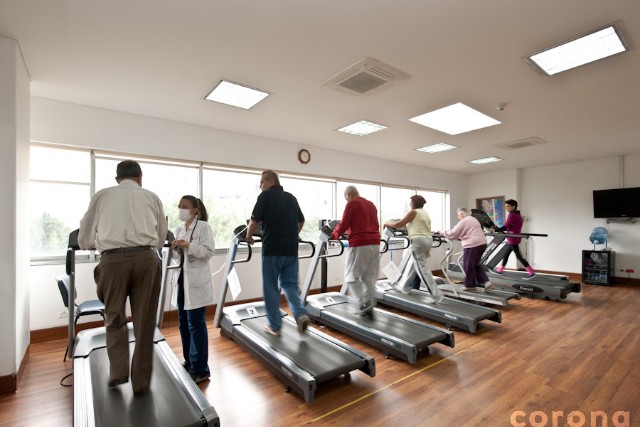}}
    \end{minipage}
    \begin{minipage}[b]{0.12\textwidth}
      \raisebox{-1\baselineskip}{\includegraphics[width=\textwidth]{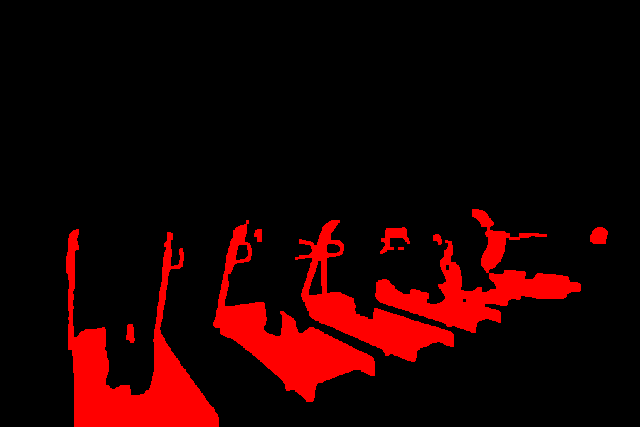}}
    \end{minipage}
    \begin{minipage}[b]{0.15\textwidth}
    \tiny
    \begin{tabular}{l}
      \SETR{floor, wall} \\
      \SETR{grandstand} \\
      \MF{floor, wall} \\
      \MF{windowpane}
    \end{tabular}
    \end{minipage}
\vspace{13px}

    \begin{minipage}[b]{0.12\textwidth}
      \raisebox{-1\baselineskip}{\includegraphics[width=\textwidth]{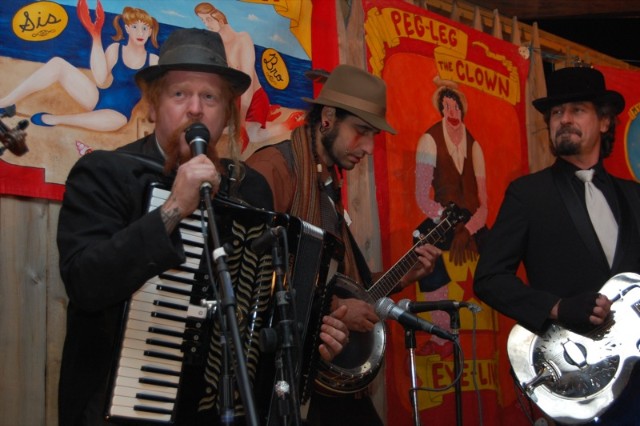}}
    \end{minipage}
    \begin{minipage}[b]{0.12\textwidth}
      \raisebox{-1\baselineskip}{\includegraphics[width=\textwidth]{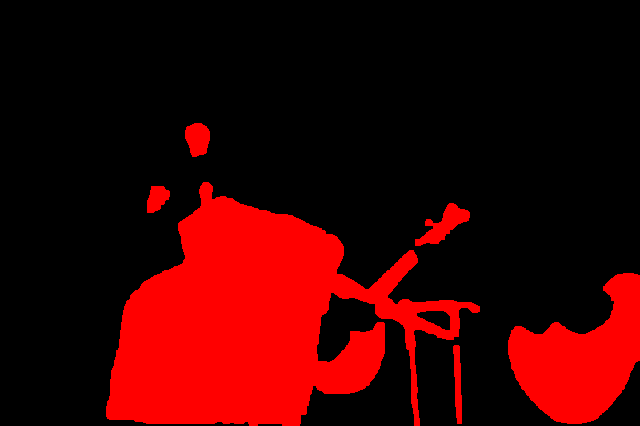}}
    \end{minipage}
    \begin{minipage}[b]{0.15\textwidth}
    \tiny
    \begin{tabular}{l}
      \SETR{person, wall,} \\
      \SETR{floor} \\
      \MF{person, curtain,} \\
      \MF{table}
    \end{tabular}
    \end{minipage}
    \begin{minipage}[b]{0.12\textwidth}
      \raisebox{-1\baselineskip}{\includegraphics[width=\textwidth]{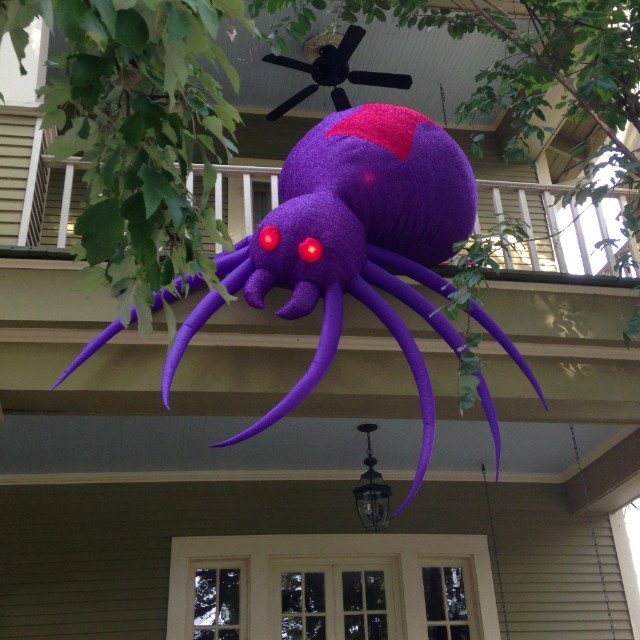}}
    \end{minipage}
    \begin{minipage}[b]{0.12\textwidth}
      \raisebox{-1\baselineskip}{\includegraphics[width=\textwidth]{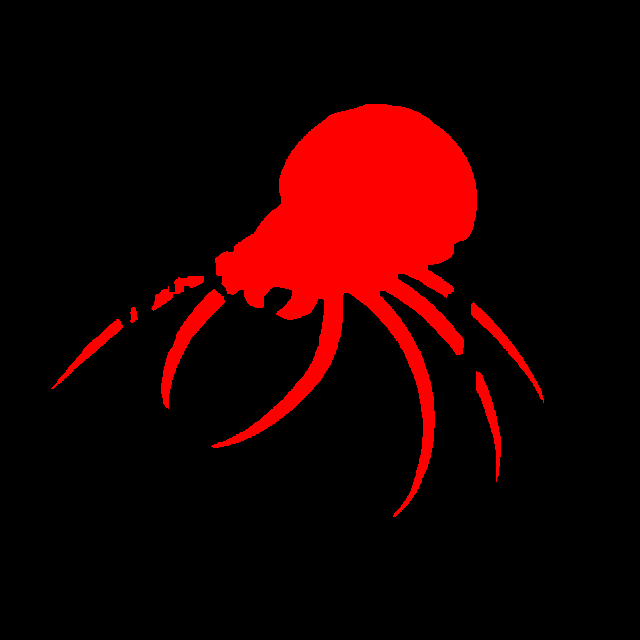}}
    \end{minipage}
    \begin{minipage}[b]{0.15\textwidth}
    \tiny
    \begin{tabular}{l}
      \SETR{building,} \\
      \SETR{sculpture} \\
      \MF{tree, building} \\
      \MF{}
    \end{tabular}
    \end{minipage}
\vspace{13px}
    
    \begin{minipage}[b]{0.12\textwidth}
      \raisebox{-1\baselineskip}{\includegraphics[width=\textwidth]{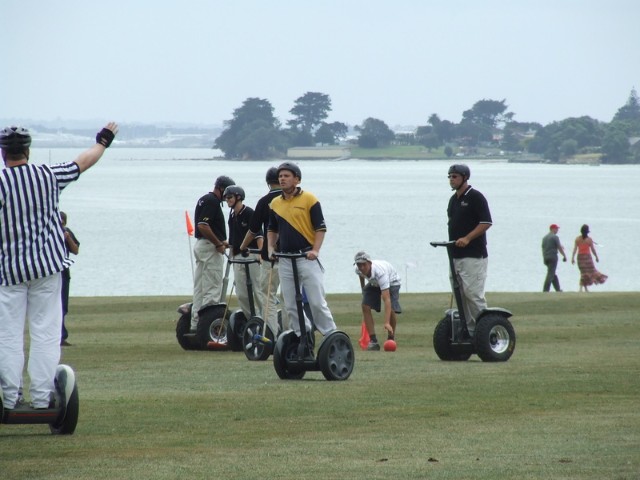}}
    \end{minipage}
    \begin{minipage}[b]{0.12\textwidth}
      \raisebox{-1\baselineskip}{\includegraphics[width=\textwidth]{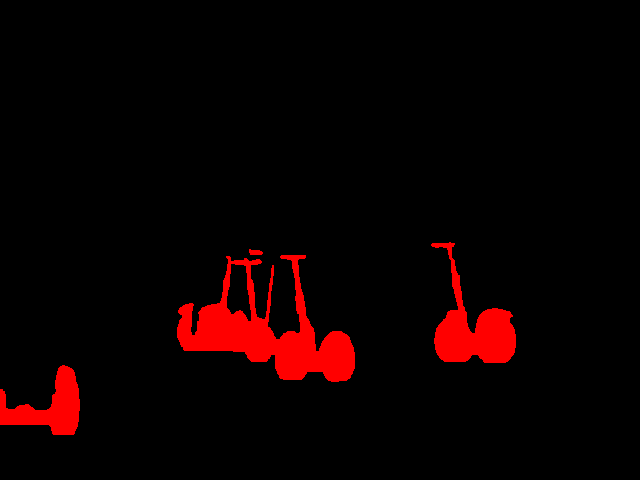}}
    \end{minipage}
    \begin{minipage}[b]{0.15\textwidth}
    \tiny
    \begin{tabular}{l}
      \SETR{grass, person,} \\
      \SETR{minibike} \\
      \MF{person, grass,} \\
      \MF{earth}
    \end{tabular}
    \end{minipage}
    \begin{minipage}[b]{0.12\textwidth}
      \raisebox{-1\baselineskip}{\includegraphics[width=\textwidth]{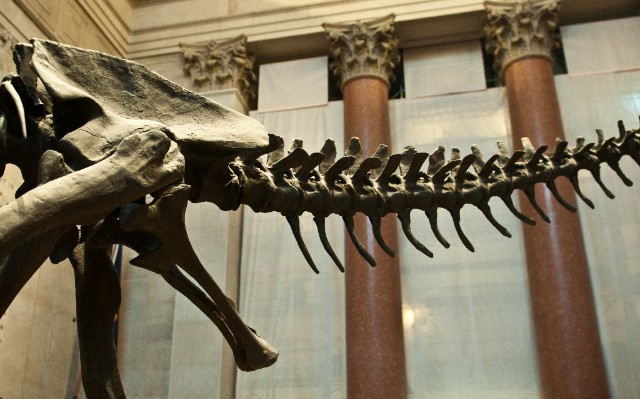}}
    \end{minipage}
    \begin{minipage}[b]{0.12\textwidth}
      \raisebox{-1\baselineskip}{\includegraphics[width=\textwidth]{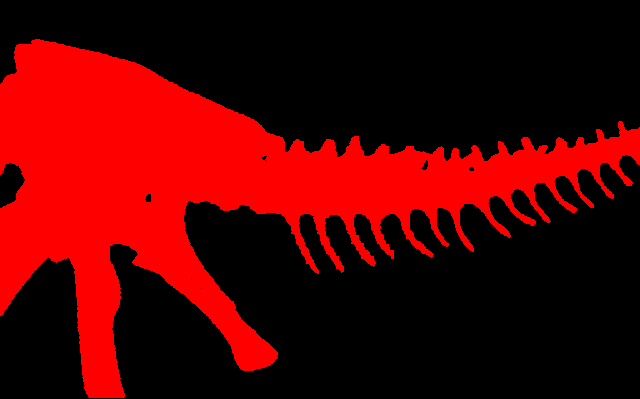}}
    \end{minipage}
    \begin{minipage}[b]{0.15\textwidth}
    \tiny
    \begin{tabular}{l}
      \SETR{sculpture, } \\
      \SETR{wall, animal} \\
      \MF{tree, wall,} \\
      \MF{person}
    \end{tabular}
    \end{minipage}
\vspace{13px}
    
    \begin{minipage}[b]{0.12\textwidth}
      \raisebox{-1\baselineskip}{\includegraphics[width=\textwidth]{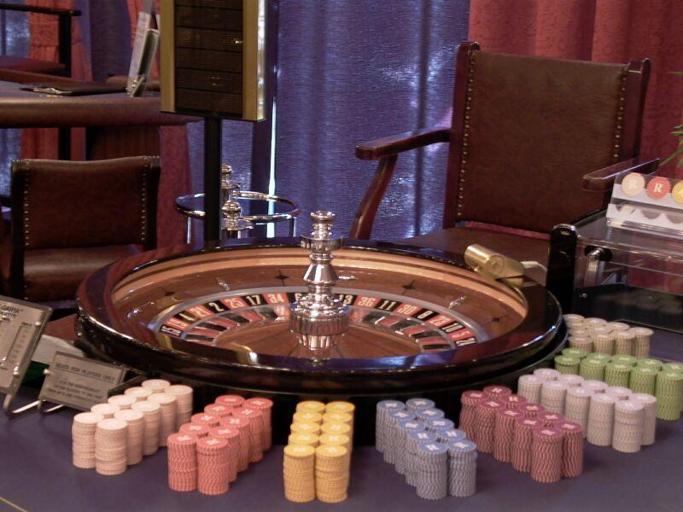}}
    \end{minipage}
    \begin{minipage}[b]{0.12\textwidth}
      \raisebox{-1\baselineskip}{\includegraphics[width=\textwidth]{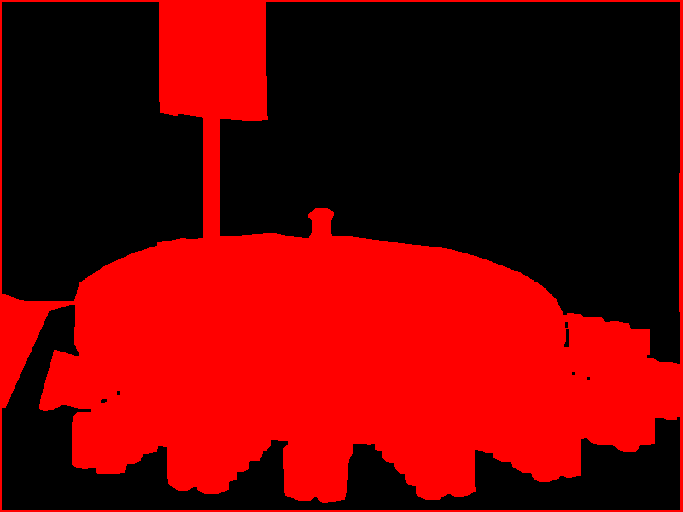}}
    \end{minipage}
    \begin{minipage}[b]{0.15\textwidth}
    \tiny
    \begin{tabular}{l}
      \SETR{tray, floor,} \\
      \SETR{wall} \\
      \MF{floor, table,} \\
      \MF{wall}
    \end{tabular}
    \end{minipage}
    \begin{minipage}[b]{0.12\textwidth}
      \raisebox{-1\baselineskip}{\includegraphics[width=\textwidth]{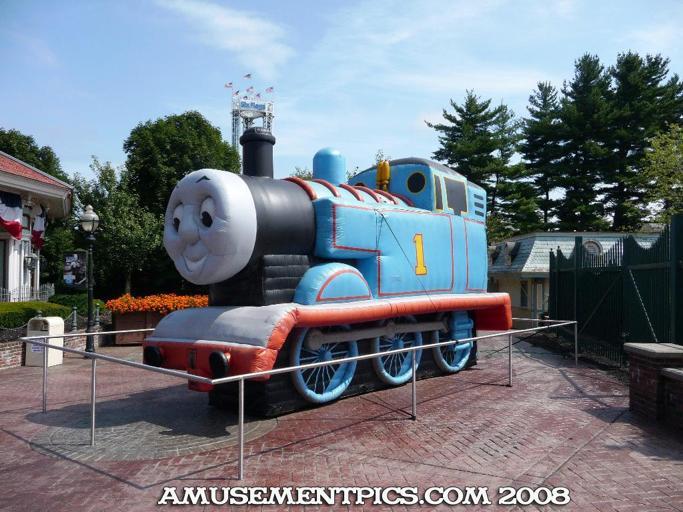}}
    \end{minipage}
    \begin{minipage}[b]{0.12\textwidth}
      \raisebox{-1\baselineskip}{\includegraphics[width=\textwidth]{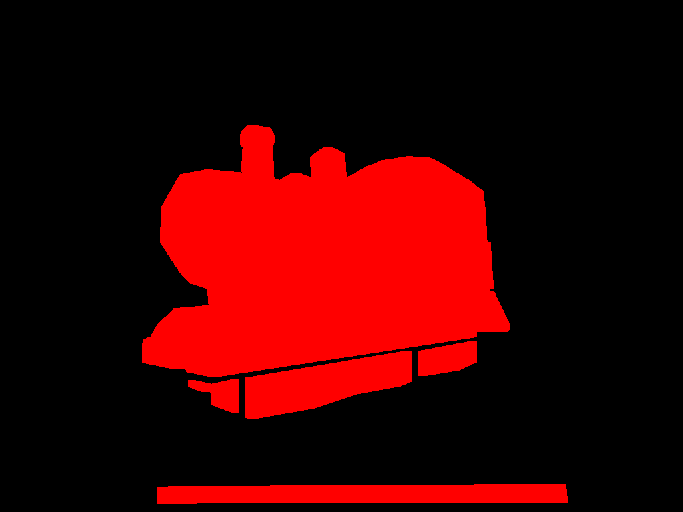}}
    \end{minipage}
    \begin{minipage}[b]{0.15\textwidth}
    \tiny
    \begin{tabular}{l}
      \SETR{truck, tree,} \\
      \SETR{building} \\
      \MF{building, earth,} \\
      \MF{road}
    \end{tabular}
    \end{minipage}
\vspace{13px}
    
    \begin{minipage}[b]{0.12\textwidth}
      \raisebox{-1\baselineskip}{\includegraphics[width=\textwidth]{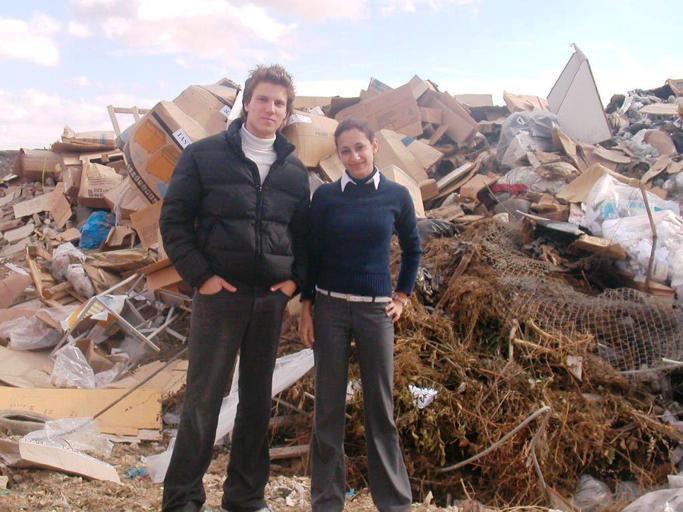}}
    \end{minipage}
    \begin{minipage}[b]{0.12\textwidth}
      \raisebox{-1\baselineskip}{\includegraphics[width=\textwidth]{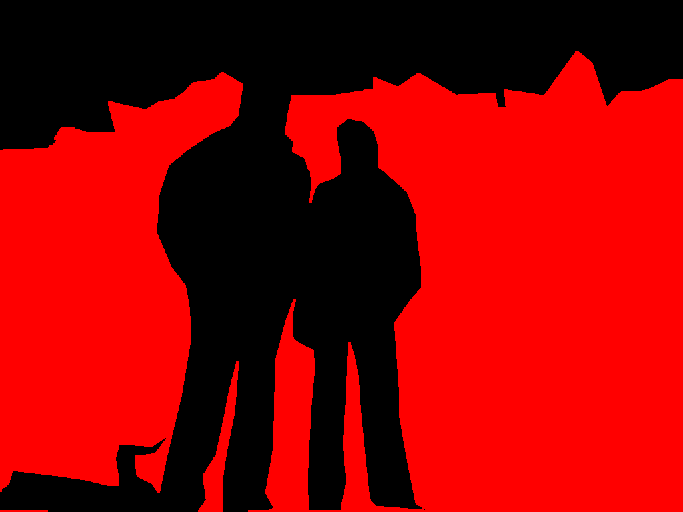}}
    \end{minipage}
    \begin{minipage}[b]{0.15\textwidth}
    \tiny
    \begin{tabular}{l}
      \SETR{earth, building,} \\
      \SETR{person} \\
      \MF{earth, door,} \\
      \MF{building}
    \end{tabular}
    \end{minipage}
    \begin{minipage}[b]{0.12\textwidth}
      \raisebox{-1\baselineskip}{\includegraphics[width=\textwidth]{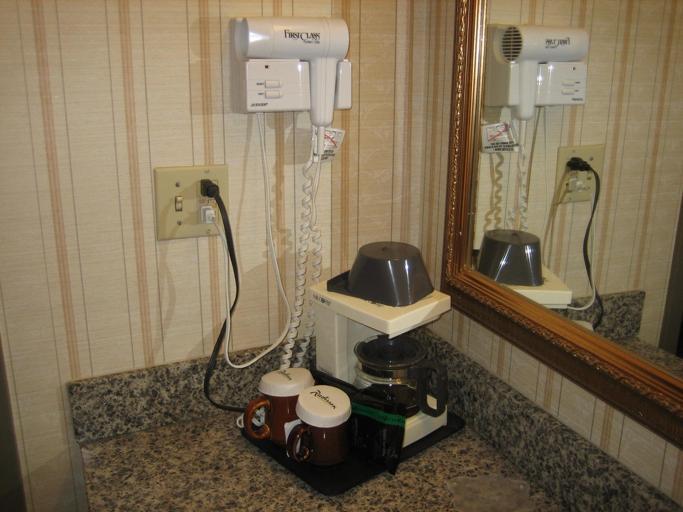}}
    \end{minipage}
    \begin{minipage}[b]{0.12\textwidth}
      \raisebox{-1\baselineskip}{\includegraphics[width=\textwidth]{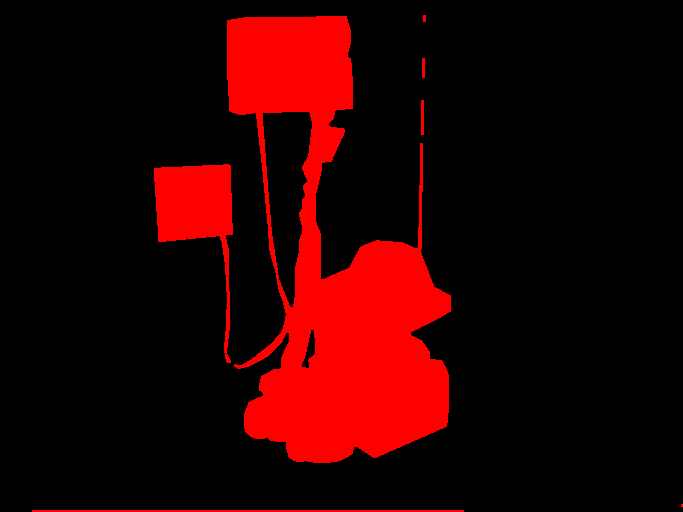}}
    \end{minipage}
    \begin{minipage}[b]{0.15\textwidth}
    \tiny
    \begin{tabular}{l}
      \SETR{wall, box,} \\
      \SETR{bottle} \\
      \MF{wall, person,} \\
      \MF{table}
    \end{tabular}
    \end{minipage}
\vspace{13px}

\caption{\textbf{Samples from the proposed ADE20k-OoD Benchmark.} We show sample images with their respective annotation masks (red=OoD). Additionally, we indicate the the top-3 classes that two state-of-the-art models ({\SETR{SETR}} and {\MF{Mask2Former}}) predict for the OoD regions. The first four rows contain examples from the OpenImages dataset, while the samples in the last two rows are sourced from ADE20k. The images featured in the benchmark depict both indoor and outdoor scenes and feature diverse out-of-distribution objects of different semantic classes.}
\label{fig:ade_samples}

\end{figure*}

A collection of samples (images and ground truth OoD masks) from the \ADEOoD{} benchmark is shown in \cref{fig:ade_samples}.

\subsubsection{Analysis of Segmentation Model Predictions.}Additionally, \cref{fig:ade_samples} indicates the most prominent predicted classes for the OoD regions obtained from SETR and Mask2Former models.

The predictions give us useful information about how the trained networks handle unknown objects, and therefore about the challenges of the current models on the proposed benchmark.
We can make the following observations:
\begin{itemize}
    \item Several times the OoD objects are classified as a spatially neighboring and frequent class, though this is not visually related to the object (e.g. \texttt{wall}, \texttt{person}, \texttt{building}). This is a form of extrapolation from the context that both models do, often with the same categories.
    \item The remaining predicted classes are visually similar to the OoD objects (e.g. \texttt{tent} for the parachute in the second row, \texttt{minibike} for the segways in the fourth row), or would likely feature in the surrounding environment (\texttt{towel} in the restroom in the first row, \texttt{fence} in the outdoor scene in the first row). These ``good guess'' predictions differ between the two models, confirming that the chosen objects are interesting for OoD detection.
\end{itemize}

\section{Diffusion Model Architecture}
\begin{figure}[!h]
    \centering
    \includegraphics[width=0.7\textwidth]{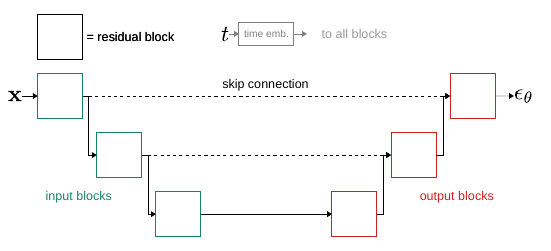}
    \caption{\textbf{Overview of the diffusion model architecture.} The architecture is composed of several residual blocks, split into input blocks and output blocks. In the usual U-Net fashion, skip-connections are present between the input and output blocks. Each block receives the timestep embedding as input.}
    \label{fig:arch}
\end{figure}

The architecture of the denoiser in our diffusion model, illustrated in \cref{fig:arch}, is based on the commonly used U-Net architecture, but replaces the convolutional layers with linear layers. The network is composed of consecutive sets of \emph{residual blocks}, structured into input blocks and output blocks. Each block receives features and timestep embedding as input. Skip connections forward input block features directly to the corresponding output block, like in a convolutional U-Net. In our experiments we use 6 input blocks and 6 output blocks.

\subsubsection{Residual Block Architecture.} 
\begin{figure}[!h]
    \centering
    \includegraphics[width=0.45\textwidth]{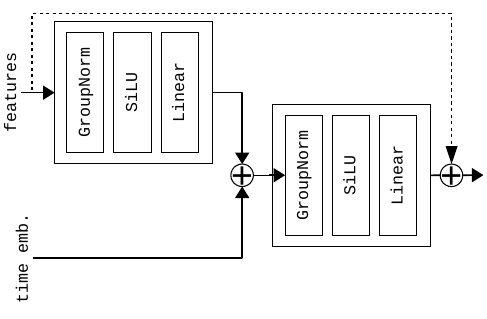}
    \caption{\textbf{Residual Block Architecture.} Features are processed individually, then summed with the timestep embedding,  and then processed by output layers. A residual connection spans the whole block, from input to output.}
    \label{fig:resblock}
\end{figure}

As shown in \cref{fig:resblock}, each residual block first processes incoming features with a sequence of a normalization layer (GroupNorm~\cite{wu2018group}), a non-linearity (SiLU~\cite{elfwing2018sigmoid}), and a linear layer. 
Then the timestep embedding is added and the result is processed by another sequence of a normalization layer, a non-linearity and a linear layer. %
The original input is summed to the output to form a residual connection. We use a constant hidden dimension throughout all linear layers.

\subsubsection{Lightweight Timestep Embedding.}
As common in diffusion models, the current diffusion timestep is supplied to the network in form of a timestep embedding. 
In order to reduce the computational costs, we compute the timestep embedding as follows:
\begin{itemize}
    \item Instead of using transformer-style periodic sinusoidal time embeddings, as done in DDPM and ViT works~\cite{ddpm,dosovitskiy2020image}, we directly use the scalar time embedding $t$.
    \item In each residual block, instead of further processing the timestep embedding as done for example in DDPM, we directly add the scalar time embedding to the incoming features (as shown in \cref{fig:resblock}).
\end{itemize}

With these timestep embedding simplifications, we observe on average a 17\% inference time speedup and no decrease in OoD detection performance.

\section{Qualitative Examples and Analysis}
\begin{figure*}[!h]
    \centering
    \begin{subfigure}[b]{0.18\textwidth}
        \centering
        \includegraphics[width=\textwidth]{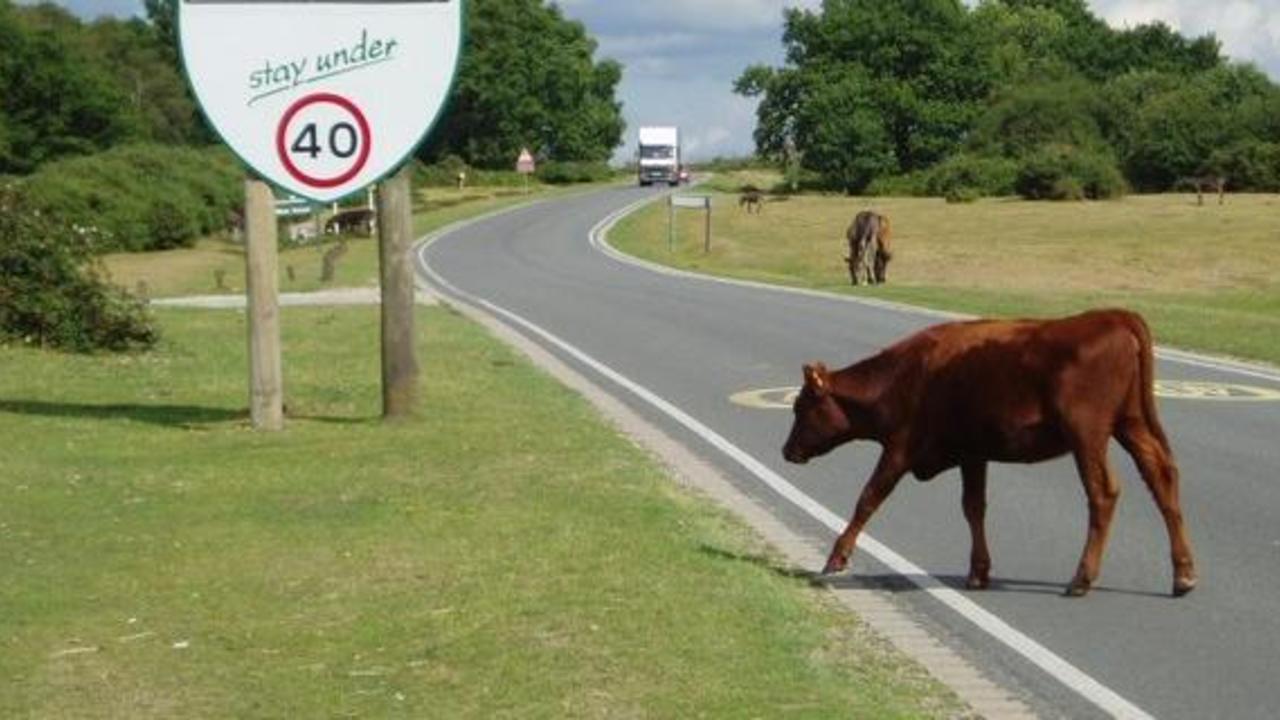}
    \end{subfigure}
    \begin{subfigure}[b]{0.18\textwidth}
        \centering
        \includegraphics[width=\textwidth]{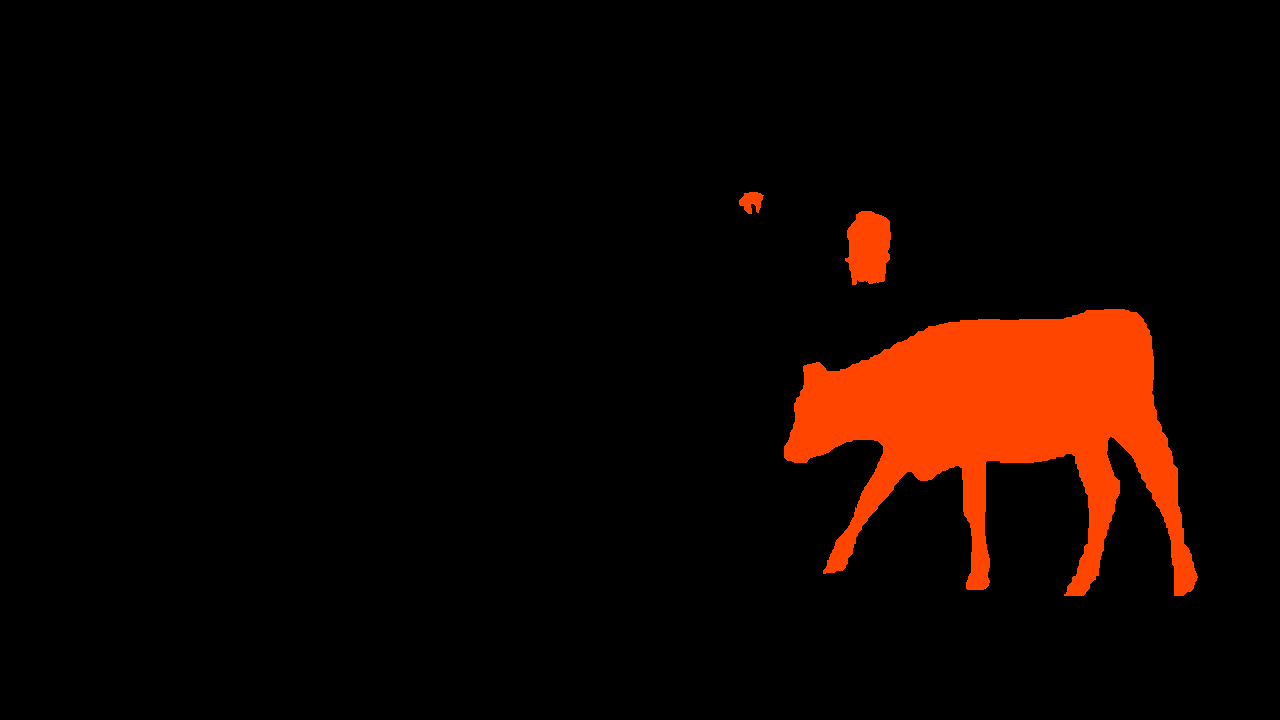}
    \end{subfigure}
    \begin{subfigure}[b]{0.18\textwidth}
        \centering
        \includegraphics[width=\textwidth]{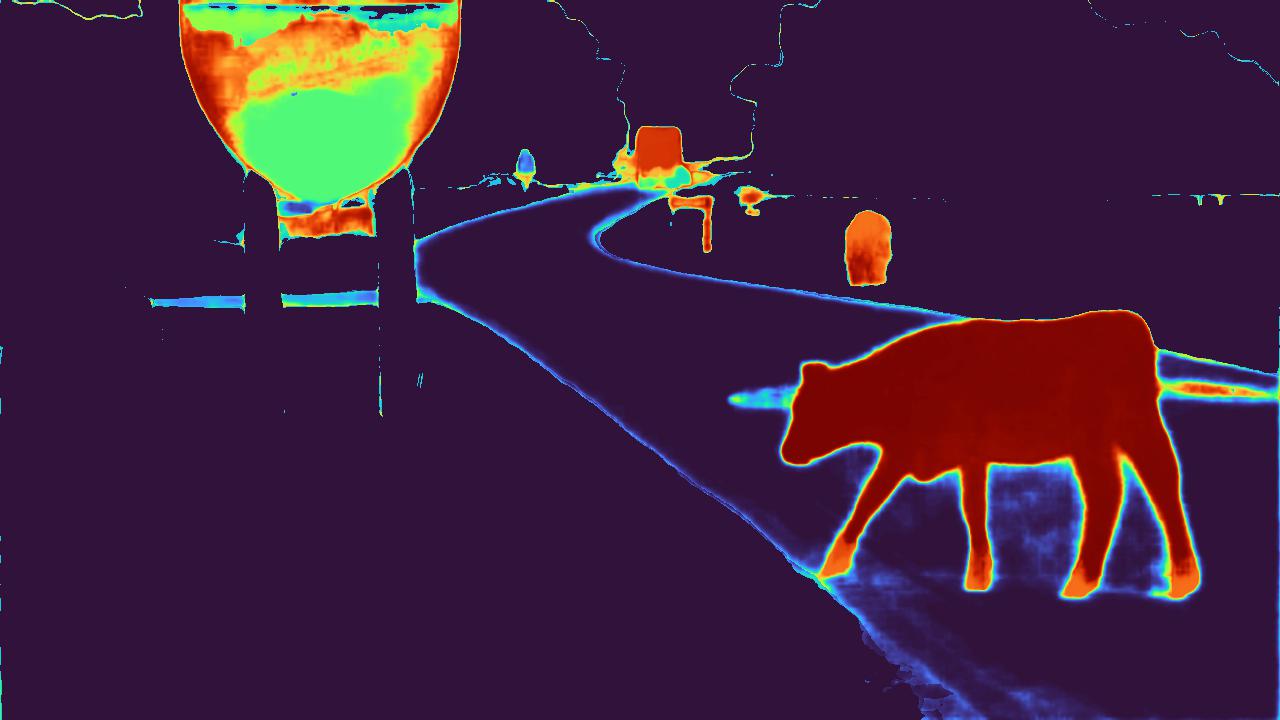}
    \end{subfigure}
    \begin{subfigure}[b]{0.18\textwidth}
        \centering
        \includegraphics[width=\textwidth]{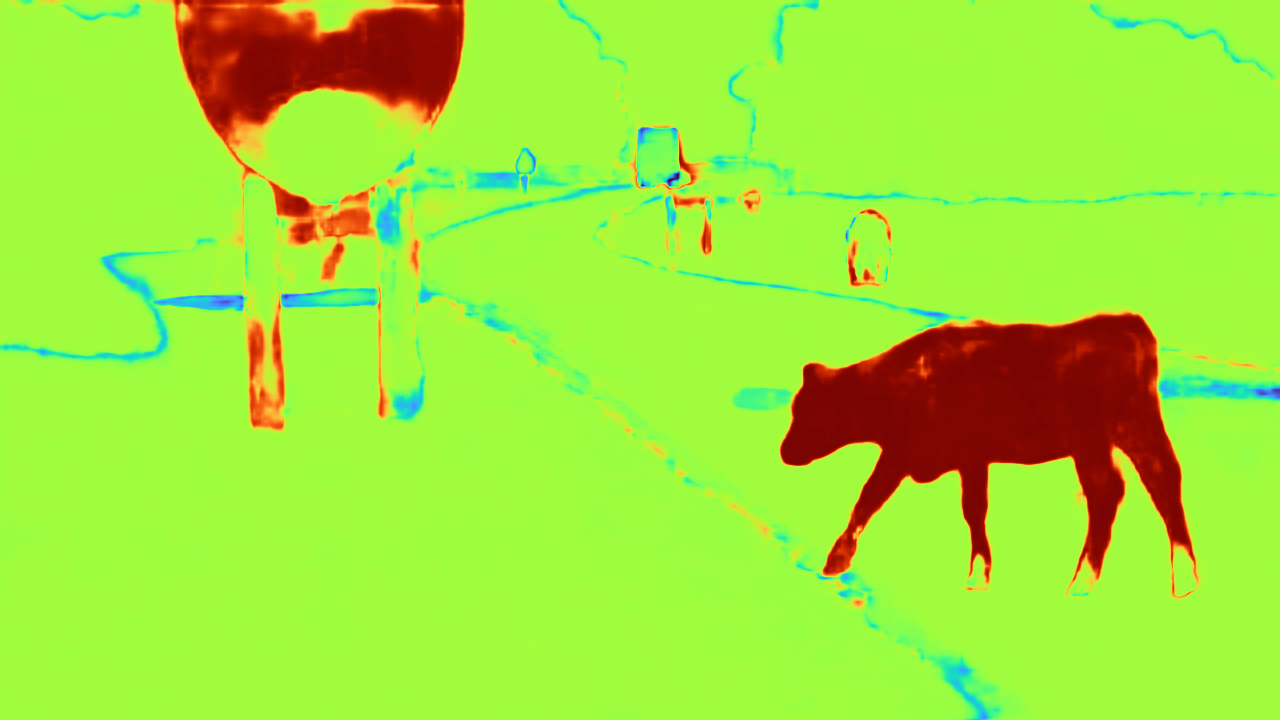}
    \end{subfigure}
    \begin{subfigure}[b]{0.18\textwidth}
        \centering
        \includegraphics[width=\textwidth]{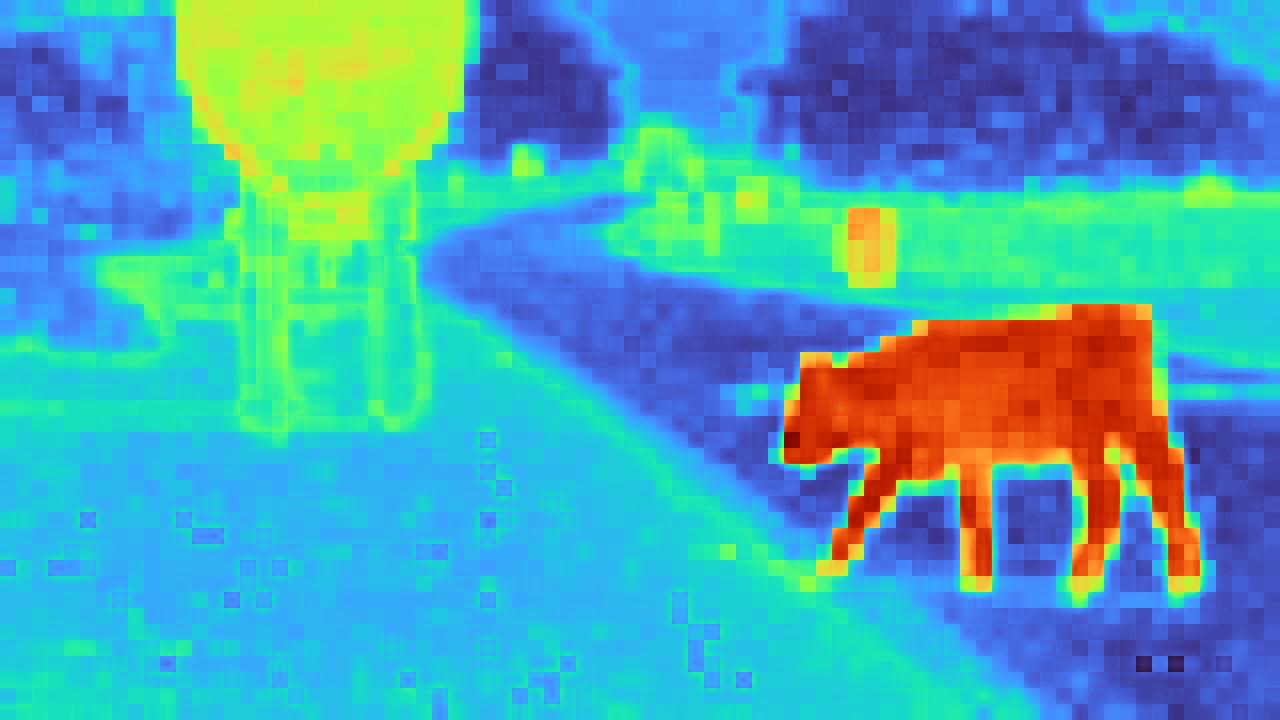}
    \end{subfigure}
    
    \begin{subfigure}[b]{0.18\textwidth}
        \centering
        \includegraphics[width=\textwidth]{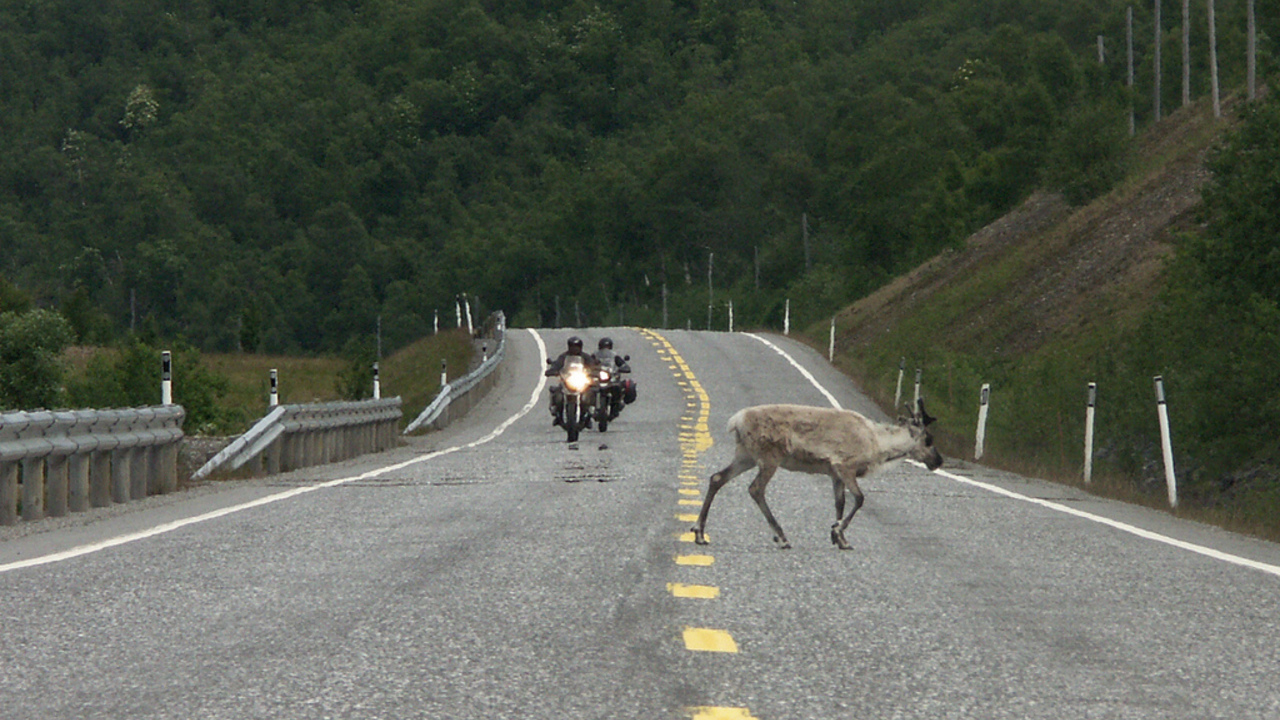}
    \end{subfigure}
    \begin{subfigure}[b]{0.18\textwidth}
        \centering
        \includegraphics[width=\textwidth]{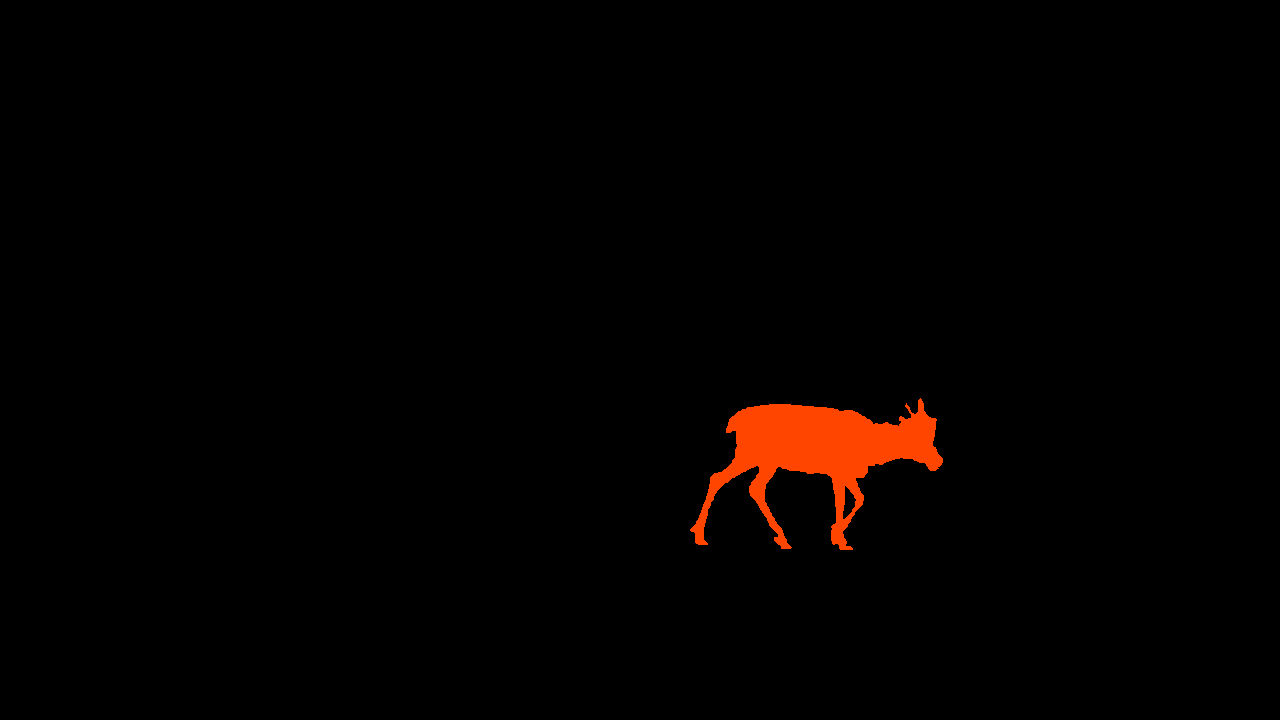}
    \end{subfigure}
    \begin{subfigure}[b]{0.18\textwidth}
        \centering
        \includegraphics[width=\textwidth]{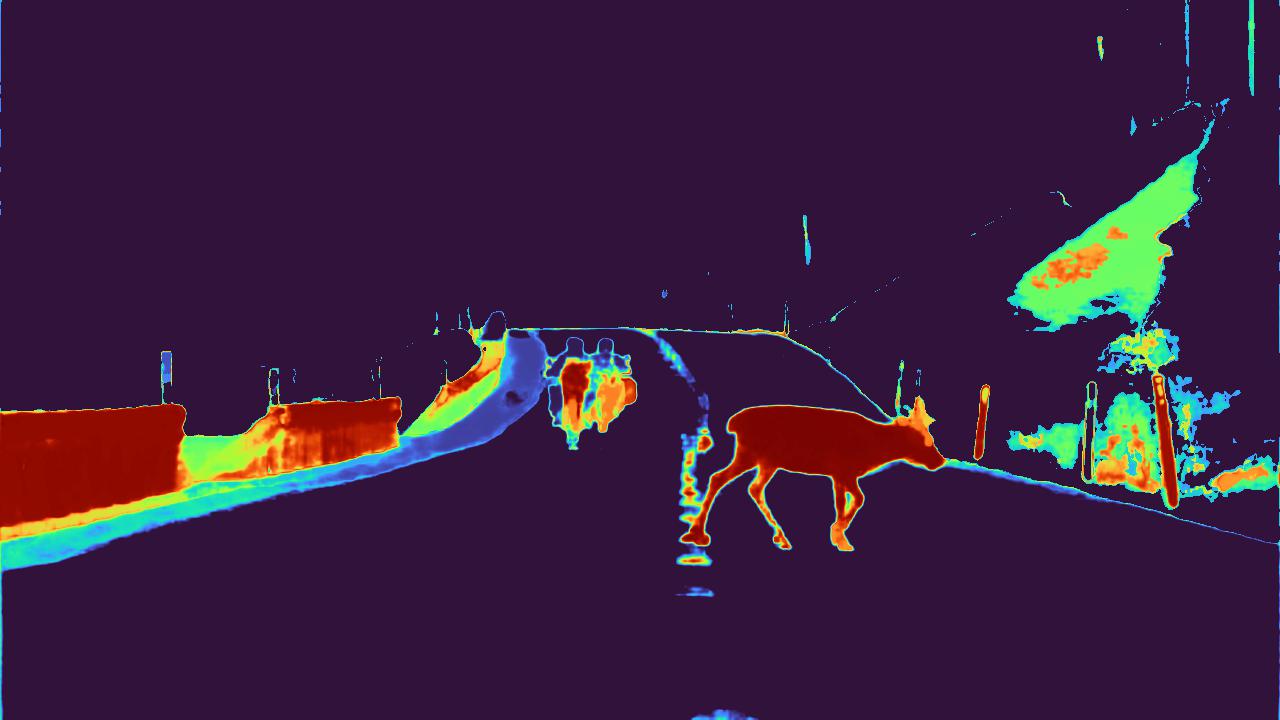}
    \end{subfigure}
    \begin{subfigure}[b]{0.18\textwidth}
        \centering
        \includegraphics[width=\textwidth]{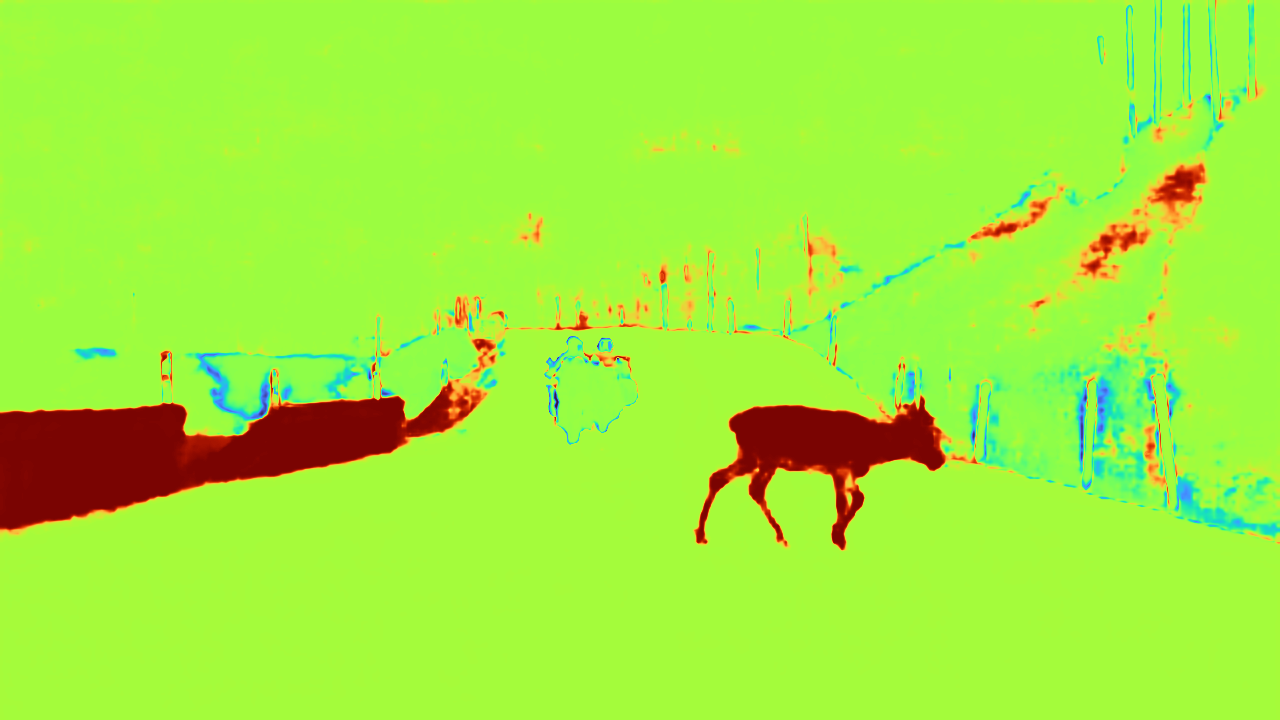}
    \end{subfigure}
    \begin{subfigure}[b]{0.18\textwidth}
        \centering
        \includegraphics[width=\textwidth]{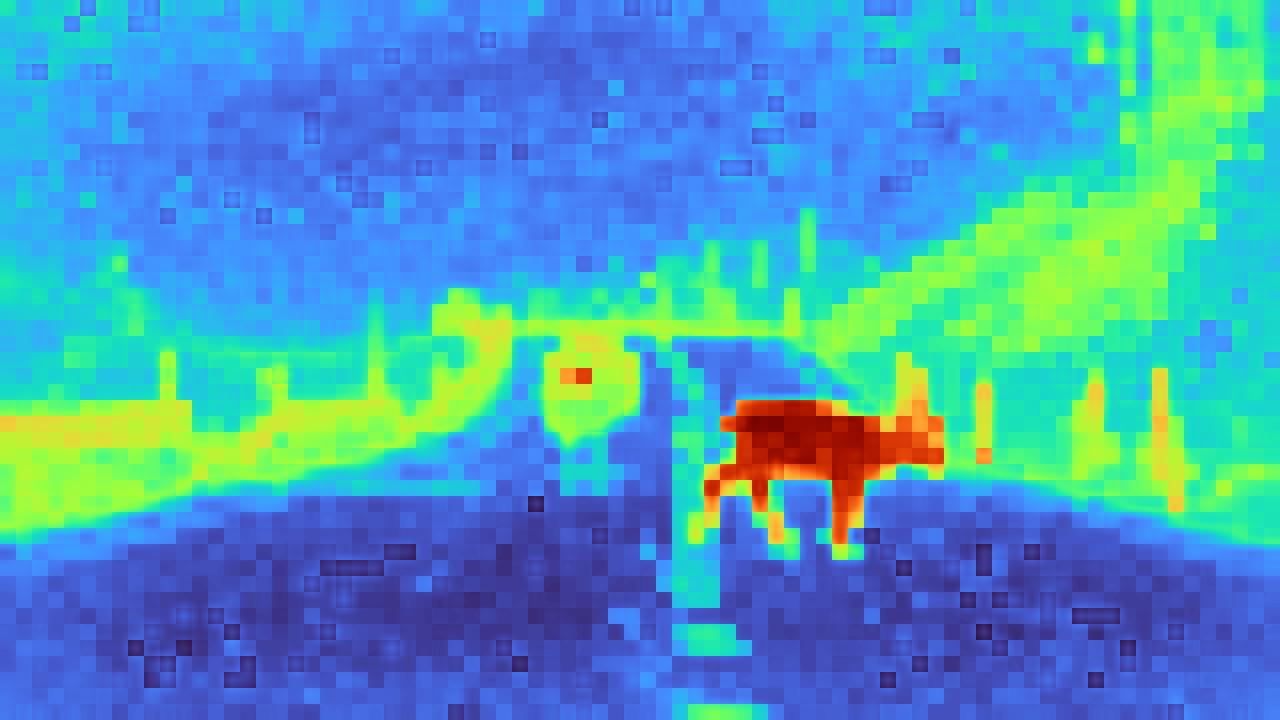}
    \end{subfigure}
    
    \begin{subfigure}[b]{0.18\textwidth}
        \centering
        \includegraphics[width=\textwidth]{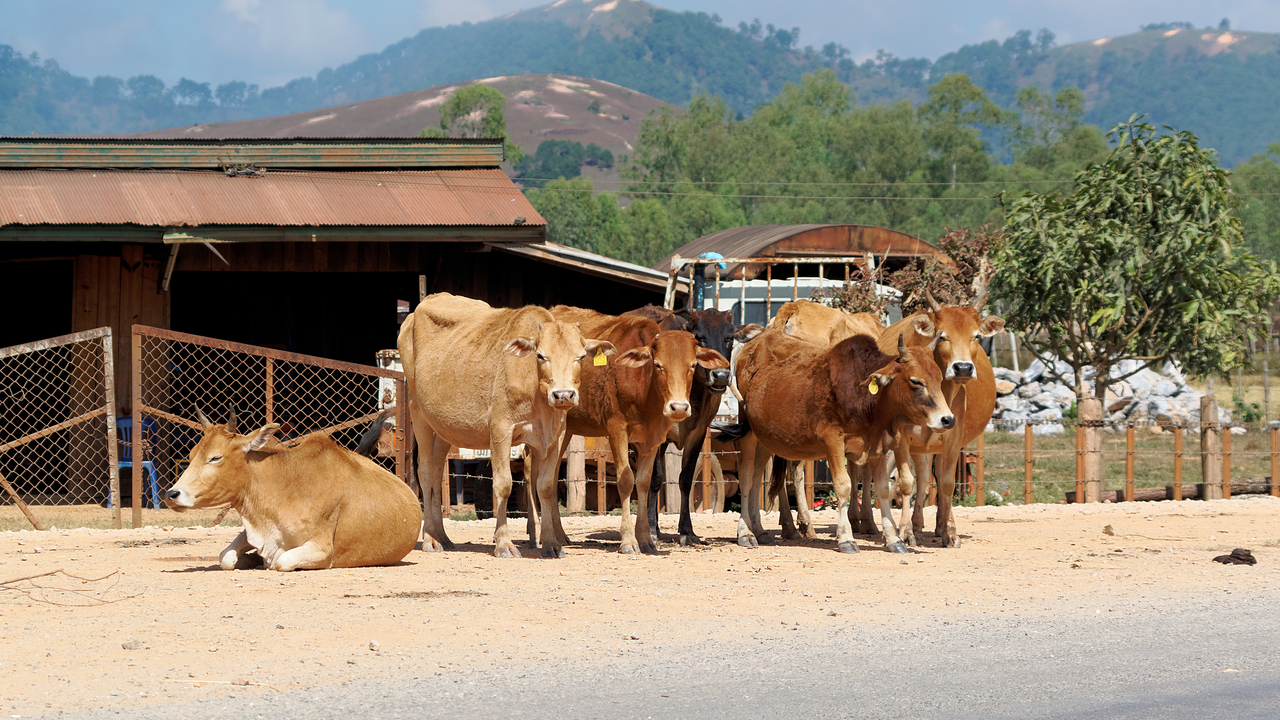}
    \end{subfigure}
    \begin{subfigure}[b]{0.18\textwidth}
        \centering
        \includegraphics[width=\textwidth]{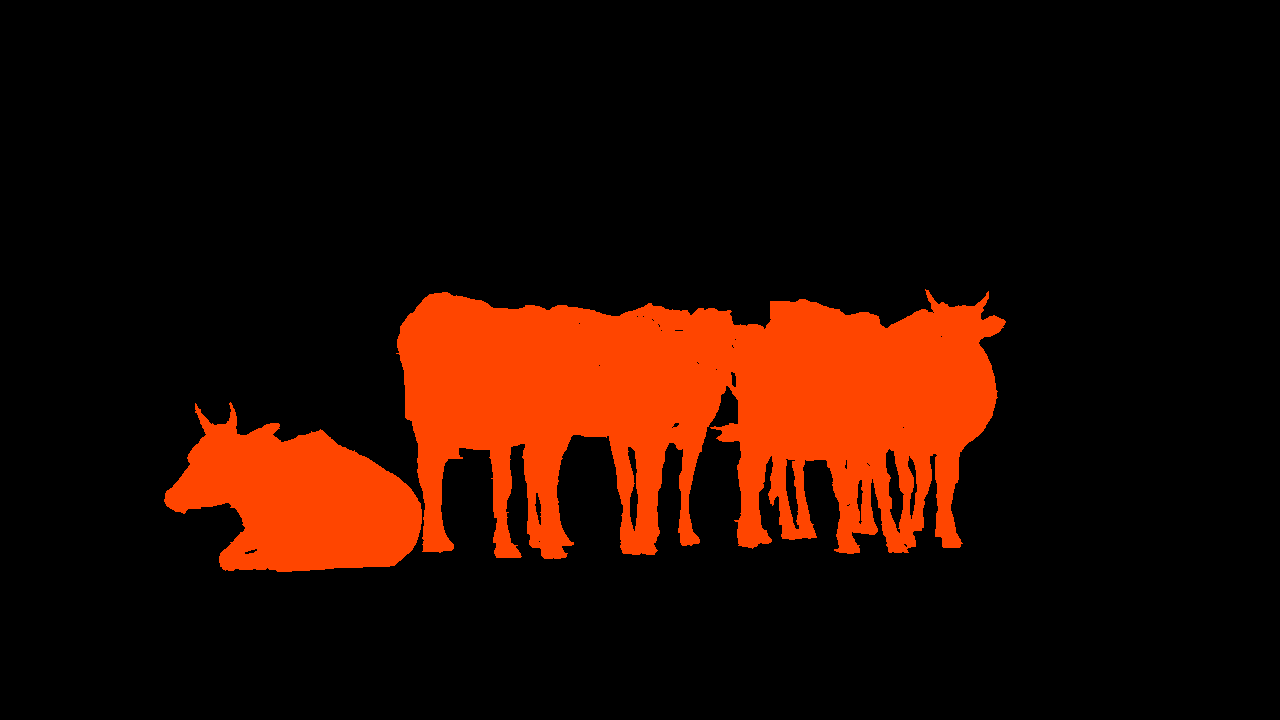}
    \end{subfigure}
    \begin{subfigure}[b]{0.18\textwidth}
        \centering
        \includegraphics[width=\textwidth]{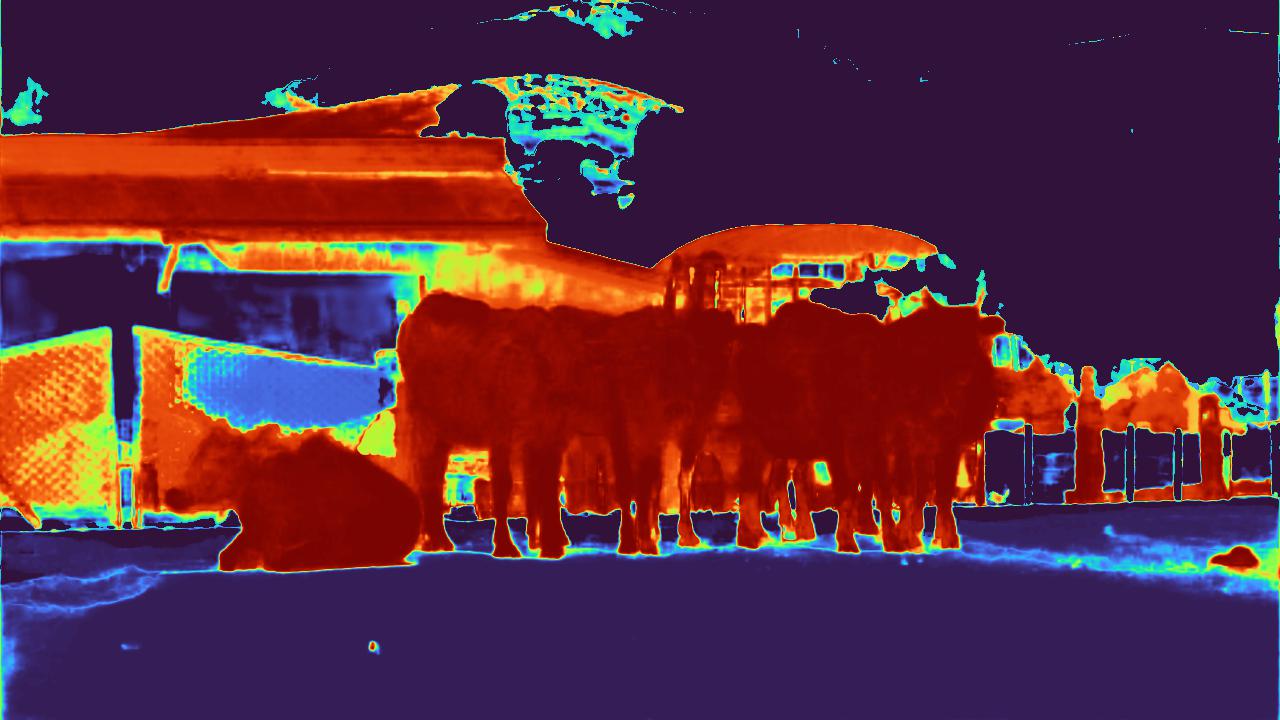}
    \end{subfigure}
    \begin{subfigure}[b]{0.18\textwidth}
        \centering
        \includegraphics[width=\textwidth]{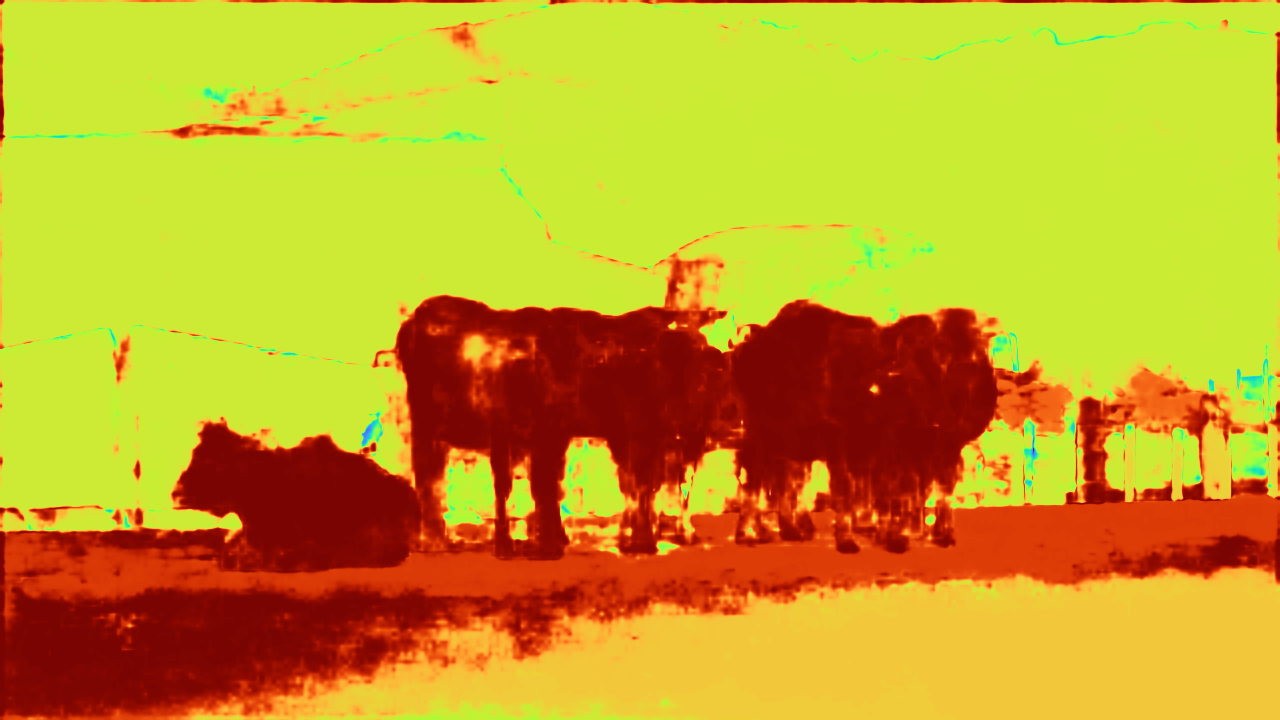}
    \end{subfigure}
    \begin{subfigure}[b]{0.18\textwidth}
        \centering
        \includegraphics[width=\textwidth]{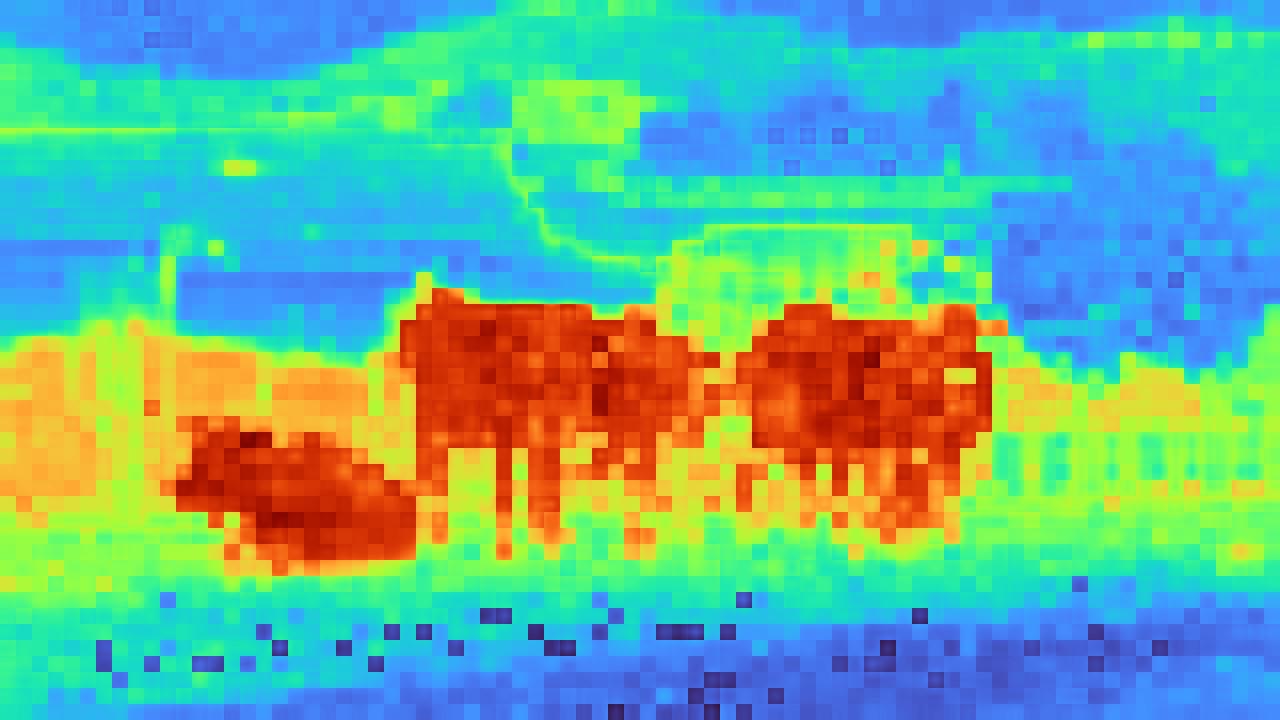}
    \end{subfigure}
    
    \begin{subfigure}[b]{0.18\textwidth}
        \centering
        \includegraphics[width=\textwidth]{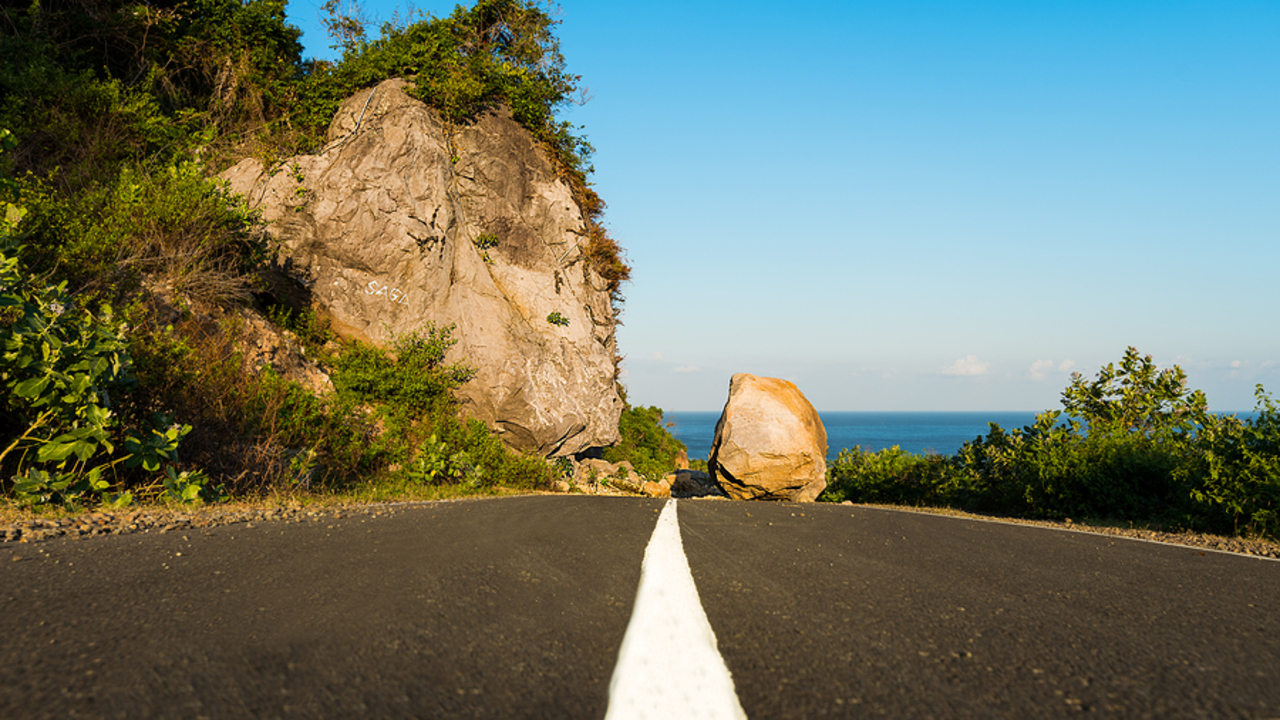}
    \end{subfigure}
    \begin{subfigure}[b]{0.18\textwidth}
        \centering
        \includegraphics[width=\textwidth]{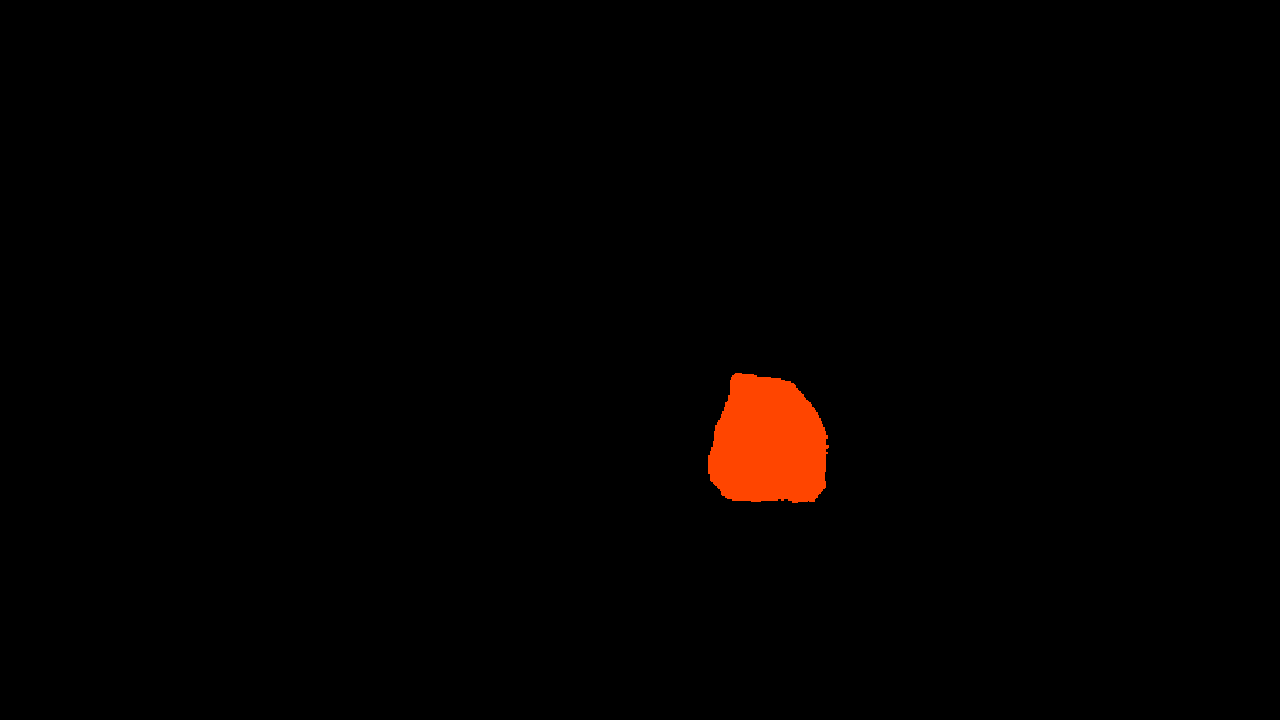}
    \end{subfigure}
    \begin{subfigure}[b]{0.18\textwidth}
        \centering
        \includegraphics[width=\textwidth]{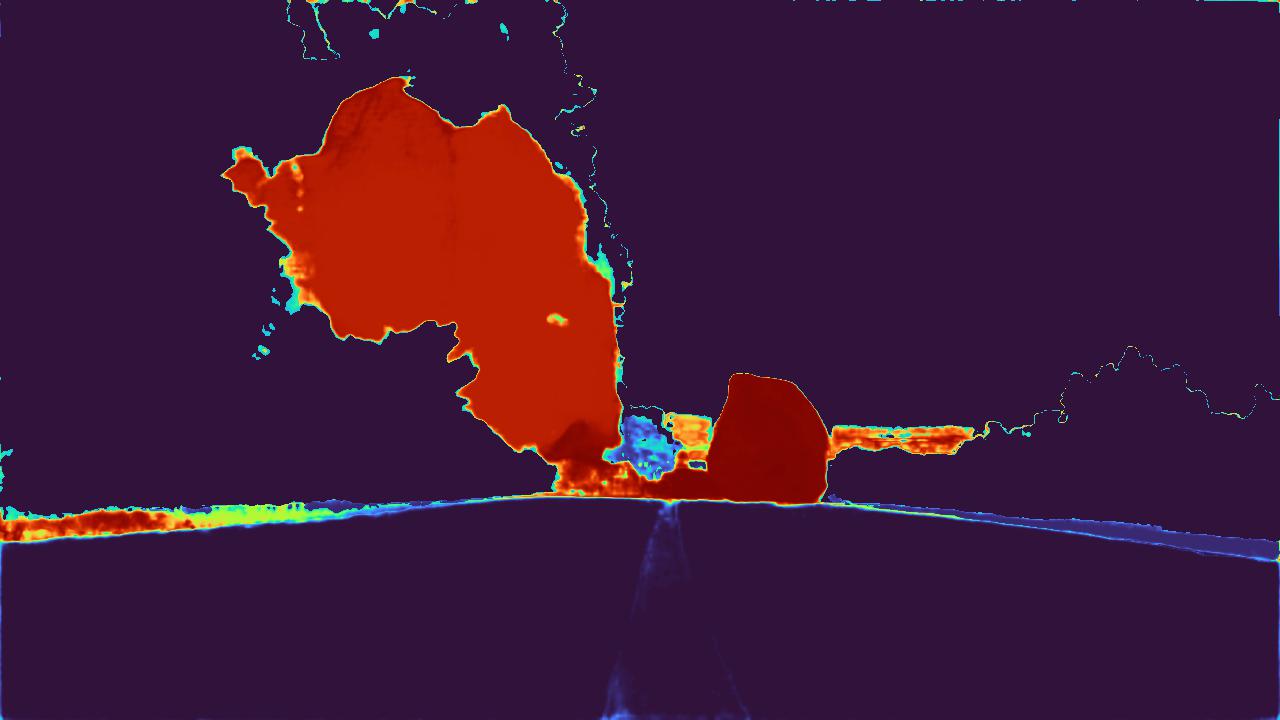}
    \end{subfigure}
    \begin{subfigure}[b]{0.18\textwidth}
        \centering
        \includegraphics[width=\textwidth]{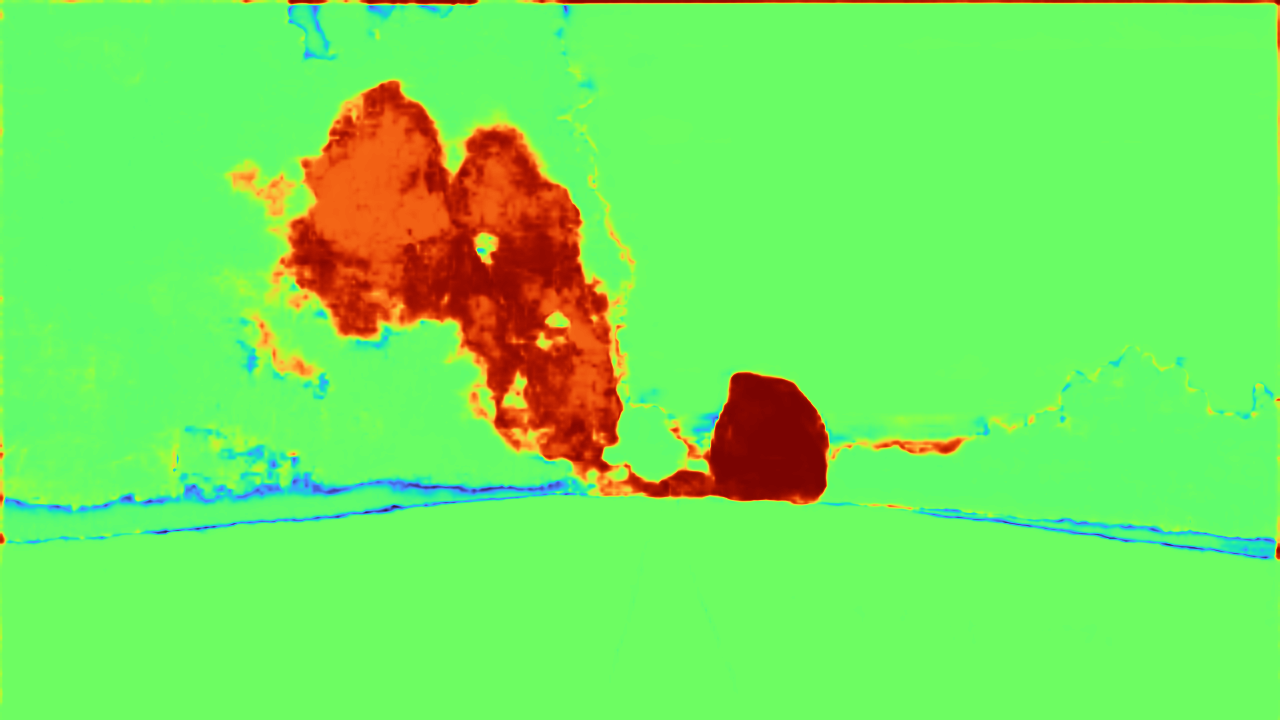}
    \end{subfigure}
    \begin{subfigure}[b]{0.18\textwidth}
        \centering
        \includegraphics[width=\textwidth]{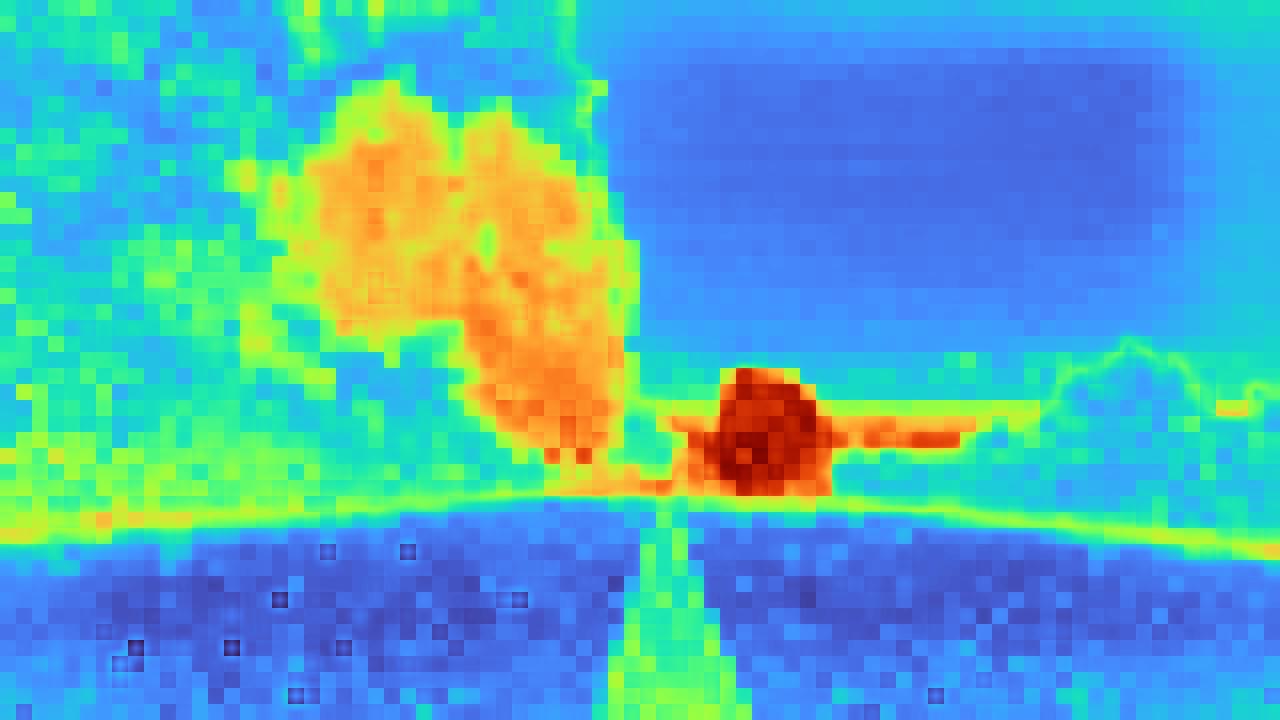}
    \end{subfigure}
    
    \begin{subfigure}[b]{0.18\textwidth}
        \centering
        \includegraphics[width=\textwidth]{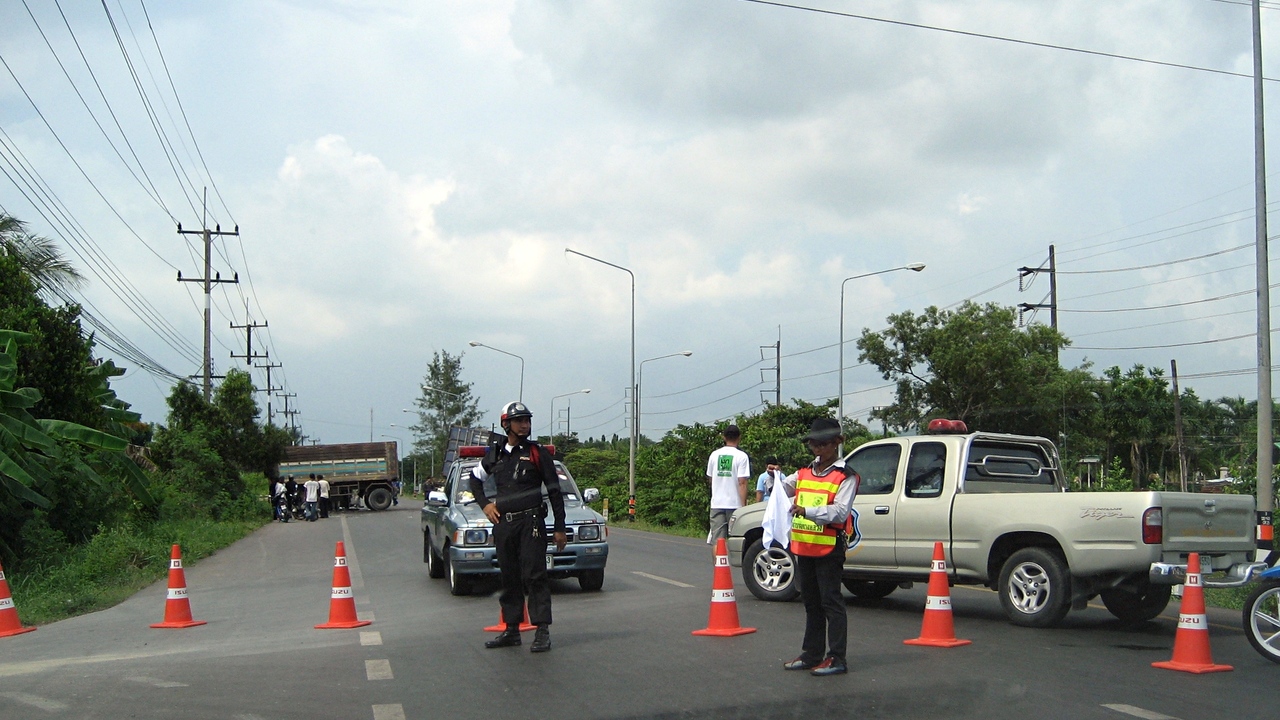}
        \caption*{\scriptsize Image}
    \end{subfigure}
    \begin{subfigure}[b]{0.18\textwidth}
        \centering
        \includegraphics[width=\textwidth]{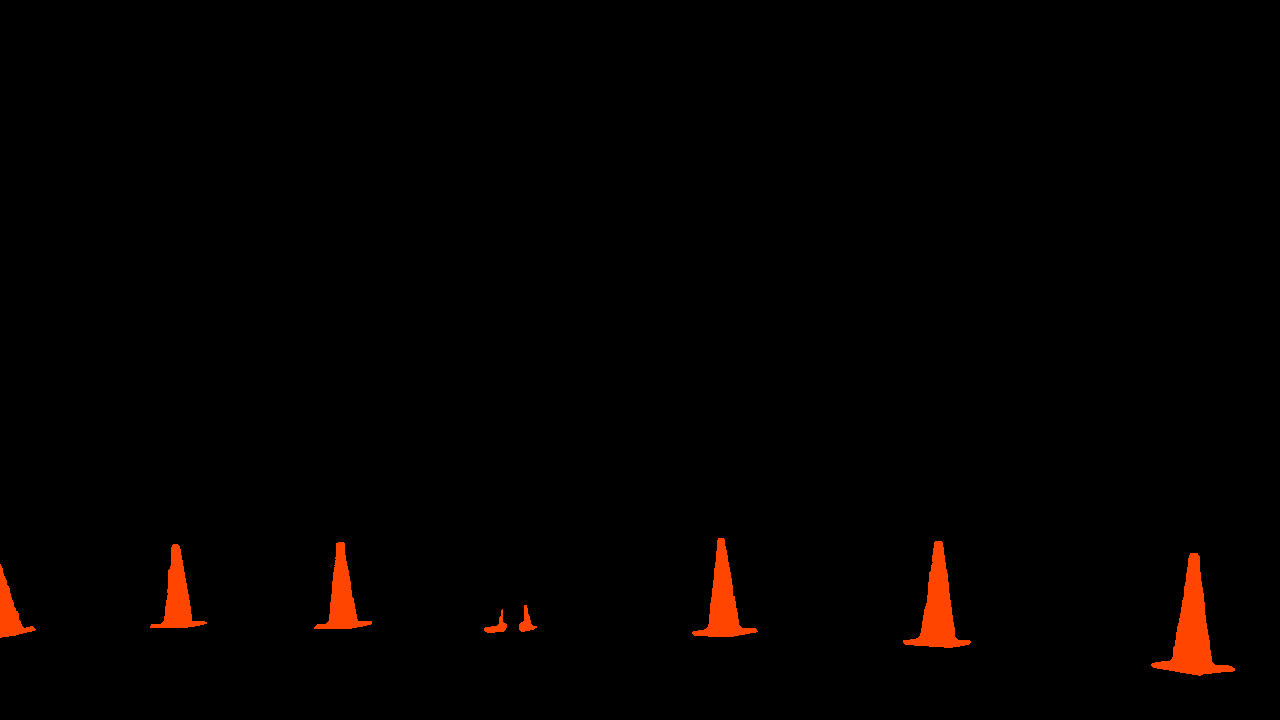}
        \caption*{\scriptsize Ground Truth}
    \end{subfigure}
    \begin{subfigure}[b]{0.18\textwidth}
        \centering
        \includegraphics[width=\textwidth]{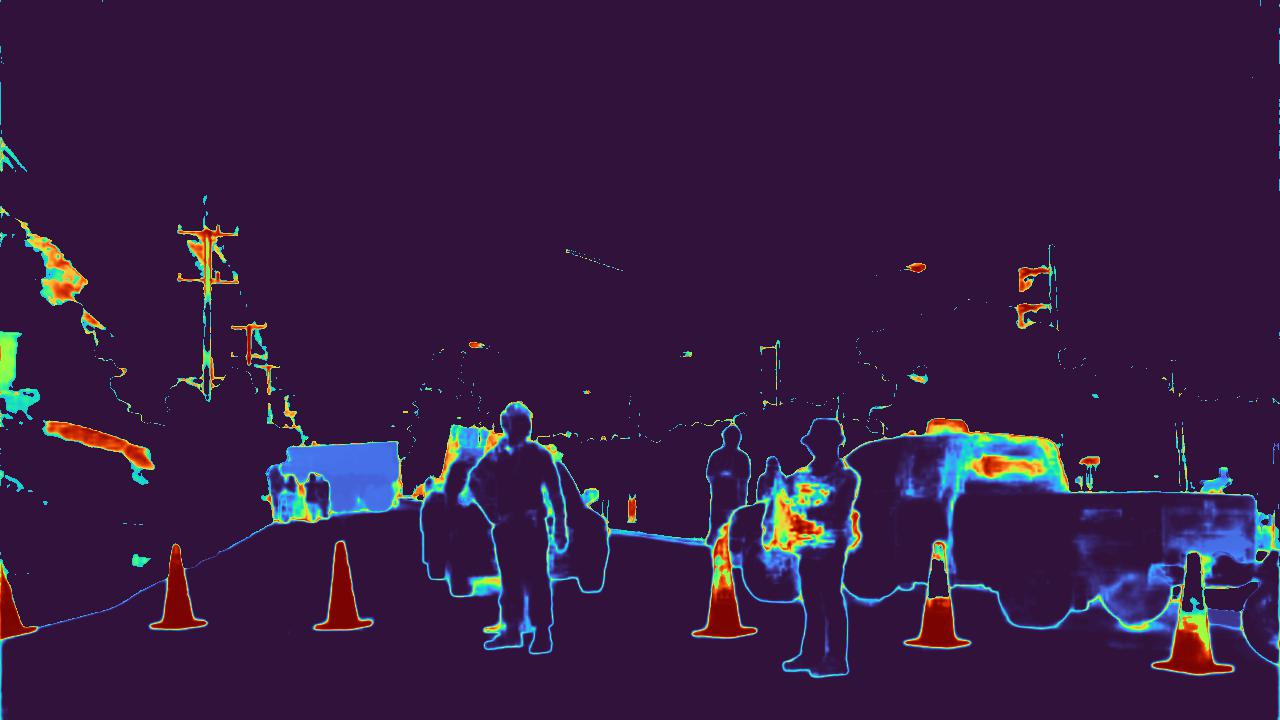}
        \caption*{\scriptsize OoD scores - M2A}
    \end{subfigure}
    \begin{subfigure}[b]{0.18\textwidth}
        \centering
        \includegraphics[width=\textwidth]{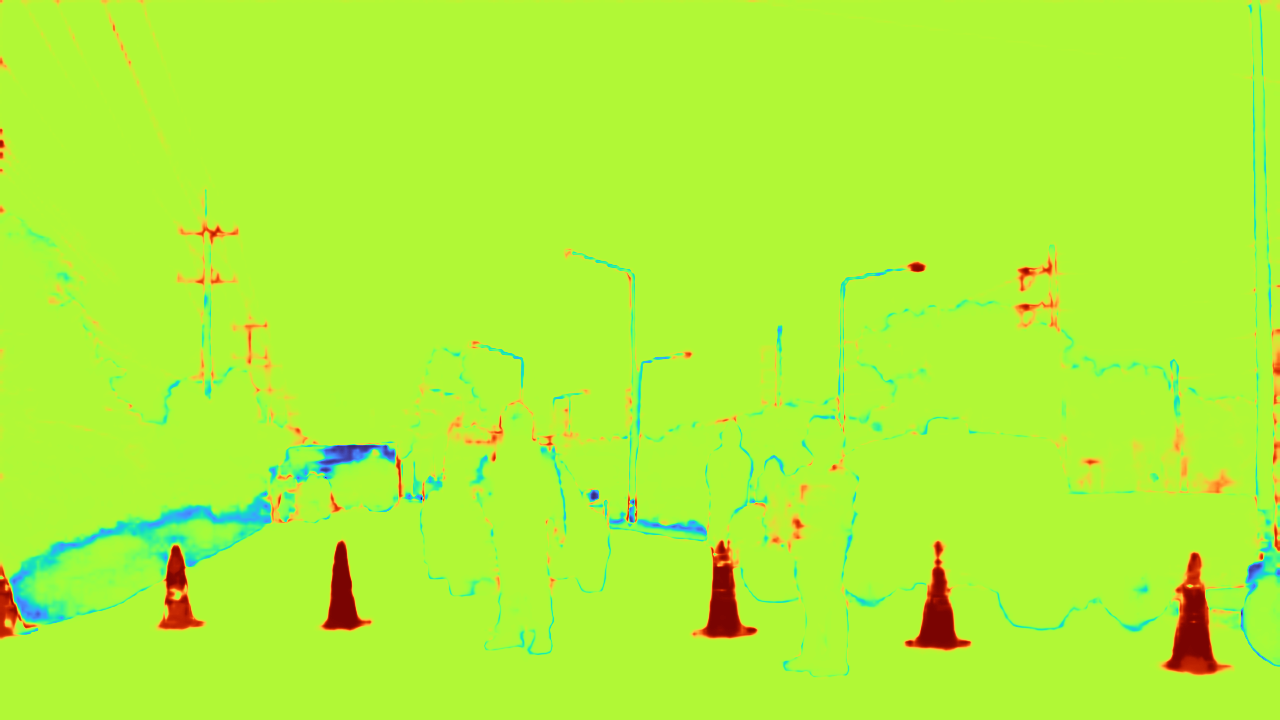}
        \caption*{\scriptsize OoD scores - RbA}
    \end{subfigure}
    \begin{subfigure}[b]{0.18\textwidth}
        \centering
        \includegraphics[width=\textwidth]{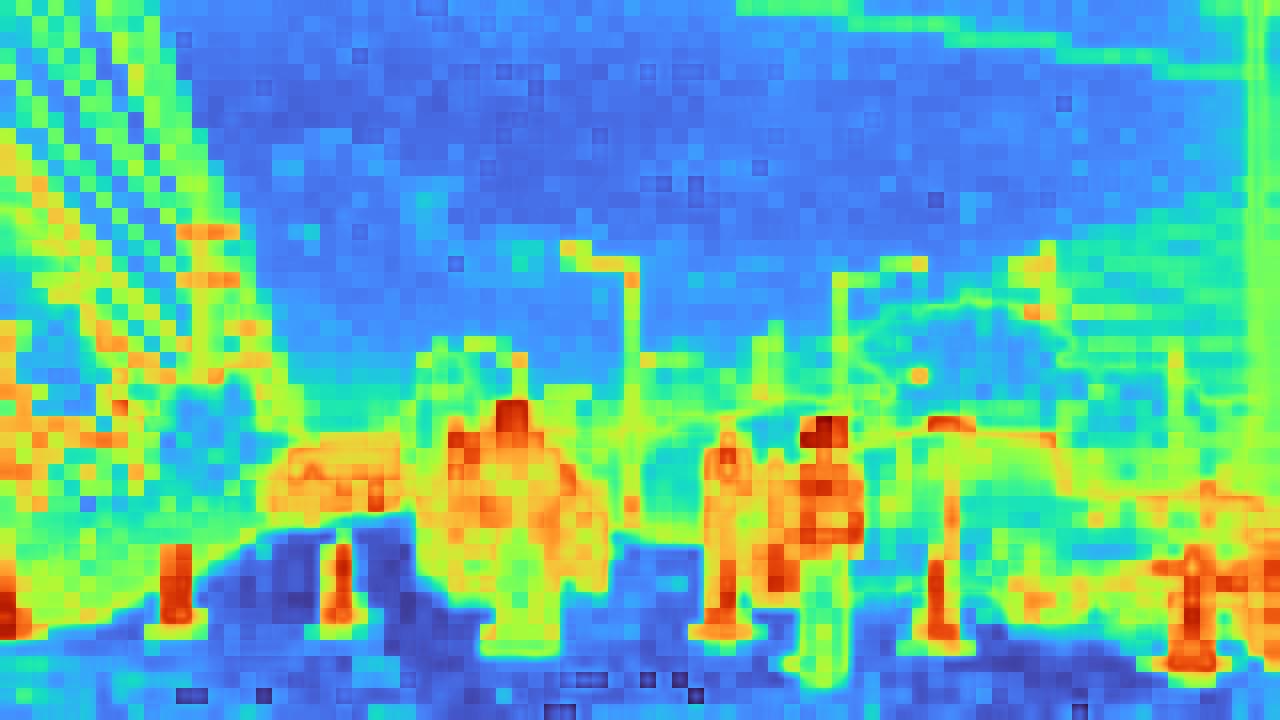}
        \caption*{\scriptsize OoD scores - Ours}
    \end{subfigure}
    
    \caption{\textbf{Qualitative comparison of our approach, Mask2Anomaly (M2A), and RbA.} The examples show images from RoadAnomaly, along with the OoD ground truth (red=OoD) and the predicted OoD scores for each method. \textcolor{myminturbo}{Purple} indicates low and \textcolor{mymaxturbo}{red} indicates high OoD scores. In all examples the methods produce some false positive predictions (i.e. the scores are high but the ground truth is black). However while our method mostly still produces lower scores than for real anomalies, M2A and RbA produce very high OoD scores, indistinguishable from the scores of true anomalies. This is, for example, visible in the street sign (first row), the guard rail (second row), the unpaved road and the building (third row).
    While all these examples are undoubtedly challenging, a good OoD detection system should still assign them lower scores.
    The last example shows a failure of our approach, which produces false positives for several small objects. 
    }
    \label{fig:quali_RA}
\end{figure*}

In \cref{fig:quali_RA} we show qualitative examples showcasing the differences between the predictions of our approach DOoD, Mask2Anomaly (M2A)\cite{rai2023unmasking}, and RbA~\cite{RbA}. The examples highlight the respective advantages and disadvantages of the approaches, all of which are high-performing, despite being based on very different paradigms.

The predictions from M2A and RbA are both based on Mask2Former, which decouples the prediction of masks and categories, as opposed to the typical per-pixel classification paradigm. For this reason, the scores are very uniform over the pixel sets corresponding to the same mask, and have smooth edges. In contrast to this, the predictions from our method are strictly local (as they are computed on each patch individually) and have no enforced spatial correlation.

Even though the predictions of M2A and RbA are more ``tidy'', the approaches have limitations. In the examples shown, it can be seen that where all approaches produce false positives, the two Mask2Former-based approaches have -- mistakenly -- high OoD scores. This can happen also for large sets of pixels, if they are predicted to belong to the same mask by Mask2Former, resulting in situations that are harmful for the benchmark results, and critical in real life applications.

Our approach DOoD, on the other hand, can struggle with small objects: in the last example the (out-of-distribution) street cones are assigned OoD scores which are partly surpassed by the scores of other small rare objects.

\section{Ablation: Compounding OoD Scores from the Diffusion Model and the Segmentation Model}
As described in the main paper (Sec. 4.3), we compound OoD scores obtained from the diffusion model with the segmentation model uncertainty (using the LogSumExp scoring function). In this section we compare the results from both individually, and the compound ones.

\begin{table}[!h]
    \centering
    \caption{\textbf{OoD detection results before and after compounding.}}
    \begin{tabular}{l|cc|cc|cc}
    \toprule
         & \multicolumn{2}{c|}{RoadAnomaly} & \multicolumn{2}{c|}{Fishyscapes Static} & \multicolumn{2}{c}{\ADEOoD{}}\\
        Scores & \multicolumn{1}{c}{AP$\uparrow$} & \multicolumn{1}{c|}{\fpr$\downarrow$} & \multicolumn{1}{c}{AP$\uparrow$} & \multicolumn{1}{c|}{\fpr$\downarrow$} & \multicolumn{1}{c}{AP$\uparrow$} & \multicolumn{1}{c}{\fpr$\downarrow$} \\
        \midrule
        Segmentation uncertainty \,        & 63.5 & 33.3 & 24.1	& 23.6 & 45.9 & 55.3 \\
        Diffusion only & 87.1 & 9.0  & 75.2 & \textbf{3.9} & 48.6 & 44.3 \\
        \hline
        Compound         & \textbf{89.1} & \textbf{8.8}  & \textbf{79.3} & 6.5 & \textbf{63.0} & \textbf{36.5} \\
    \bottomrule
    \end{tabular}
    \label{tab:compound_scores_ablation}
\end{table}

\begin{figure}[!h]
    \centering
    \begin{tabular}{ccccccc}
    
    \rotatebox{90}{\tiny Image} & &
        \includegraphics[width=0.15\textwidth]{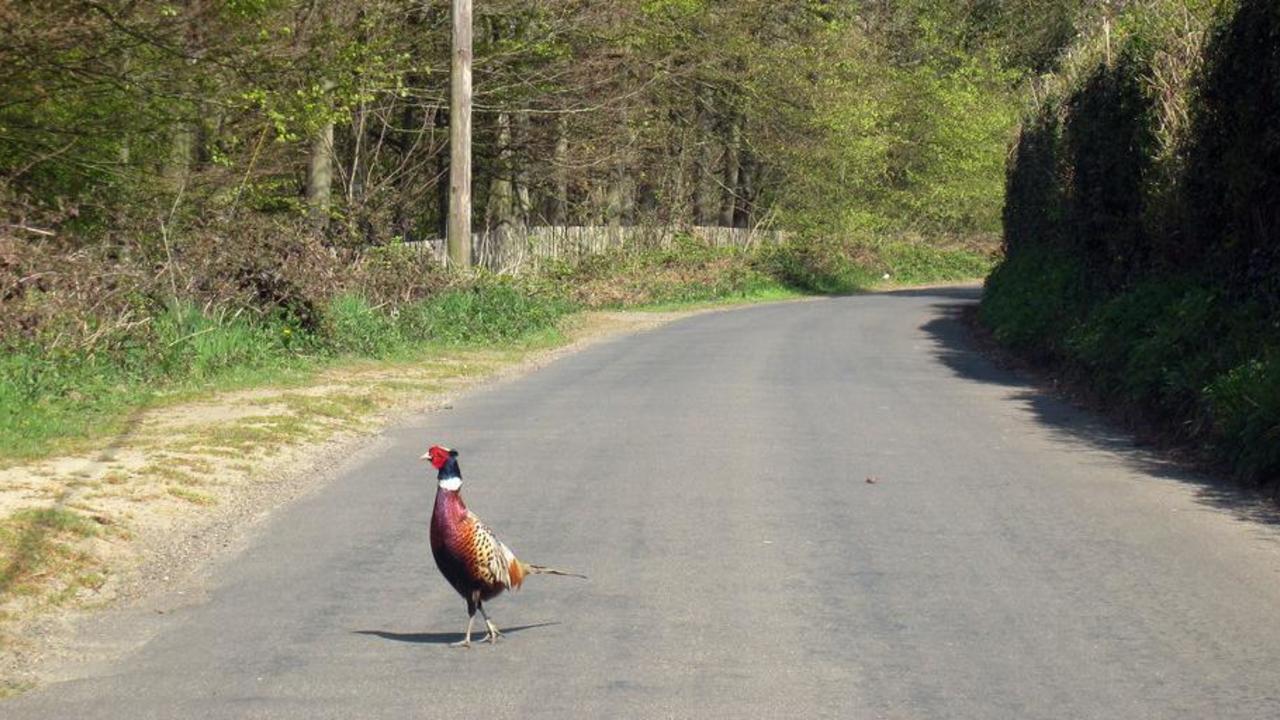}
    &
        
        \includegraphics[width=0.15\textwidth]{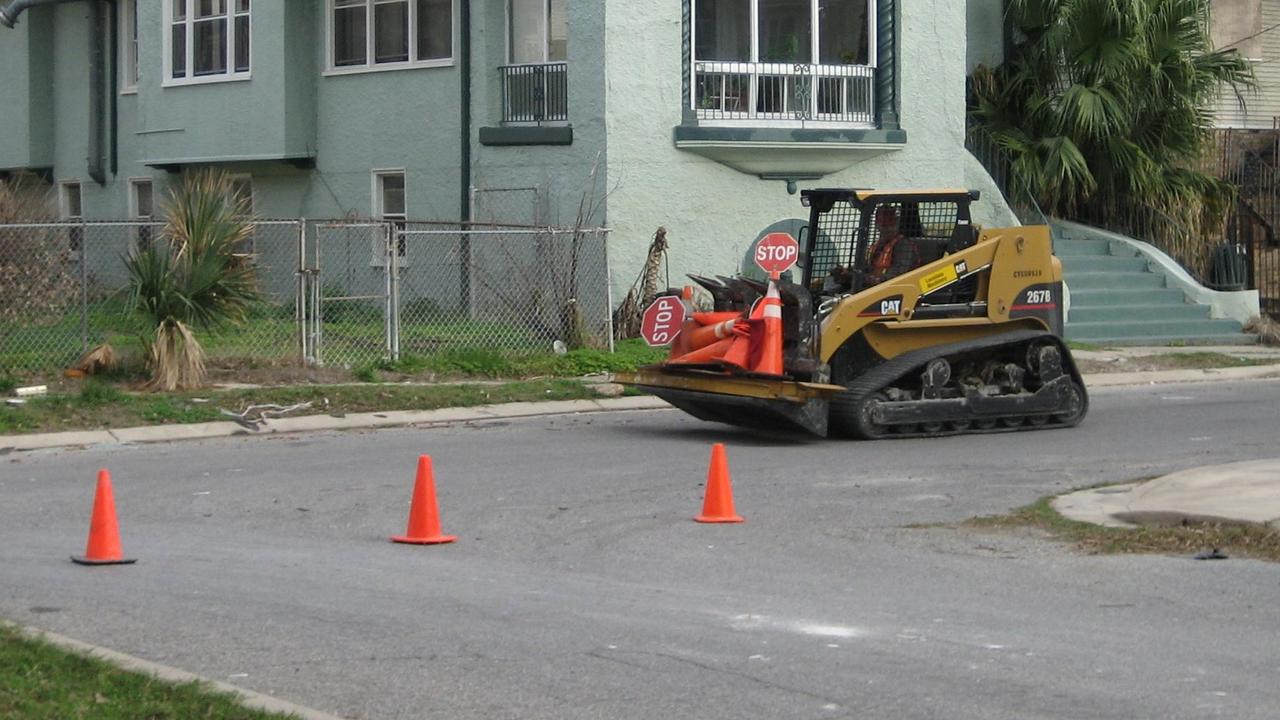}
    &
        
        \includegraphics[width=0.15\textwidth]{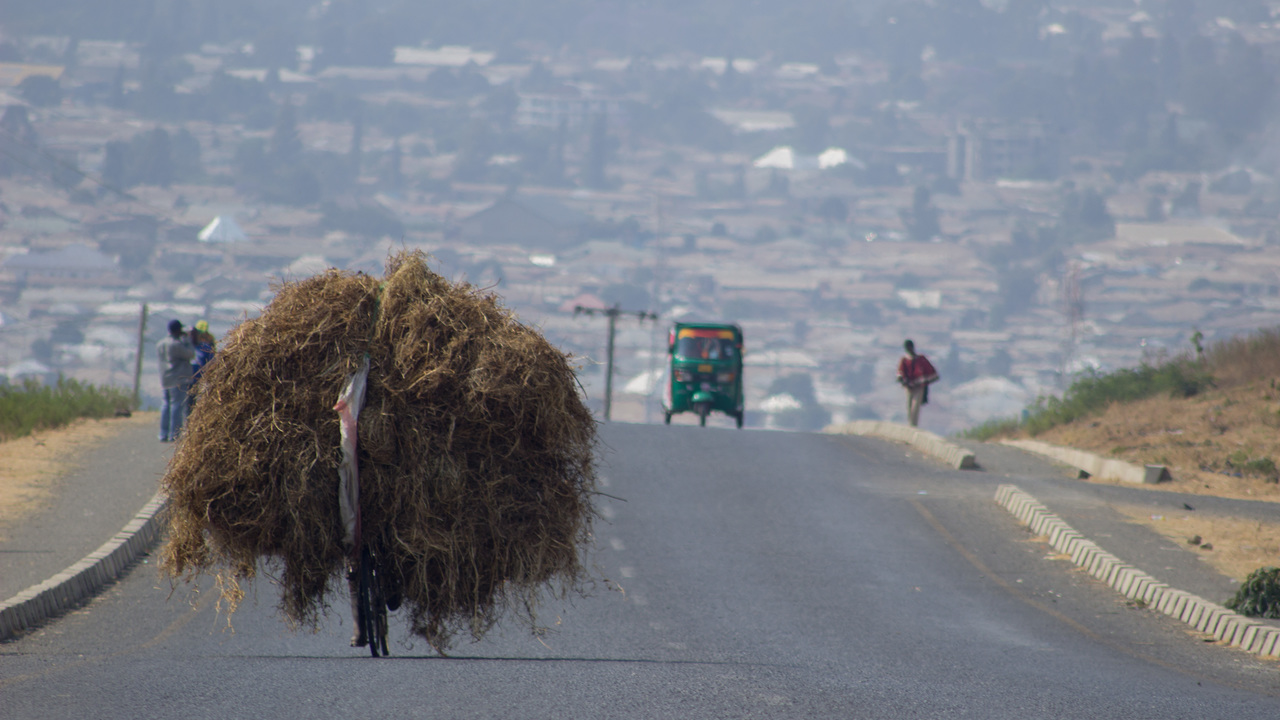}
    &
        
        \includegraphics[width=0.13\textwidth]{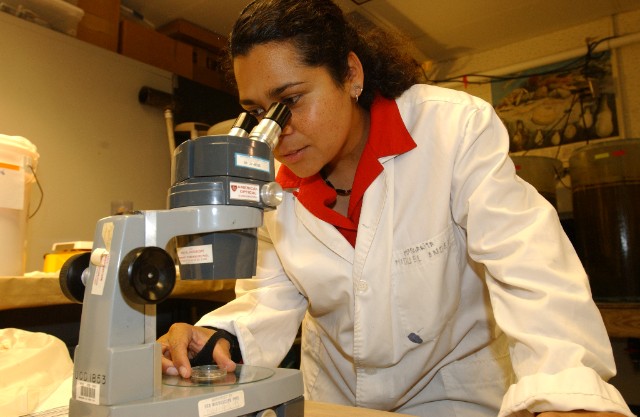}
    &
        
        \includegraphics[width=0.128\textwidth]{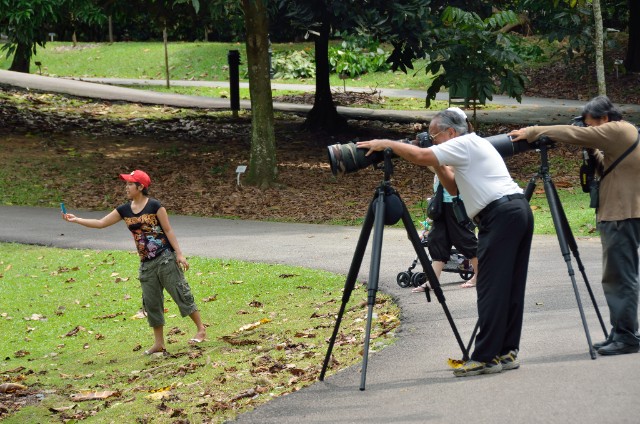}
    \\
    
    \rotatebox{90}{\tiny OoD} & \rotatebox{90}{\tiny Mask} &
        \includegraphics[width=0.15\textwidth]{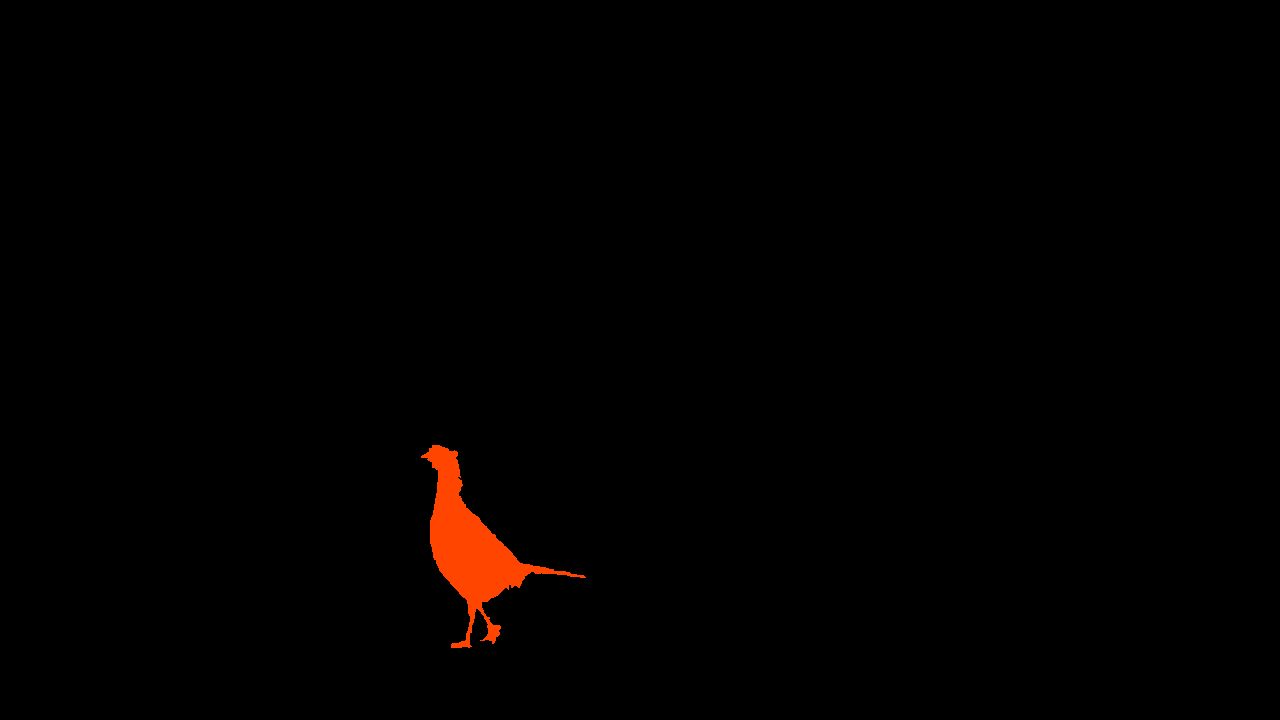}
    &
        
        \includegraphics[width=0.15\textwidth]{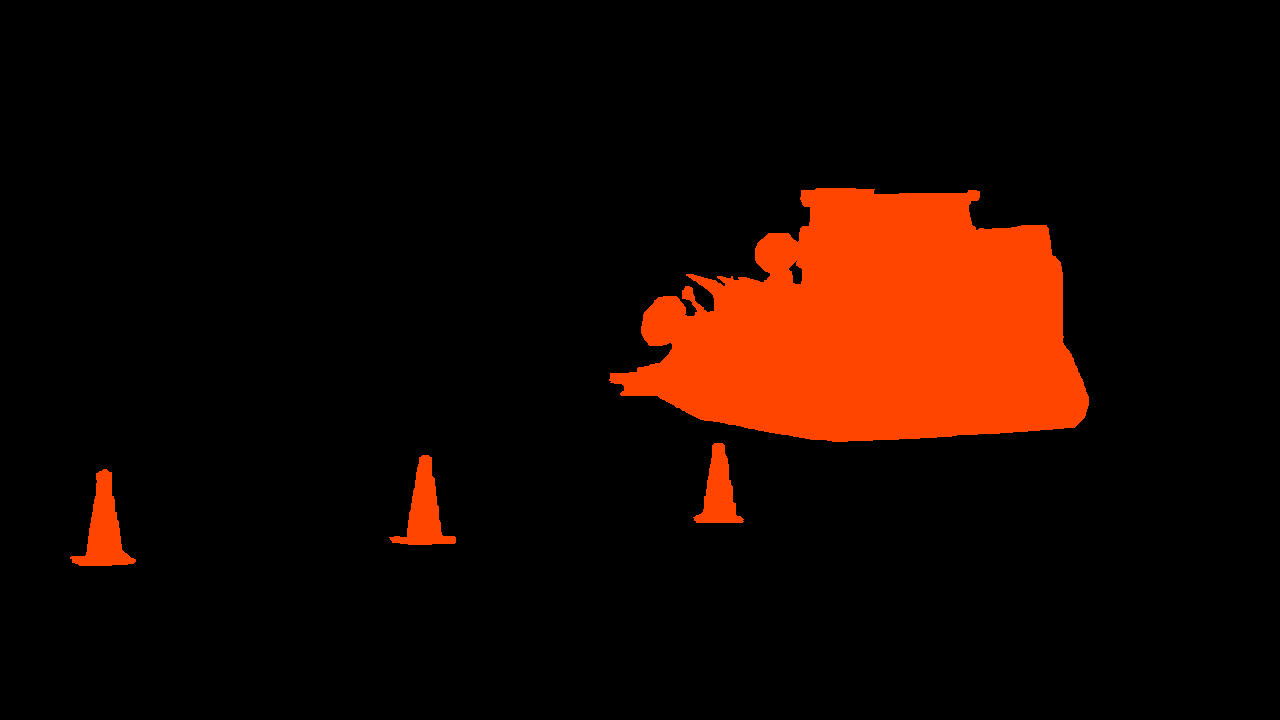}
    &
        
        \includegraphics[width=0.15\textwidth]{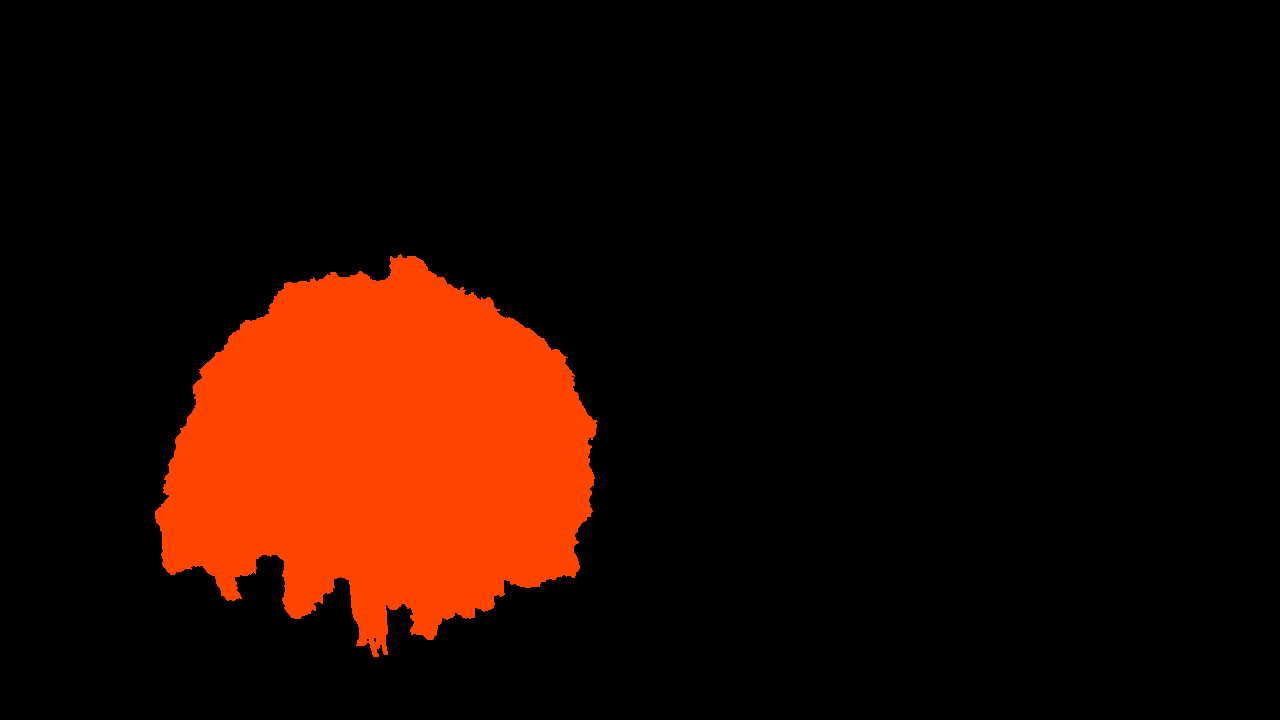}
    &
        
        \includegraphics[width=0.13\textwidth]{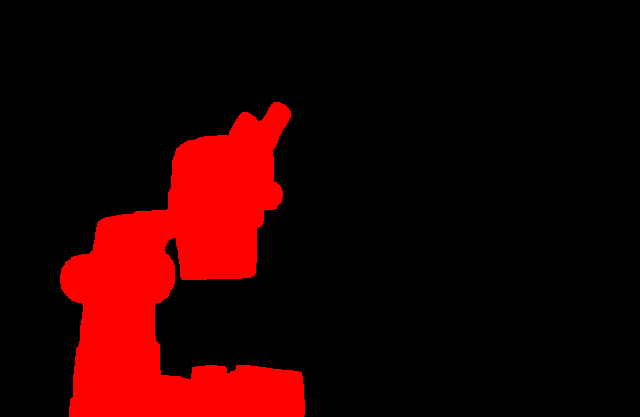}
    &
        
        \includegraphics[width=0.128\textwidth]{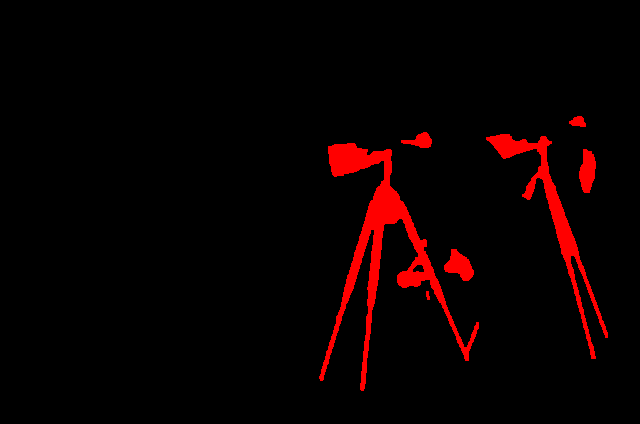}
    \\
    
    \rotatebox{90}{\tiny Diffusion} & \rotatebox{90}{\tiny scores} &
        \includegraphics[width=0.15\textwidth]{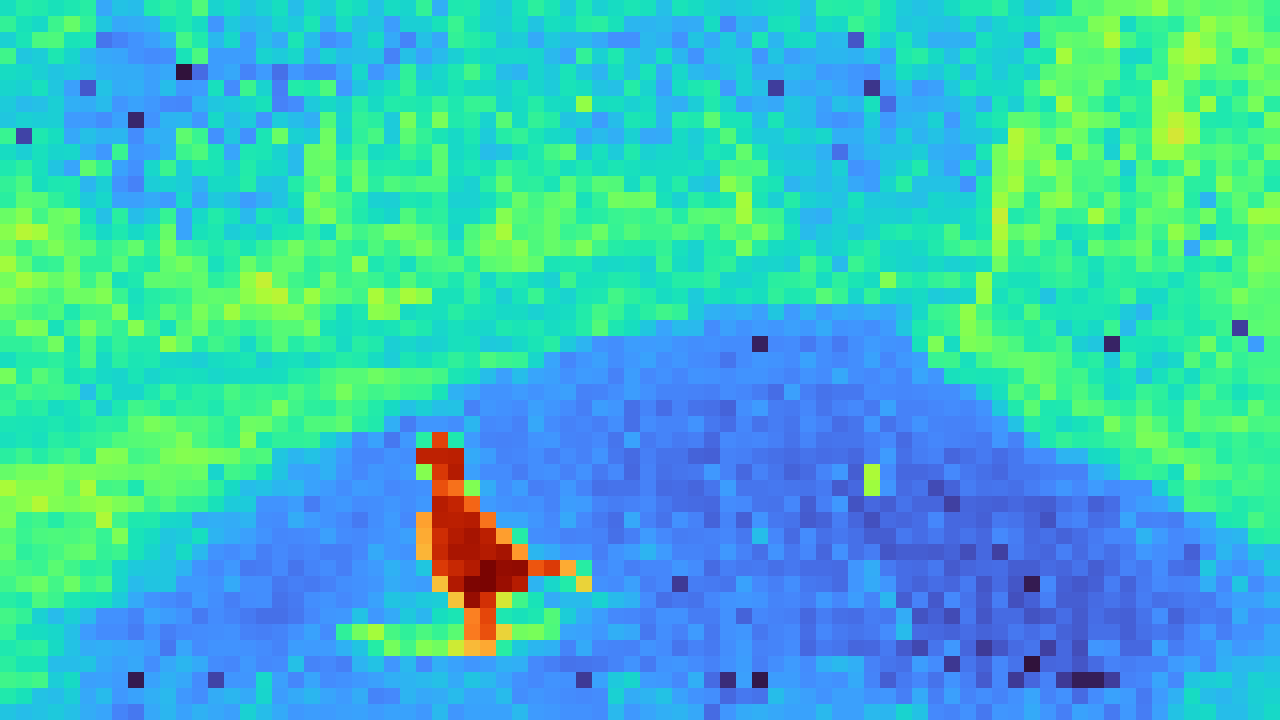}
    &
        
        \includegraphics[width=0.15\textwidth]{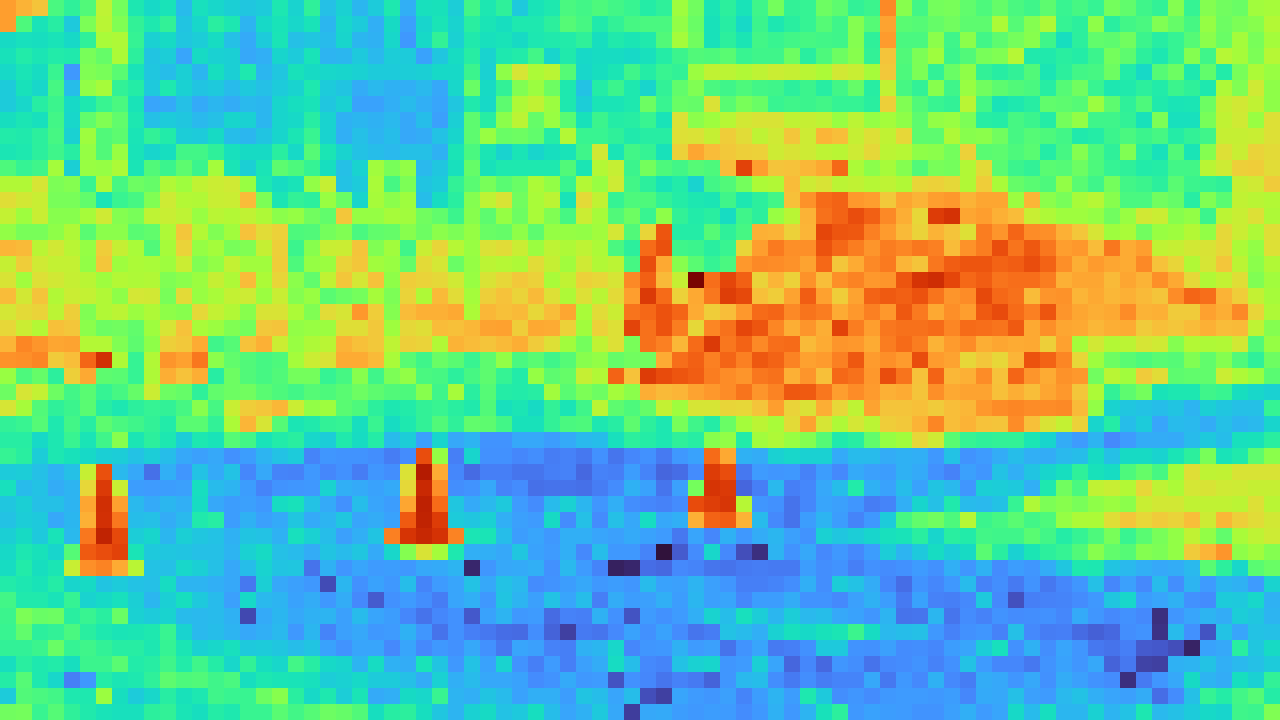}
    &
        
        \includegraphics[width=0.15\textwidth]{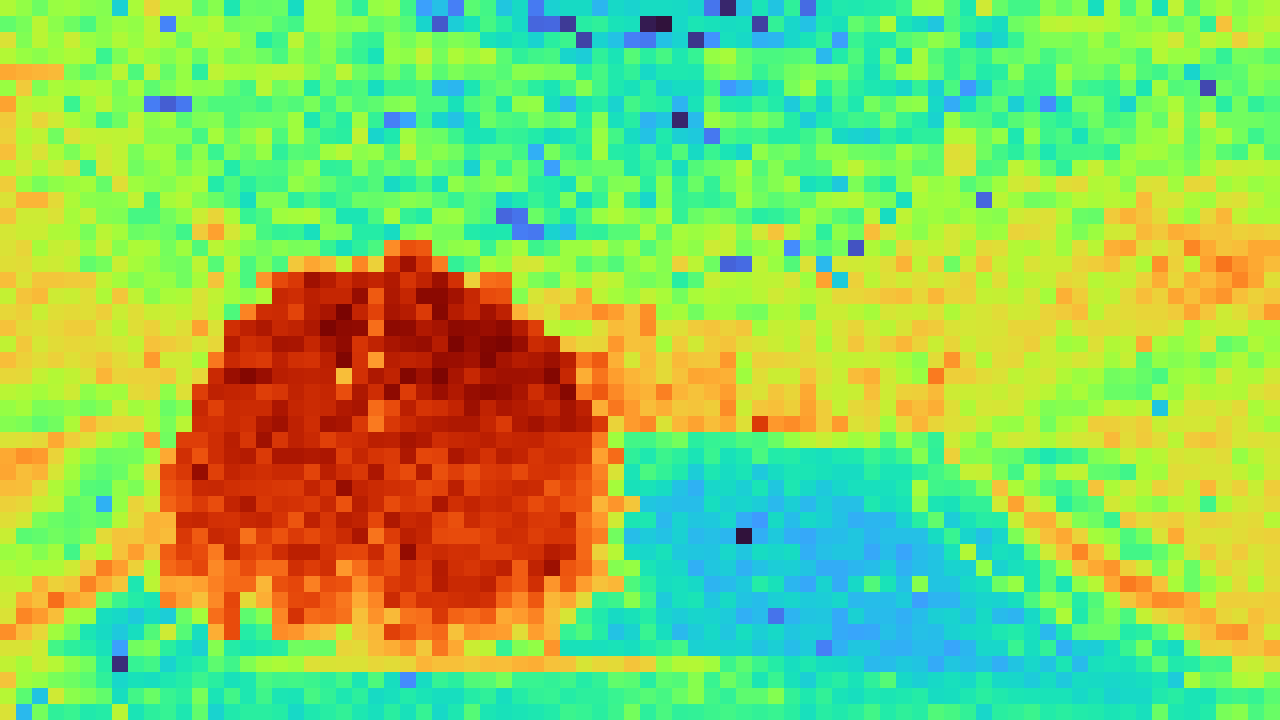}
    &
        
        \includegraphics[width=0.13\textwidth]{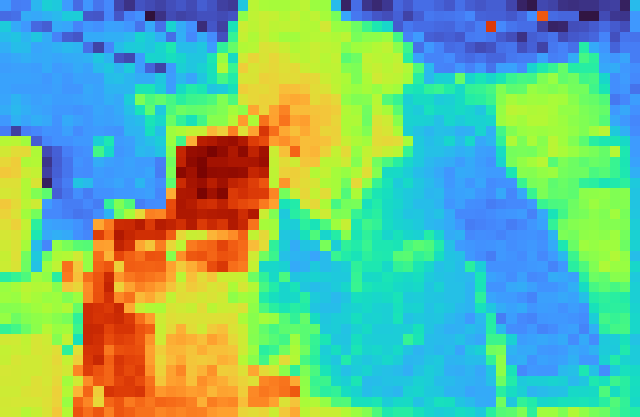}
    &
        
        \includegraphics[width=0.128\textwidth]{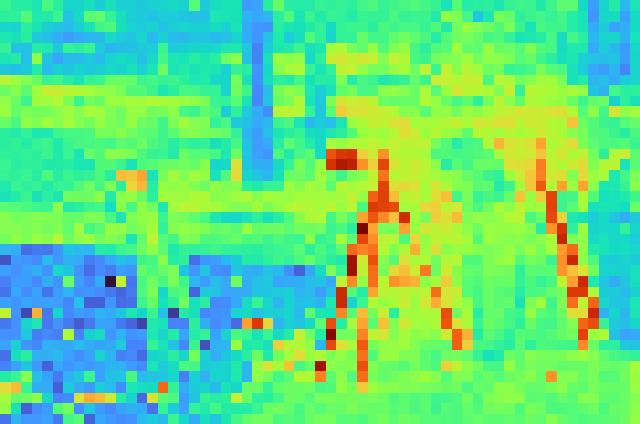}
    \\
    
    \rotatebox{90}{\tiny Compound} & \rotatebox{90}{\tiny scores} &
        \includegraphics[width=0.15\textwidth]{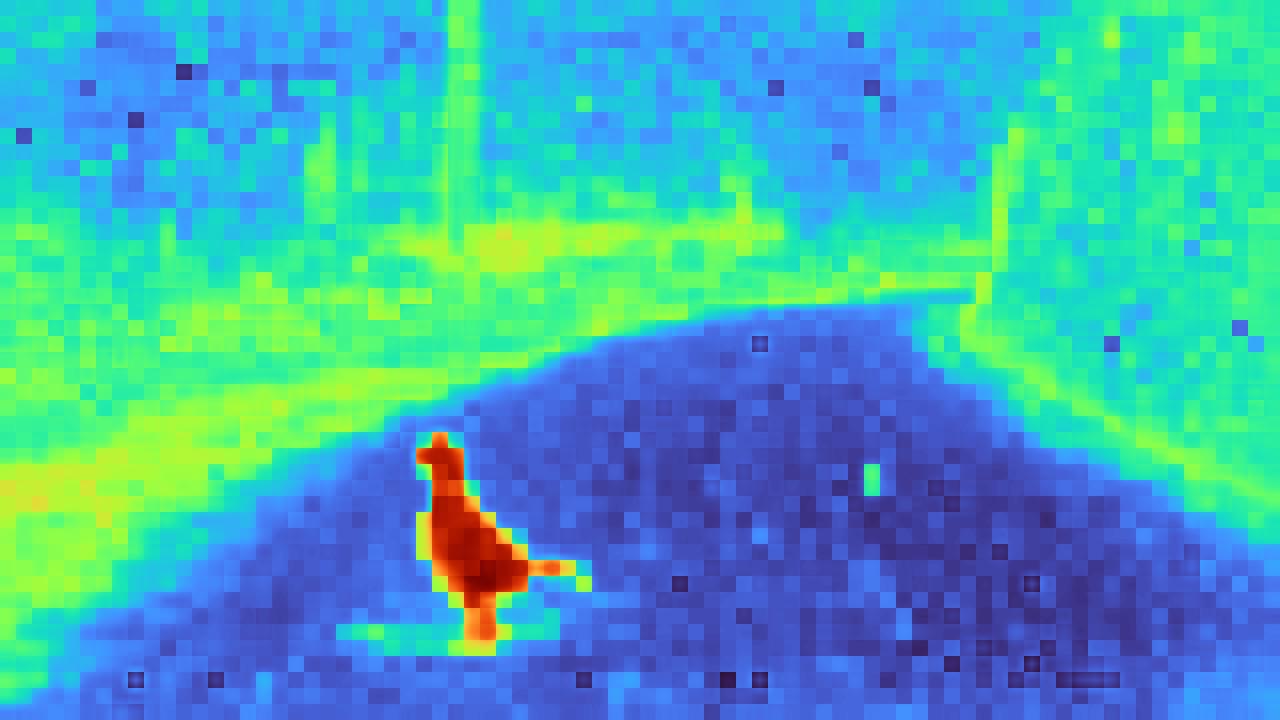}
    &
        
        \includegraphics[width=0.15\textwidth]{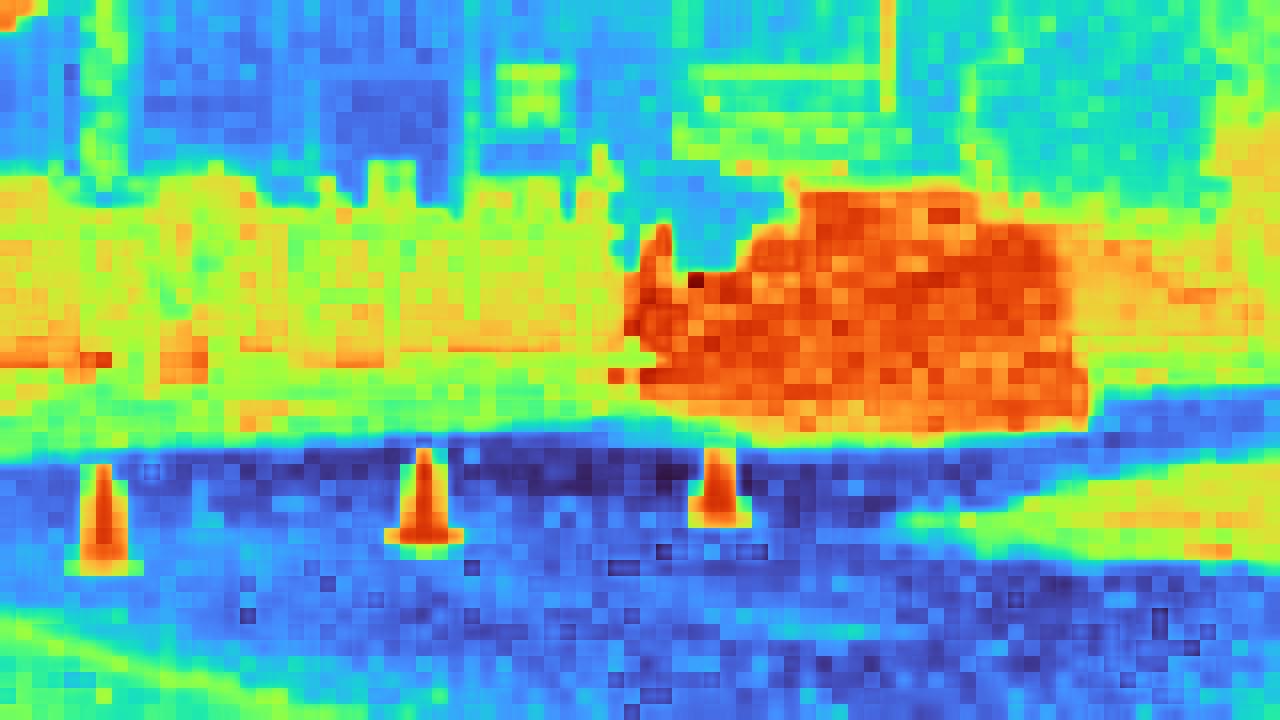}
    &
        
        \includegraphics[width=0.15\textwidth]{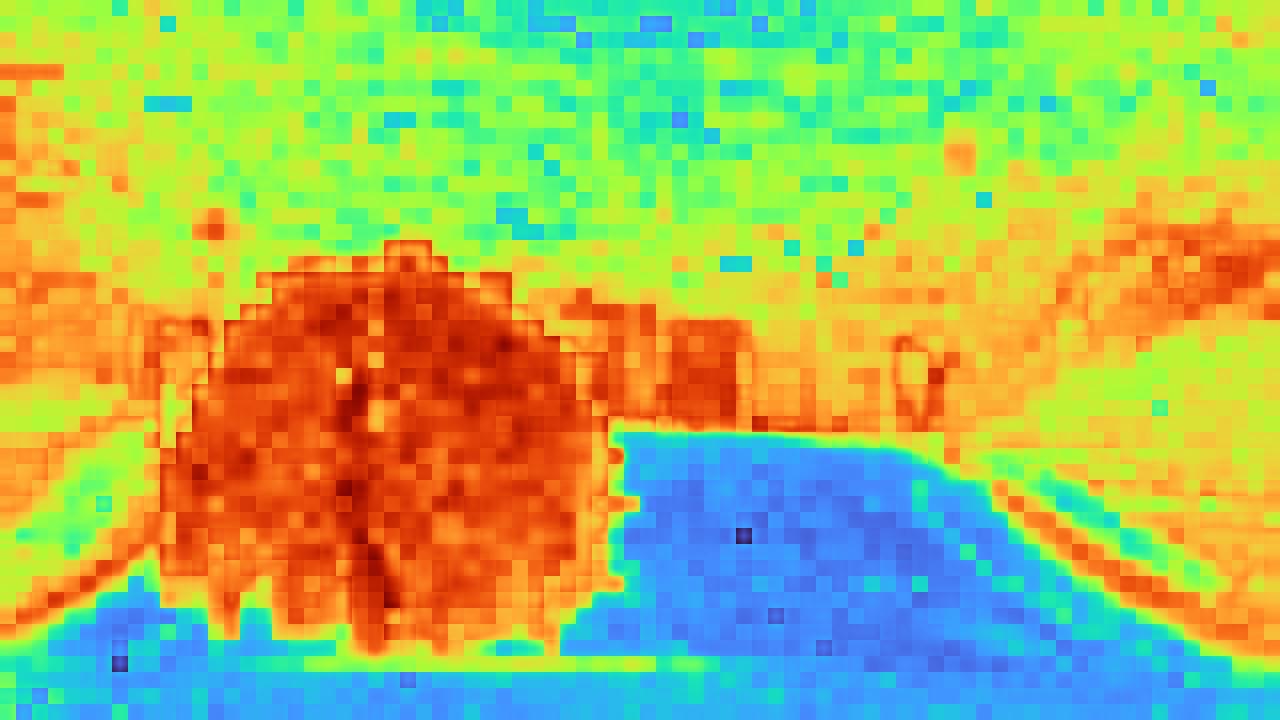}
    &
        
        \includegraphics[width=0.128\textwidth]{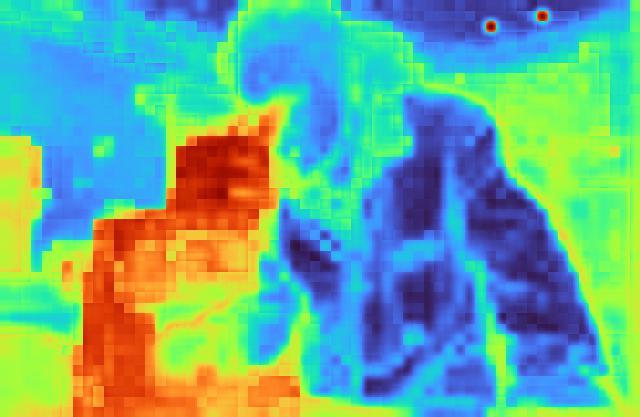}
    &
        
        \includegraphics[width=0.125\textwidth]{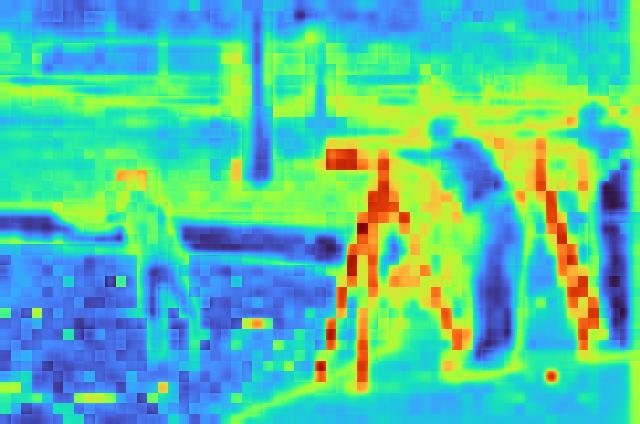}
    \end{tabular}
    
    \caption{\textbf{Qualitative comparison between pure diffusion scores and compound scores.} In the examples above, taken from the RoadAnomaly and ADE20k-OoD benchmarks, it can be seen that the diffusion scores already detect the anomalous objects before compounding. Compounding usually has a positive effect, noticeable for example in the example in the second column where it reduces the score magnitudes for background pixels and increases the ones corresponding to the vehicle. In some cases the effect is negative, such as in the example in the third column, where some in-distribution background objects provoke high uncertainty in the segmentation model.}
    \label{fig:diff_vs_unc}
    
\end{figure}

Quantitative results are shown in \cref{tab:compound_scores_ablation} on three benchmarks and qualitative examples are shown in \cref{fig:diff_vs_unc} for samples from RoadAnomaly and \ADEOoD{}.

$ $
\newline
The quantitative results clearly show that:
\begin{enumerate}
    \item Diffusion scores alone are always better than segmentation uncertainty.
    \item Compounding has a positive effect on the overall score, with its results being better than both components.
\end{enumerate}

\newpage
\section{Results on Dynamic Environmental Conditions}
We tested our approach DOoD on WD-Pascal~\cite{bevandic2019simultaneous}, a benchmark containing images of road scenes in adverse environmental conditions. Quantitative results, along with an example image from the benchmark and our method's prediction for it, are shown below:
\begin{figure}[!h]
\caption{\textbf{Results on dynamic environmental conditions.}}
\begin{minipage}[b]{0.4\linewidth}
    \centering
    \scriptsize
    \begin{tabular}{l|c|c}
        \toprule
         & AP & FPR \\
         \midrule
        cDNP & 68.55 & 12.05 \\
        RbA  & 73.00 & 9.58 \\
        DOoD (Ours) & \textbf{81.56} & \textbf{6.75} \\
        \bottomrule
    \end{tabular}
\end{minipage}
\begin{minipage}[b]{0.55\linewidth}
    \centering
    \setlength{\tabcolsep}{6pt}
    \begin{tabular}{cc}
        \includegraphics[width=0.38\textwidth]{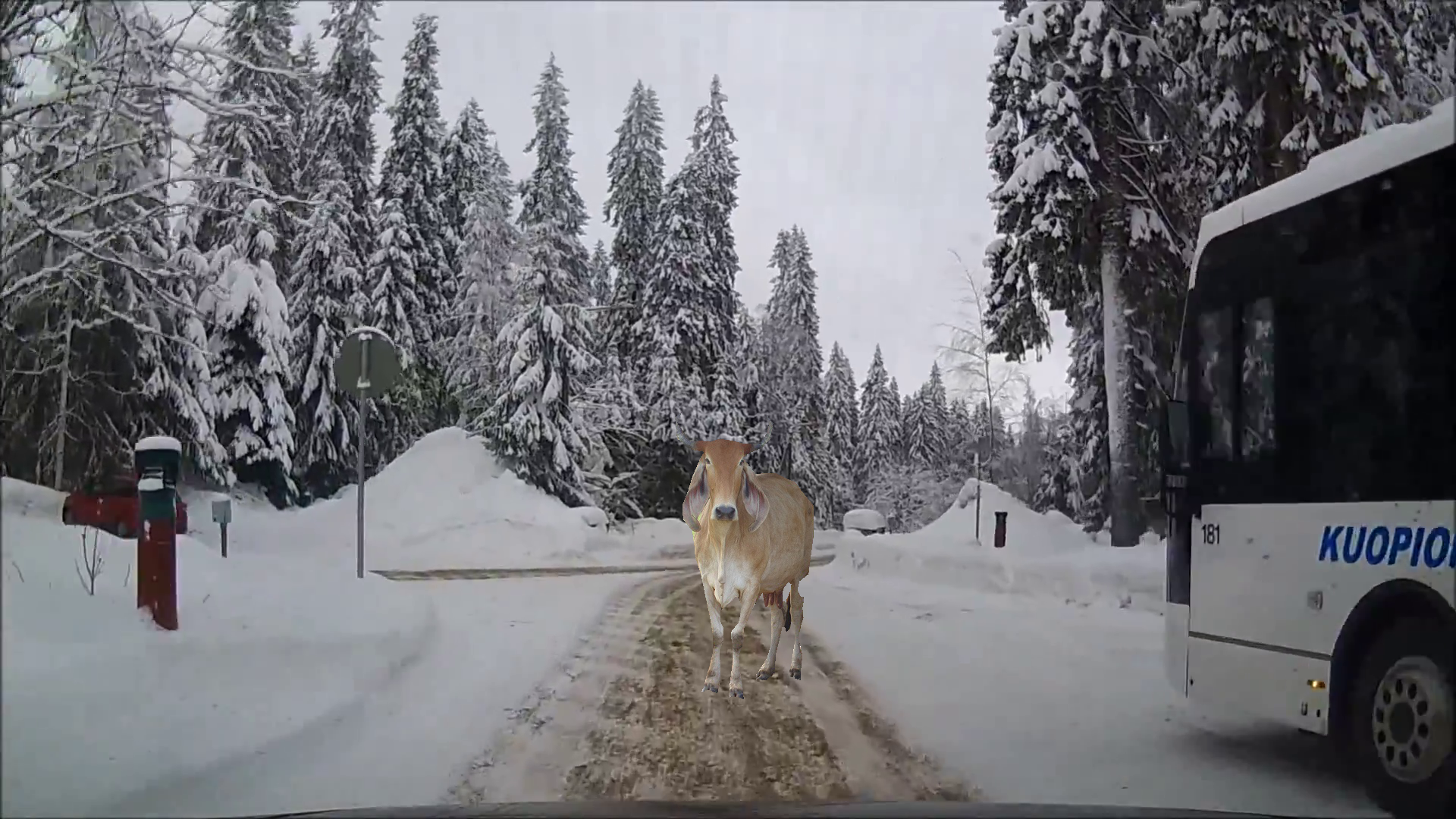} & \includegraphics[width=0.38\textwidth]{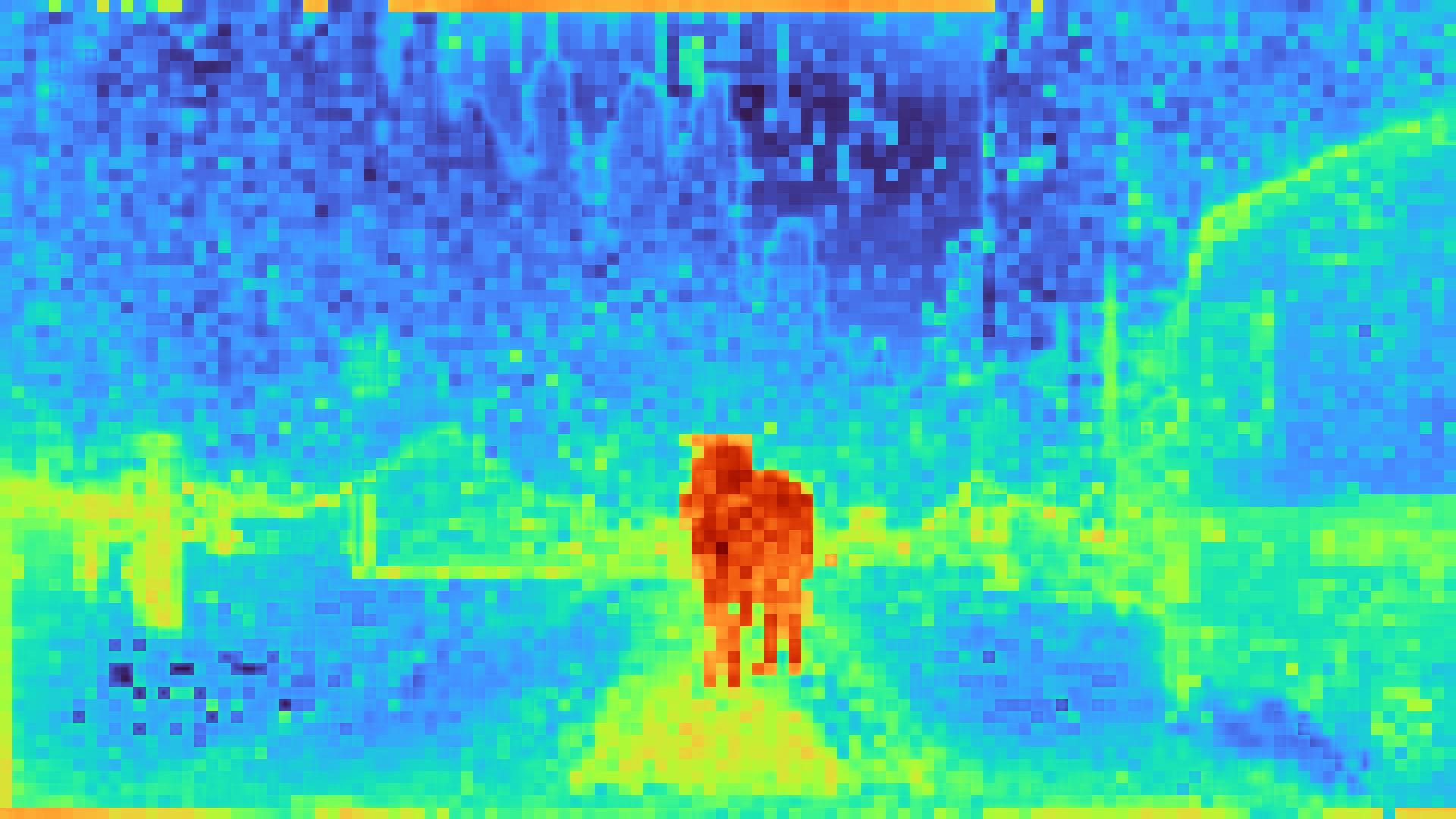}
    \end{tabular}
\end{minipage}
\end{figure}

\section{Results on Remote Sensing Data}

The focus of our work is out-of-distribution detection for semantic segmentation on general natural images. For completeness, in the following, we demonstrate the applicability of our method DOoD to even more general data domains via the example of remote sensing data. We constructed a benchmark by training on WHDLD~\cite{8954885} and testing on DLRSD~\cite{8954885}, using classes only found in the latter (airplane, car, court, sand, ship, tank) as OoD. A qualitative example and early results on this benchmark are shown below:
\begin{figure}[!h]
\caption{\textbf{Early results on remote sensing data.}}
\begin{minipage}[b]{0.55\linewidth}
    \centering
    \setlength{\tabcolsep}{4pt}
    \begin{tabular}{ccc}
        \includegraphics[width=0.23\textwidth]{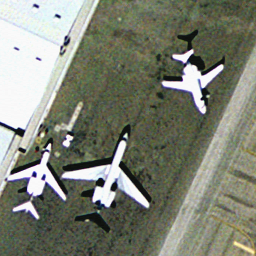} & \includegraphics[width=0.23\textwidth]{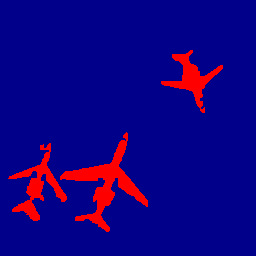} &
        \includegraphics[width=0.23\textwidth]{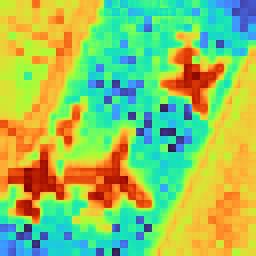}
    \end{tabular}
\end{minipage}
\begin{minipage}[b]{0.4\linewidth}
    \centering
    \small
    \begin{tabular}{l|c|c}
        & AP & FPR \\
        \hline
        cDNP & 39.7 & 45.8 \\
        DOoD (Ours) & \textbf{48.4} & \textbf{42.5} \\
    \end{tabular}
    \label{tab:remote_sensing}
\end{minipage}
\end{figure}

%
%

\end{document}